%% file: main.tex
\documentclass{article}

\usepackage[preprint]{neurips_2026}

\usepackage[utf8]{inputenc}
\usepackage[T1]{fontenc}
\usepackage{hyperref}
\usepackage{url}
\usepackage{booktabs}
\usepackage{amsfonts}
\usepackage{amssymb}
\usepackage{pifont}
\usepackage{graphicx}
\usepackage{wrapfig}
\usepackage{makecell}
\usepackage{nicefrac}
\usepackage{microtype}
\usepackage[table]{xcolor}
\usepackage{amsmath}
\usepackage[nameinlink,capitalise,noabbrev]{cleveref}
\usepackage{multirow}
\usepackage{subcaption}
\usepackage{tcolorbox}
\tcbuselibrary{skins}

\definecolor{ncmetabg}{HTML}{F1F4F7}
\definecolor{ncmetaedge}{HTML}{DCE6F5}
\definecolor{ncmetablue}{HTML}{1877F2}
\definecolor{ncapricot}{HTML}{F7EBDD}

\newcommand{\evalname}{PhyGround}

\usepackage{xcolor}
\newif\ifshowcolors
\showcolorsfalse

\ifshowcolors

\else

\fi

\title{PhyGround: Benchmarking Physical Reasoning in Generative World Models}

\author{%
  \normalfont
  \textbf{Juyi Lin}$^{1}$, \textbf{Arash Akbari}$^{1}$, \textbf{Yumei He}$^{2}$, \textbf{Lin Zhao}$^{1}$,
  \textbf{Haichao Zhang}$^{1}$,  \textbf{Arman Akbari}$^{1}$,\\
  \textbf{Xingchen Xu}$^{3}$, \textbf{Zoe Y.~Lu}$^{2}$, \textbf{Enfu Nan}$^{1}$, \textbf{Hokin Deng}$^{4}$,
  \textbf{Edmund Yeh}$^{1}$,  \\
  \textbf{Sarah Ostadabbas}$^{1}$, \textbf{Yun Fu}$^{1}$, \textbf{Jennifer Dy}$^{1}$, \textbf{Pu Zhao}$^{1}$, \textbf{Yanzhi Wang}$^{1}$ \\[2pt]
  $^{1}$Northeastern University \quad
  $^{2}$Tulane University \\
  $^{3}$University of Washington \quad
  $^{4}$Carnegie Mellon University
}

\makeatletter
\newcommand{\maketitleboxed}[1]{%
  \par
  \begingroup
    \renewcommand{\thefootnote}{\fnsymbol{footnote}}
    \renewcommand{\@makefnmark}{\hbox to \z@{$^{\@thefnmark}$\hss}}
    \long\def\@makefntext##1{%
      \parindent 1em\noindent
      \hbox to 1.8em{\hss $\m@th ^{\@thefnmark}$}##1%
    }
    \thispagestyle{empty}%
    \begin{tcolorbox}[
      enhanced,
      colback=ncmetabg,
      colframe=ncmetaedge,
      boxrule=0.35pt,
      arc=12pt,
      left=0.55cm, right=0.55cm, top=0.45cm, bottom=0.4cm,
      interior style={shade, shading angle=315,
        left color=white!96!ncmetabg,
        right color=ncmetablue!4!ncapricot!8!ncmetabg},
      before skip=0pt, after skip=0.4em,
      grow to left by=1.5pt, grow to right by=1.5pt,
    ]
      \centering
      {\LARGE\bf \raisebox{-0.32\height}{\includegraphics[height=3.2em]{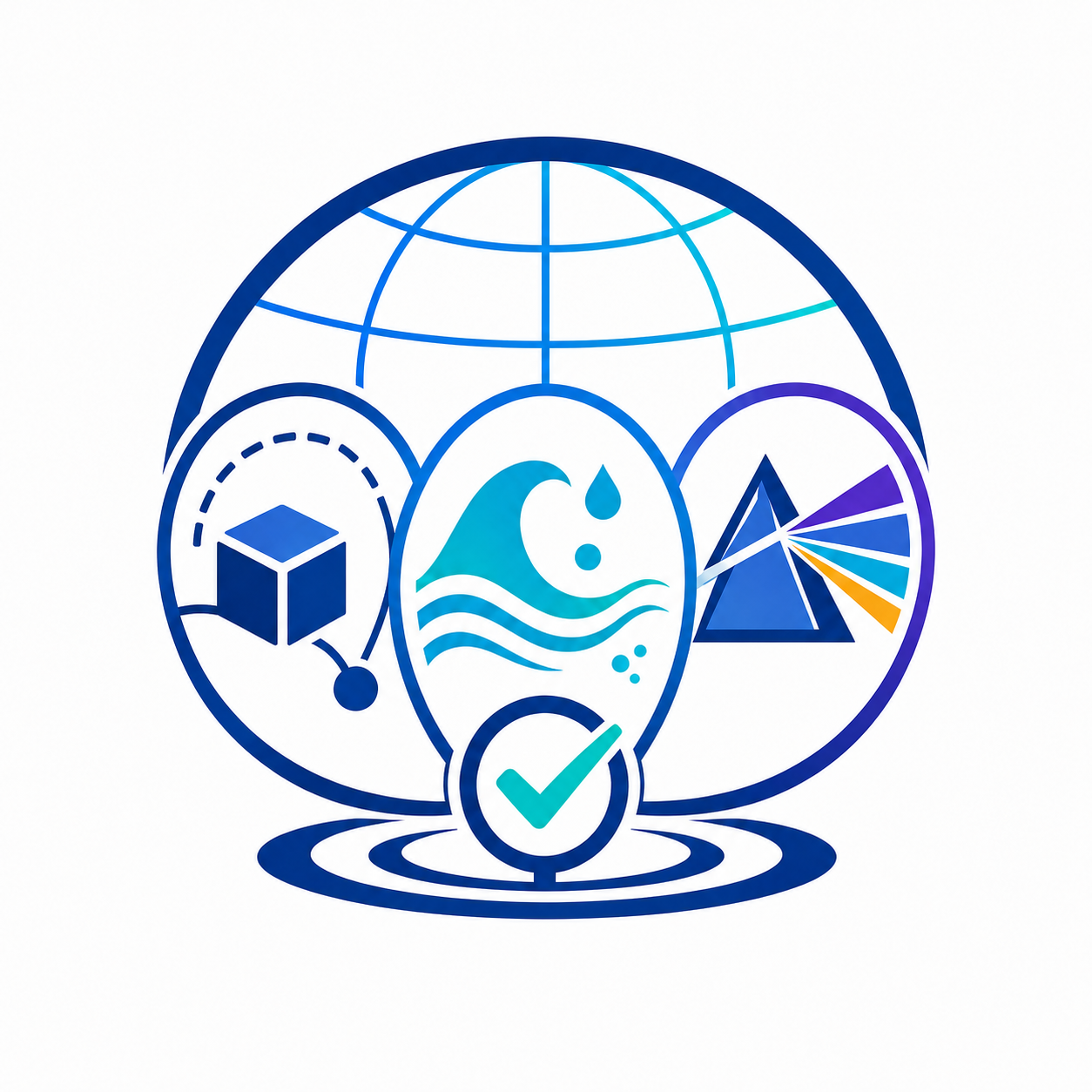}}\,\@title\par}%
      \vskip 0.22in
      \def\And{\end{tabular}\hfil\linebreak[0]\hfil\begin{tabular}[t]{c}\bf\rule{\z@}{24\p@}\ignorespaces}%
      \def\AND{\end{tabular}\hfil\linebreak[4]\hfil\begin{tabular}[t]{c}\bf\rule{\z@}{24\p@}\ignorespaces}%
      \begin{tabular}[t]{c}\bf\rule{\z@}{24\p@}\@author\end{tabular}\par
      \vskip 0.18in
      \begingroup
        \leftskip=1.5em \rightskip=1.5em
        \centerline{\large\bf Abstract}\vspace{0.6ex}
        \small #1\par
        \vspace{0.6ex}
        {\parfillskip=0pt plus 1fil\relax
         \noindent\small\textbf{Project page:}~\href{https://phyground.github.io/}{\color{blue}\texttt{https://phyground.github.io/}}\par}
      \endgroup
    \end{tcolorbox}%
    \@thanks
    \@notice
  \endgroup
  \let\maketitle\relax
  \let\thanks\relax
}
\makeatother

\begin{document}

\maketitleboxed{\input{sections/0_abstract}}

\begin{figure}[!ht]
\centering
\includegraphics[width=\linewidth]{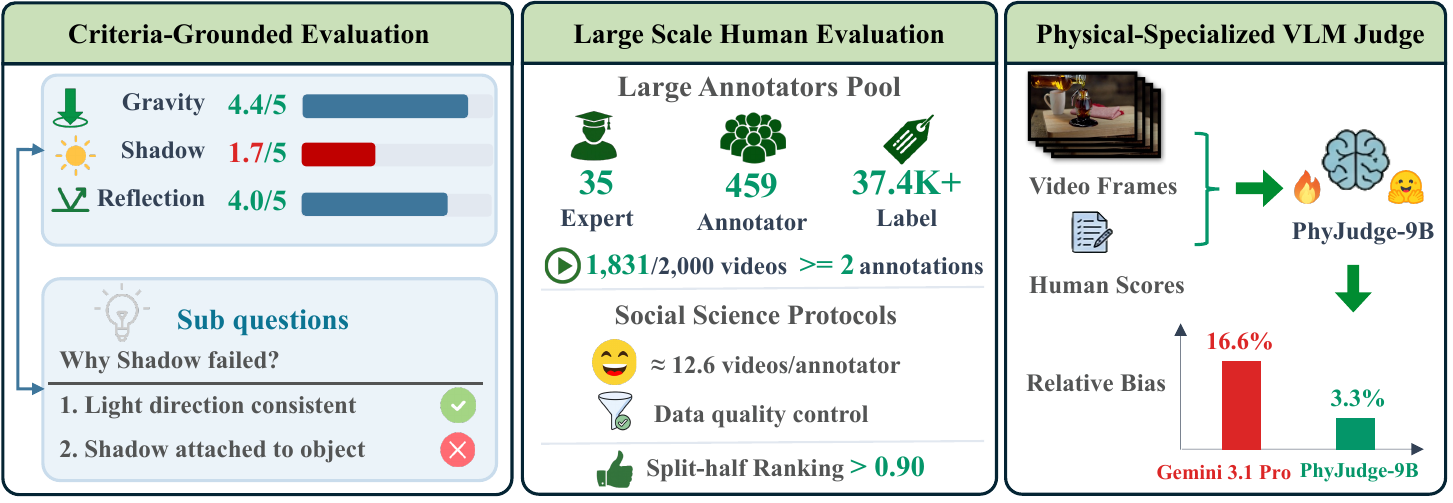}
\caption{\textbf{Overview of PhyGround}. PhyGround decomposes each video model's holistic physical reasoning score into scores for 13 physical laws. We recruited 459 annotators to conduct a large-scale, quality-controlled human study. Based on these human annotations, we released PhyJudge-9B, a fine-tuned judge model that supports reproducible automated evaluation.}
\label{fig:teaser}
\end{figure}

\input{sections/1_introduction}

\input{sections/2_related_work}

\input{sections/3_benchmark_design}
\input{sections/5_experiments}

\input{sections/6_conclusion}

\bibliographystyle{plain}
\bibliography{main}

\appendix

\input{appendix/law_selection}
\input{appendix/humaneval}

\input{appendix/aggregation_sensitivity}
\input{appendix/scoreperlaw}

\input{appendix/trainingdetail}
\input{appendix/vlmevalsetting}
\input{appendix/videogen}
\input{appendix/limitation}
\input{appendix/poorvideobylaw}
\input{appendix/poorvideobymodel}

\clearpage

\end{document}

%% file: sections/0_abstract.tex
Generative world models are increasingly used for video generation, where learned simulators are expected to capture the physical rules that govern real-world dynamics. However, evaluating whether generated videos actually follow these rules remains challenging. Existing physics-focused video benchmarks have made important progress, but they still face three key challenges, including the coarse evaluation frameworks that hide law-specific failures, response biases and fatigue that undermine the validity of annotation judgments, and automated evaluators that are insufficiently physics-aware or difficult to audit. To address those challenges, we introduce \textbf{PhyGround}, a criteria-grounded benchmark for evaluating physical reasoning in video generation. The benchmark contains 250 curated prompts, each augmented with an expected physical outcome, and a taxonomy of 13 physical laws across solid-body mechanics, fluid dynamics, and optics. Each law is operationalized through observable sub-questions to enable per-law diagnostics. We evaluate eight modern video generation models through a large-scale, quality-controlled human study, grounded on social science lab experiment design. A total of 459 annotators provided 5{,}796 complete annotations and over 37.4K fine-grained labels; after quality control, the retained annotations exhibited high split-half model-ranking correlations (Spearman's $\rho > 0.90$). To support reproducible automated evaluation, we release \textbf{PhyJudge-9B}, an open physics-specialized VLM judge. PhyJudge-9B achieves substantially lower aggregate relative bias than Gemini-3.1-Pro (3.3\% vs.\ 16.6\%). We release prompts, human annotations, model checkpoints, and evaluation code on the project page.

%% file: sections/1_introduction.tex
\section{Introduction}
\label{sec:intro}

Generative world models have been increasingly applied in video generation \cite{li2025worldmodelbench, wan2025,rupprecht2026human}. Ideally, world models should be capable of understanding and predicting physical phenomena \cite{ding2025understanding, li2025worldmodelbench}. However, the outputs of current models often violate basic physical principles. For example, objects may float without support, fluids may disappear or lose mass, and shadows may conflict with the underlying light source. These failures undermine the physical authenticity of generated videos and limit the applicability of world models in domains that require reliable simulation, prediction, and visual reasoning. To better understand the physical reasoning capability of current world models and identify areas for improvement, it is critical to evaluate the extent to which generated videos comply with physical laws. Such evaluation requires a reliable benchmark for measuring, comparing, and diagnosing the physical plausibility of outputs from world models. Although recent physics-focused video benchmarks have made meaningful progress in the assessment of physical adherence in generative world models, three key challenges remain unresolved.

First, existing evaluation frameworks often operate at the domain level, producing broad overall ratings that obscure law-specific failures. However, a physical domain is not a single capability. It contains multiple physical constraints that may be satisfied or violated independently. For example, a fluid scene may involve flow dynamics, boundary interaction, displacement, and continuity. A generated video may depict plausible flow while violating continuity by making liquid disappear or appear from nowhere. Averaging these distinct criteria into a single ``fluid dynamics'' score can therefore mask important failure modes and create a misleading impression of physical competence. Moreover, existing frameworks often rely on binary judgments, making it impossible to distinguish differences in severity (a slightly deviated trajectory versus an object flying upward against gravity).

Second, human evaluation for physics-focused video benchmarks is vulnerable to response biases rooted in human judgment. Annotators watch videos, interpret scoring rules, identify relevant visual evidence, and assign ratings under limited time and attention. If the annotation process is not carefully designed, benchmark scores may reflect fatigue, response shortcuts, and ordering effects. These concerns are well documented in social science research on response behavior \citep{reis2000handbook,campbell1963experimental,shadish2002experimental,krosnick1991response,orne1962social}. They are especially salient in physics-focused video evaluation, where annotators must distinguish videos that merely look plausible from those that actually satisfy the relevant physical constraints. A small annotator pool further amplifies this problem, as aggregate scores may be driven by individual biases rather than stable model performance.

Third, automatic evaluation remains insufficiently physics-aware and auditable. Standard video metrics such as FVD\cite{unterthiner2019fvd}, SSIM, PSNR\cite{hore2010imagepsnrssim}, and CLIP-based similarity\cite{radford2021learningclip} are useful for measuring distributional similarity, pixel-level fidelity, or semantic alignment, but they are not designed to detect violations of physical laws. Recent VLM-as-judge approaches provide a more semantic alternative, but many rely on closed-source models \cite{videphy2_2025, videosciencebench2025,li2025worldmodelbench}. Because their architectures, weights, training data, model versions, and API configurations evolve over time and are not fully controlled by benchmark users, the evaluation results may be difficult to reproduce or audit over time. Moreover, general-purpose VLM judges are not necessarily specialized for fine-grained physical diagnosis, especially when visually plausible videos contain subtle violations.

To address these challenges, we introduce \textbf{PhyGround}, a criteria-grounded benchmark for physical evaluation. As illustrated in Figure~1, it advances physics-focused video benchmarking in three ways.

\textbf{First, Criteria-Grounded Per-Law Evaluation Taxonomy.}
We construct a criteria-grounded physics taxonomy that evaluates physical reasoning at a more fine-grained level, both for specific laws and for sub-question design. 
The taxonomy covers three observable domains, including solid-body mechanics, fluid dynamics, and optics, and defines thirteen per-law criteria, including laws that are often under-specified in prior benchmarks, such as shadow consistency. Furthermore, each law is operationalized through sub-questions for VLMs and scored on a 1-5 Likert scale, capturing the extent of violations. As a result, PhyGround produces per-law and per-domain diagnostic scores.

\textbf{Second, Large-Scale, Quality-Controlled Human Evaluation.}
We conduct a large-scale human evaluation of eight video generation models using a controlled annotation protocol. A total of 459 annotators provided 5{,}796 complete annotations and over 37.4K fine-grained labels.
Beyond scale, we design the annotation process to improve reliability. For example, drawing on social science experimental protocols, we randomize the arrival of annotators and video presentation order, cap per-annotator workloads, and apply the survey design principles to make sure to improve task clarity and question accuracy. We then filter low-quality annotation based on multiple metrics, score consistency, dimension discriminability, peer agreement, and behavioral engagement. These procedures produce stable aggregate rankings, with split-half model-ranking correlations exceeding Spearman $\rho > 0.90$.

\textbf{Third, PhyJudge-9B, A Fully Open Physics-Specialized VLM Judge.} We release PhyJudge-9B, an open VLM judge fine-tuned on PhyGround human annotations.  It is supervised fine-tuned on structured score supervision from human annotations and is trained for PhyGround's per-law scoring task. Compared with closed-source baseline Gemini-3.1-Pro, PhyJudge-9B achieves substantially lower aggregate relative bias (3.3\% vs.\ 16.6\%). 

To summarize, PhyGround provides a fully open-source, criteria-grounded benchmark for physical evaluation. We hope to provide the community with a solid foundation for reproducible and auditable physical video evaluation. All benchmark prompts, human annotations, model checkpoints, and evaluation code will be publicly released.

%% file: sections/2_related_work.tex
\section{Related Work}
\label{sec:related}

\paragraph{Physics-focused Benchmarks for Video Generation.}
General-purpose video benchmarks such as EvalCrafter~\cite{liu2024evalcrafter} and VBench~\cite{huang2024vbench} rely on FVD, SSIM, and CLIP-style metrics that measure visual fidelity rather than physical correctness. Recently, a growing number of physics-focused benchmarks have emerged to address this gap\cite{phygenbench2024,wang2026veryvbvr,guo2025t2vphysbench,zhang2025morpheus,zhang2026physion,wang2025phydetex,cao2024physgame,sun2025contentcrave,xu2026visionreward,song2025vf}. 
VideoPhy~\cite{bansal2024videophy} and VideoPhy-2~\cite{videphy2_2025} use a binary followed/violated rubric that collapses mild and severe violations, with about ten annotators. PhyGenBench~\cite{phygenbench2024} curates only 160 manually crafted prompts and assumes each video corresponds to only one physical law.
Physics-IQ~\cite{physicsiq2026} grades 66 scenarios with reference-based pixel metrics that presume a unique ground-truth continuation and do not localize per-law failures. WorldModelBench~\cite{li2025worldmodelbench} has 350 prompts with 65 annotators but its binary 0/1 scoring obscures law-specific weaknesses, easily saturates at full scores, and thus lacks the resolution to distinguish the advantages of recent video generation models. A complementary line inverts the setup: Impossible Videos~\cite{bai2025impossible} prompts models to generate counterfactual scenes that violate physical, biological, or social laws, testing rule-breaking rather than rule-following. Its 260 prompts and GPT-4o-based judgments target creative generalization rather than per-law physical-correctness diagnosis. PhyWorldBench~\cite{gu2025phyworldbench} and VideoScience-Bench~\cite{videosciencebench2025} push other frontiers (anti-physics, undergraduate science) but rely on closed-source MLLM judges. 
\paragraph{VLM-as-a-judge for Video Physics Evaluation.}
Standard metrics such as FVD\cite{unterthiner2019fvd}, SSIM\cite{hore2010imagepsnrssim}, and CLIP\cite{radford2021learningclip} measure distributional similarity or semantic alignment, not violations of physical laws.
Learned video-quality metrics like VideoScore~\cite{he2024videoscore} score fine-grained quality dimensions but are not trained on physics-specific labels. Recent benchmarks pair their datasets with VLM-based automatic evaluators \cite{zheng2025vbench20,motamed2025travl,huangcosmoseval,han2025videobench,liu2025aigve,wang2025love,he2025videoscore2,liu2025improving,tong2025mj,wang2025aigv,zhang2025qbench,zhang2026vqinsight}.
On the open side, VideoPhy~\cite{bansal2024videophy} fine-tunes VideoCon-Physics, VideoPhy-2~\cite{videphy2_2025} releases VideoPhy-2-AutoEval, and WorldModelBench~\cite{li2025worldmodelbench} releases a 2B judger reported to outperform GPT-4o by 8.6\% on world-modeling violations.
These open evaluators are trained on holistic single-score labels and architecturally cannot produce per-law diagnostics or severity calibration even when reproducible. A second line uses closed-source VLMs: PhyGenEval~\cite{phygenbench2024}, VideoScience-Bench~\cite{videosciencebench2025}, and PhyWorldBench ~\cite{gu2025phyworldbench} all rely on GPT-4o or Gemini variants, raising reproducibility concerns because model versions, weights, and API behavior change outside benchmark users' control.

\paragraph{Positioning of PhyGround.}
\input{tables/table_benchmark_comparison.tex}
\Cref{tab:benchmark_comparison} compares PhyGround with six physics-focused video generation benchmarks. First, PhyGround supports more diagnostic evaluation by combining explicit expected-outcome prompts with per-law decomposition. Second, PhyGround provides a more reliable human-evaluation foundation through a substantially larger quality-controlled annotator pool. 
Third, PhyGround enables reproducible automated evaluation by providing an open-source judge.

%% file: tables/table_benchmark_comparison.tex
\begin{table}[!t]
\centering
\caption{Comparison of physics-focused video generation benchmarks. \textbf{Laws}: number of distinct physics laws checked at evaluation (\texttt{--} if not score at law granularity). \textbf{Annotators}: number of human annotators reported for evaluation (\texttt{--} if not reported). \textbf{Video/Ann.} $\downarrow$: average number of generated videos labeled per annotator (workload per annotator; lower means each annotator focuses on fewer videos). \textbf{Explicit}: prompts state the expected physical outcome. \textbf{Sub-Q}: automatic judge prompt includes law-specific sub-questions. \textbf{Open Judge}: open-source evaluator is provided.}
\label{tab:benchmark_comparison}
\small
\setlength{\tabcolsep}{4pt}
\begin{tabular}{lrrrrrccc}
\toprule
Benchmark & Prompts & Videos & Laws & Annotators & Video/Ann.\ $\downarrow$ & Explicit & Sub-Q & Open Judge \\
\midrule
PhyGenBench        & 160   & 1,280       & 27          & 3           & 512         & \ding{55} & \ding{51} & \ding{55} \\
Physics-IQ         & 66    & 396         & \texttt{--} & \texttt{--}           & \texttt{--}           & \ding{55} & \ding{55} & \ding{51} \\
VideoPhy-2         & 688   & 6,800       & 19          & 12          & 1,700       & \ding{55} & \ding{51} & \ding{51} \\
VideoScience-Bench & 200   & \texttt{--} & \texttt{--} & \texttt{--} & \texttt{--} & \ding{51} & \ding{51} & \ding{55} \\
WorldModelBench    & 350   & 4,421       & 5           & 65          & 128         & \ding{55} & \ding{55} & \ding{51} \\
Impossible Videos  & 260   & 902         & 6           & \texttt{--} & \texttt{--} & \ding{55} & \ding{55} & \ding{55} \\
\midrule
\textbf{PhyGround (Ours)}
& \textbf{250}
& \textbf{2,000}
& \textbf{13}
& \textbf{459}
& \textbf{12.6}
& \ding{51}
& \ding{51}
& \ding{51} \\
\bottomrule
\end{tabular}
\end{table}

%% file: sections/3_benchmark_design.tex
\section{Benchmark Design}
\label{sec:benchmark}

\begin{wrapfigure}{r}{0.5\linewidth}
\centering
\includegraphics[width=\linewidth]{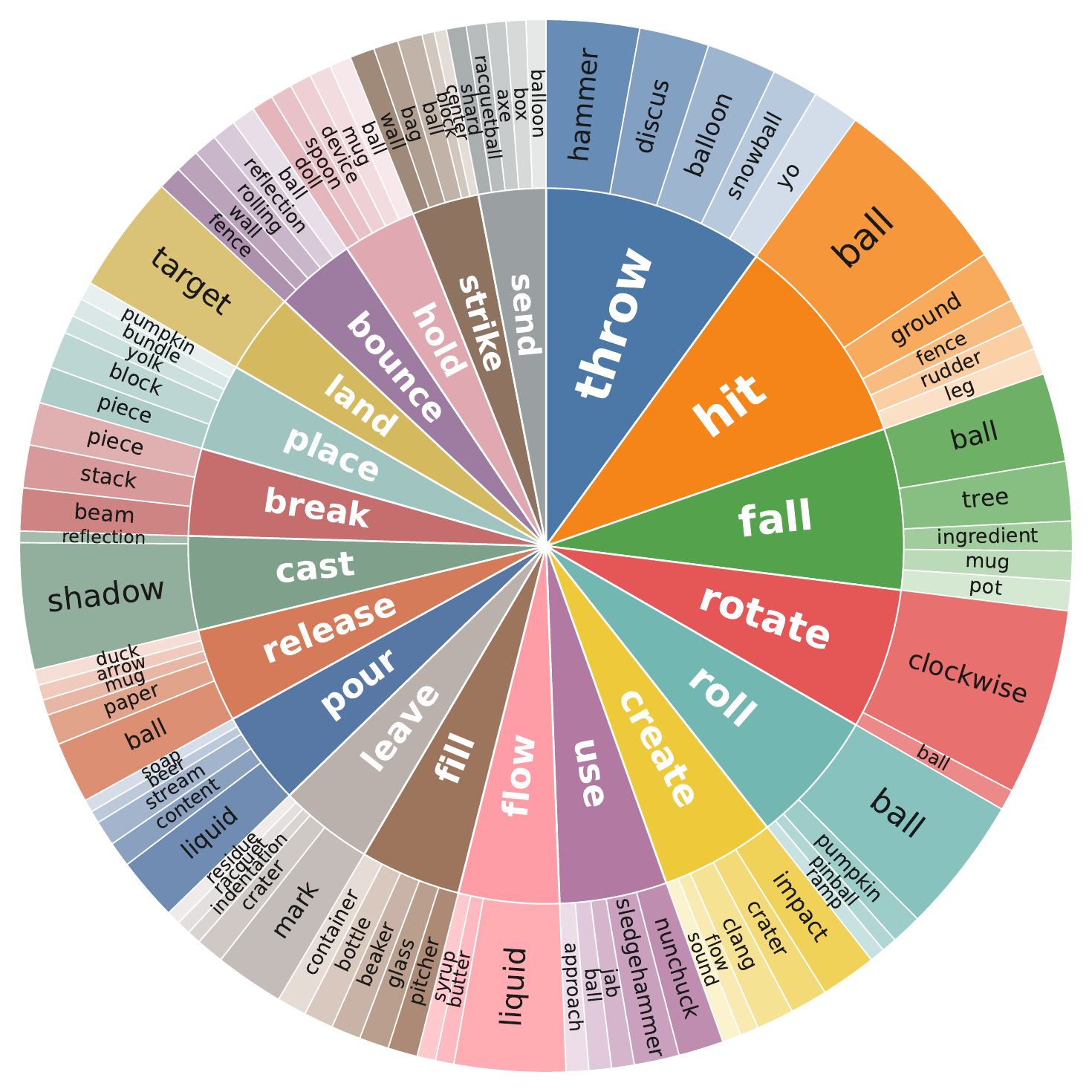}
\caption{Distribution of root verbs (inner ring, top 20 by frequency) and their direct nouns (outer ring, top 5 per verb) across the 250 PhyGround prompts.}
\label{fig:prompt_diversity}
\end{wrapfigure}

\subsection{Prompt Suite Construction}
\label{sec:prompts}

To provide a more objective, explicit evaluation of physical correctness, we curate 250 prompts from four complementary sources and retain those with unambiguous physical outcomes, visually verifiable events, and concise everyday scenarios (criteria and rationale in \Cref{sec:appendix_prompt_selection}).

\paragraph{Expected-outcome Augmentation.}
Previous benchmarks were mostly based on implicit prompts: they described only the process, without specifying the physical outcome, such as a ball falling or a ball bouncing. The original intention was to see whether video models could infer the outcome. However, many physical phenomena do not necessarily occur, which allows the video to have multiple possible endings and greatly affects the reproducibility of the evaluation. Implicit prompts can tell whether a generated video looks plausible vaguely, but they do not always reveal whether the expected physical outcome actually occurs. Therefore, we revise each implicit prompt by adding an explicit expected physical outcome. For instance, ``A ball rolls across a table'' becomes ``A ball rolls across a table and \emph{reaches the edge, falls off}.''
This forces the model to attempt the physical phenomenon, raising the bar on physical correctness and giving evaluators a clear scoring reference. All prompts use a text+image-to-video (ti2v) format with the first frame as image condition.

\paragraph{Curating Procedure.}
First frames are drawn from existing video corpora, in particular, WorldModelBench and Physics-IQ provide reference frames directly, OpenVid and VideoPhy-2 first frames are extracted as the first frame of their source video.
Because some source videos can contain content that conflicts with our grading criteria, we manually verify that every first-frame image and text condition is of good quality before we include it in the prompt suite.

\paragraph{Prompt Diversity.}

 \Cref{fig:prompt_diversity} illustrates the distribution of root verbs and direct nouns across the 250 prompts. The 250 prompts cover a broad and well-balanced range of physical actions and entities: no single verb dominates, and each verb fans out into multiple distinct objects spanning rigid-body motion (throw/hit/roll/bounce), gravity and contact (fall/land/place/break), fluids (pour/flow/fill), optics (cast/shadow/reflection), and tool use (use, hold, strike).  

\subsection{Physics Criteria Taxonomy}
\label{sec:taxonomy}

We define a curated set of thirteen physical laws organized across three domains.
In particular, instead of treating ``fluid dynamics'' as a single holistic category, PhyGround separately evaluates flow dynamics, boundary interaction, displacement, and continuity, allowing different physical constraints within the same domain to be diagnosed independently.
Formally, let $\mathcal{L} = \{l_1, \ldots, l_{13}\}$ denote the law set and $\mathcal{D} = \{\text{solid body}, \text{fluid}, \text{optical}\}$ the domain set.
Each video is evaluated with 2 to 4 applicable laws from $\mathcal{L}$ based on its prompt content.
The selection criteria for laws and the excluded categories are detailed in \Cref{sec:appendix_law_selection}.

Following social science research design principles on validity and measurement quality \citep{reis2000handbook}, we use a 1-5 scale to preserve differences in violation severity, where 1 indicates that fully implausible and 5 indicates fully plausible. This approach avoids the limitation of a binary measure, which would collapse all deviations into the same category and obscure the distinction between minor imperfections and severe physical failures.

\subsection{Automatic VLM Judge}

PhyGround provides an automatic VLM judge for reproducible large-scale evaluation. The judge takes the video frames, the augmented prompt, and a criterion-specific rubric as input, and outputs a 1-5 score for one general dimension or one applicable physical  law. For physical-law scoring, the same per-law sub-questions are included in the judge prompt as an observation checklist.  In practice, we query the judge separately for each criterion, each video requires 5--7 VLM queries in total: three for the general dimensions and 2-4 for the applicable physical laws. This criterion-wise querying keeps each prompt focused on a single rubric, reduces interference among heterogeneous criteria, and produces more reliable, easily parsed diagnostic scores.
  
\paragraph{Sub-question Decomposition.}
For the VLM judge, we equip each law with 2-3 sub-questions (SubQ), each targeting a specific observable property.
These sub-questions operationalize an abstract law into visually verifiable observation items and are explicitly listed in the VLM judge's prompt as a reasoning checklist that constrains its observation perspective and reduces subjectivity.

\subsection{Scoring}
\label{sec:scoring}

\paragraph{General Physical Adherence.}
We evaluate each video along three complementary dimensions. \textbf{Semantic Alignment (SA)} measures the extent to which the video content matches the text prompt. \textbf{Physical Temporal Validity (PTV)} assesses whether the temporal sequence of physical events follows a plausible physical order. \textbf{Object Persistence (Persistence)} evaluates whether objects maintain consistent appearance, shape, and existence throughout the video.

\paragraph{Per-law Physics Scores.}

For each video, we evaluate only the subset of physical laws applicable to its prompt, resulting in 2-4 law-level scores per video. Each law is scored independently, with the score reflecting the degree to which the generated video follows that law.

\paragraph{Human Annotation.} We design the annotation process using established principles for valid and reliable human measurement \citep{campbell1963experimental,reis2000handbook,shadish2002experimental}. Annotators complete the task on desktop devices only to ensure consistent viewing conditions and reduce variation in screen size, resolution, and interface layout. To limit fatigue and low-effort responding \citep{krosnick1991response}, each annotator rates roughly 12.6 videos on average. Participants first review the consent form, study overview, voluntary-participation statement, and task instructions. The task is described without revealing our hypotheses, and model identities are hidden to reduce demand effects \citep{orne1962social}. After consenting, participants report basic demographic information and complete a short training module that illustrates different levels of physical realism and explains the five-point Likert scale. During evaluation on our annotation platform, each participant is randomly assigned videos from the full pool, and video order is randomized independently to reduce assignment bias, ordering effects, and carryover effects. Each video must be watched at least once before rating. Annotators score three general dimensions and 2-4 physical laws, each on a 1-5 Likert scale. Finally, we apply a two-round quality-control pipeline based on score consistency, dimension discriminability, peer agreement, and behavioral engagement.

\subsection{Judge Training}
\label{sec:judge_training}

We perform supervised fine-tuning with LoRA adapters, with training data drawn from the human annotation platform (\Cref{sec:experiments}).
Each sample is a triplet $(v, q_k, y_k)$: video frames $v$, the augmented prompt $q_k$ instantiated for a single criterion $k$ (containing the expected outcome), and the structured JSON score label $y_k = \{k: s\}$. Here, $s \in \{1, \dots, 5\}$ denotes the score obtained by aggregating human annotations.
Let the dataset be $\mathcal{D}$; only the LoRA parameters $\theta$ are trainable, and the training objective minimizes the token-level negative log-likelihood:
\begin{equation}
\mathcal{L}(\theta) = -\mathbb{E}_{(v, q_k, y_k) \sim \mathcal{D}}\left[\sum_{t=1}^{|y_k|} \log p_\theta\bigl(y_{k,t} \mid y_{k,<t}, v, q_k\bigr)\right].
\end{equation}
Here $y_{k,t}$ denotes the $t$-th token of the serialized JSON label $y_k$, and $|y_k|$ is its token length.
Training uses a holdout validation strategy, we evaluate the model on Veo-3.1 hold-out split and train on the remaining 7 models human annotations.
The LoRA configuration, score-supervision format, and training hyperparameters in \Cref{sec:appendix_judge_training}.
We refer to the resulting fine-tuned judge as \textbf{PhyJudge-9B}.

%% file: sections/5_experiments.tex
\section{Experiments}
\label{sec:experiments}

\subsection{Setup}
\label{sec:setup}

\paragraph{Video Generation Models.}
We evaluate 8 video generation models spanning a representative range of architectures and scales, including one closed-source model (Veo-3.1~\citep{veo3}) and seven open-source models (Wan2.2-27B-A14B~\citep{wan2025}, OmniWeaving~\citep{omniweaving2025}, Cosmos-Predict2.5-14B and Cosmos-Predict2.5-2B~\citep{cosmos2025}, Wan2.2-TI2V-5B~\citep{wan2025}, LTX-2.3-22B and LTX-2-19B~\citep{hacohen2026ltx2}).
All models are evaluated under the unified ti2v format with 250 prompts, producing 2{,}000 total videos.
Unless otherwise stated, each model is evaluated under its native default generation settings, including output resolution and frame rate, without post-hoc normalization (see \Cref{sec:appendix_videogen} for details).

\paragraph{Evaluation Design.}
In total, we recruited 459 annotators, who in total contributed 5{,}796 complete annotations and 37.4K individual labels through our evaluation platform, with all 2{,}000 videos. Each video receives at least two annotators (1{,}410 with at least three). To the best of our knowledge, this is the largest human-annotator pool reported for a physics-focused video generation benchmark (\Cref{tab:benchmark_comparison}).
For each law, the rubric anchor shown to annotators is the per-law main question together with its accompanying note (e.g., for gravity: ``unsupported objects fall downward, projectiles follow curved trajectories, poured liquids descend''), keeping the annotation interface concise.
After two rounds of quality control, cross-validating four mutually independent signals (score constancy, per-video dimension distinctness measured by copy-paste rate, peer MAE, and behavioral signals. Full procedure in \Cref{sec:appendix_humaneval}) 352 annotators are retained, yielding 4{,}576 complete annotations and 29.5K individual labels, with 1{,}831 of the 2{,}000 videos still covered by at least two annotators.

For VLM evaluation, we include Claude (Opus 4.7), Gemini-3.1-Pro\cite{google2026gemini31pro}, Qwen3.5 9B-base, and Qwen3.5 27B-base \cite{qwen3.5}, with both direct-answer and Chain-of-Thought prompting variants where applicable, as baseline models in our setup.

\subsection{Main Results}
\label{sec:results}

\input{tables/TABLE_leaderboard.tex}

\Cref{tab:leaderboard} shows the scores for all models. Each general dimension (SA, PTV, Persist.) is averaged per video across annotators and then across videos. Each physics-domain score (Solid-Body, Fluid, Optical) is the mean over all (video, law) annotation units in that domain, so each law is weighted by the number of videos it applies to. Overall averages the General and Physics scores: $0.5 \times \text{mean(SA, PTV, Persist.)} + 0.5 \times $ the (video, law) mean pooled across the three physics domains.
At a high level, no model achieves strong physics adherence: no model exceeds 3.3/5 (66\%) on Overall, and no model crosses this threshold on the solid-body domain either, indicating substantial room for improvement in physical reasoning.
However, reading only Overall provides an incomplete picture, motivating the cross-model analysis along the Solid-Body / Fluid / Optical columns presented below.

\paragraph{Per-domain decomposition reveals complementary model profiles.}
Holistic rankings can obscure substantial domain-level differences. For example, the top open-source Wan2.2-27B-A14B and the closed-source Veo-3.1 obtain the same Overall score (both 3.28), yet their strengths differ sharply: Wan2.2-27B-A14B matches or exceeds Veo-3.1 on 4 of the 7 evaluation dimensions, achieving the best scores in the table on solid-body (3.23 vs.\ 3.12), temporal coherence (PTV, 3.37), and persistence (3.50), whereas Veo-3.1 leads decisively on fluid (3.65 vs.\ 3.18, a gap of 0.47) and optical (3.69 vs.\ 3.55). A similar effect appears near the bottom: Cosmos-Predict2.5-2B and LTX-2-19B differ by only 0.02 on Overall (2.58 vs.\ 2.56), but the former is weak on fluid (2.77) and strong on optical (3.35), while the latter is competitive on fluid (3.04), has the lowest solid-body score in the table (2.46), and is weak on optical (2.78). Thus, models with nearly identical holistic scores may imply very different choices for deployment, model selection, and targeted improvement.

This decomposition also exposes cases where a low holistic score hides a functioning sub-capability. Although Cosmos-Predict2.5-2B ranks only 7th on Overall, its optical score reaches 3.35, essentially tied with Cosmos-Predict2.5-14B (3.36) and exceeding LTX-2.3-22B, Wan2.2-TI2V-5B, and LTX-2-19B (all $<3.0$). Its within-model range, from solid-body 2.53 to optical 3.35, is the largest among the 8 evaluated models and crosses nearly a full rating level, showing that its Overall score is pulled down by uneven domain performance rather than uniformly weak behavior across all domains.

Finally, aggregating scores by physics domain rather than by model reveals a benchmark-level difficulty profile: the cross-model means are 2.79 on solid-body, 3.08 on fluid, and 3.24 on optical. This suggests that scenarios involving rigid-body motion, force transmission, and collision response remain particularly challenging. 

The value of the per-domain decomposition is not that it correlates better with human ratings than holistic scoring, but that \textbf{it extracts an information dimension that holistic scoring is structurally unable to contain}. \Cref{sec:appendix_scoreperlaw} further drills down to per-law granularity, revealing finer-grained observations that are averaged out even by the three-domain aggregation.

\subsection{Annotation Reliability}
\label{sec:reliability}

\paragraph{Leaderboard Stability.}
We assess ranking robustness by measuring score shifts under annotator subsampling.
Dropping a random $30\%$ of annotators changes per-model scores by only $0.033$ RMS, while sub-sampling down to $20$ annotators inflates this to $0.204$ RMS, a sixfold gap that highlights the advantage of using a larger candidate pool (full sweep in \Cref{sec:appendix_humaneval}).
In a split-half check, we partition the 352 annotators into two equal groups and compare the resulting model rankings over 100 random partitions. The halves agree strongly, with Spearman $\rho = 0.93 / 0.91 / 0.93$ and Kendall $\tau$-b $= 0.83 / 0.80 / 0.83$ on General/Physics/Overall (\Cref{tab:robustness_split_half}).
Thus, while individual annotators may disagree on specific videos, the aggregate benchmark reliably distinguishes model quality.

\paragraph{Inter-Annotator Agreement.}
We compute same-prompt pairwise preference agreement using the same win/loss-style metric as WorldModelBench \cite{li2025worldmodelbench} to summarize inter-annotator reliability. Physical (avg) averages each annotator's per-law scores into a single physics score per video before comparing, and Total score is General (avg) $+$ Physical (avg). \Cref{tab:pairwise_preference_agreement} reports this agreement on the General/Physical/Total scores. Agreement is similar for the General and Physical scores (76.0\% vs.\ 74.5\%), while the Total score obtains 73.0\%. These values are comparable to those reported by WorldModelBench (70\%) and VideoPhy-2 (75-80\%).

\input{tables/humaneval_tables}

\subsection{Judge Results}
\label{sec:judge_results}

\begin{figure}[t]
  \centering
  \includegraphics[width=\linewidth]{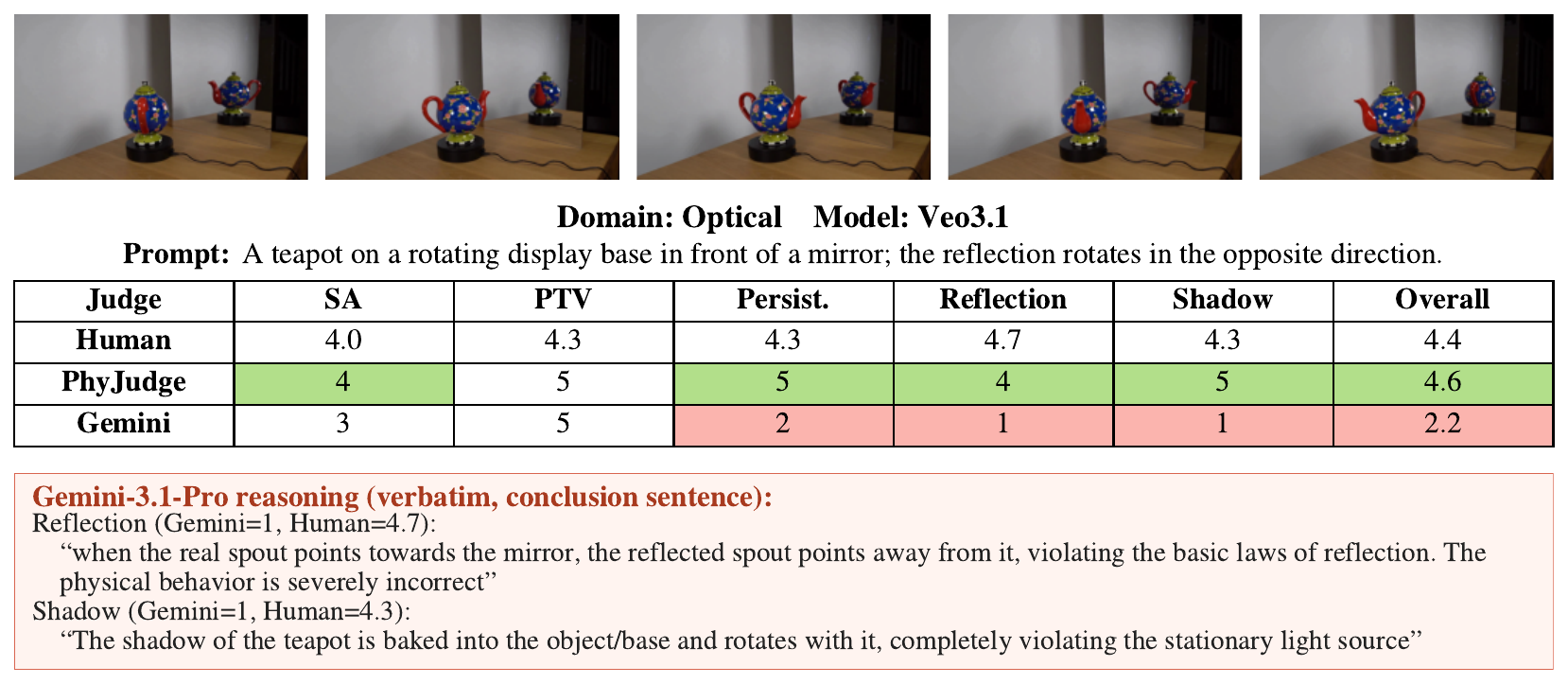}
  \caption{Head-to-head comparison of PhyJudge-9B and Gemini-3.1-Pro on a representative Optical case (Veo-3.1, rotating-mirror teapot). Green cells mark dimensions where PhyJudge-9B tracks the human mean more closely than Gemini; red cells mark dimensions where Gemini diverges from the human mean by at least two rating levels. The panel beneath the score table reproduces Gemini's verbatim conclusion sentence on each disputed law.}
  \label{fig:judge-cmp}
\end{figure}

\input{tables/TABLE_pred_error_subtable.tex}
\input{tables/gemini_per_law.tex}

The main judge results are reported in \Cref{tab:pred_error_subq_subtable} (judge per-dimension aggregate relative bias). PhyJudge-9B's aggregate relative bias is lower than that of the closed-source Gemini-3.1-Pro and Claude (Opus 4.7).
Our evaluation protocol is strictly held-out: all PhyJudge-9B tests are conducted on Veo-3.1 videos, and the Veo-3.1 videos and their human ratings do not appear in the training set of PhyJudge-9B.  All judges are evaluated with the same set of per-law sub-questions. Rows marked ``CoT'' use a reasoning-augmented prompt: the judge first answers each sub-question with brief evidence and a written justification before emitting the final integer score, whereas the non-CoT rows constrain the same model to a score-only output.
\Cref{tab:gemini_per_law} gives the raw-score view behind this design choice: adding sub-questions moves Gemini-3.1-Pro much closer to human means on the motion-heavy Fluid and Optical domains ($1.86\to 3.29$ and $2.64\to 3.28$), while adding CoT on top of SubQ is not consistently beneficial.

\paragraph{A representative head-to-head case.}
\Cref{fig:judge-cmp} drills into a single Optical case---a teapot on a rotating display in front of a mirror, generated by Veo-3.1.
Humans rate the clip 4.4 because reflection and shadow track the teapot correctly, and PhyJudge-9B closely follows with 4.6, while Gemini-3.1-Pro scores both reflection and shadow at 1 and misreads a physically plausible scene as a near-total optical violation.
Both judges are run with the same sub-question prompt schema as in the main judge results: Gemini-3.1-Pro with chain-of-thought (CoT) reasoning on top of the sub-questions, and PhyJudge-9B with the score-only sub-question prompt that matches its training format.
Examples for all other laws are shown at \Cref{sec:appendix_judge_examples}.

Beyond accuracy, PhyJudge-9B is released with fixed weights and the full scoring pipeline and does not depend on provider-side model versions or API behavior, so it can be independently reproduced and audited.
On frame sampling, the most common implementation choice, we sweep 1/2/4/8\,fps for PhyJudge-9B as well as for the Qwen3.5-9B-base and Qwen3.5-27B-base off-the-shelf judges, and find that within each judge the within-fps spread is small and exhibits no systematic directional bias, whereas across judges the fine-tuned PhyJudge-9B is markedly closer to human labels than either base judge---larger base scale alone does not close the gap (\Cref{tab:fps_pred_error}); all main results use 4\,fps.

%% file: tables/TABLE_leaderboard.tex
\begin{table}[t]
\centering
\caption{Human evaluation results on PhyGround (1--5 scale, higher is better) over 250 prompts. SA, PTV, and Persist.\ denote Semantic Alignment, Physical Temporal Validity, and Persistence; Solid-Body, Fluid, and Optical denote the three physics-domain scores. Overall combines the General and Physics views with equal weight. \textbf{Bold} marks the best score in each column and \underline{underline} marks the second-best. $^\dagger$: closed-source.}
\label{tab:leaderboard}
\resizebox{\textwidth}{!}{%
\begin{tabular}{l ccc ccc c}
\toprule
& \multicolumn{3}{c}{General Physics $\uparrow$} & \multicolumn{3}{c}{Domain Physics $\uparrow$} & \\
\cmidrule(lr){2-4} \cmidrule(lr){5-7}
Model & SA & PTV & Persist. & Solid-Body & Fluid & Optical & Overall $\uparrow$ \\
\midrule
    Wan2.2-27B-A14B & \underline{3.10} & \textbf{3.37} & \textbf{3.50} & \textbf{3.23} & 3.18 & \underline{3.55} & \textbf{3.28} \\
    Veo-3.1$^\dagger$ & \textbf{3.26} & \underline{3.29} & \underline{3.42} & \underline{3.12} & \textbf{3.65} & \textbf{3.69} & \textbf{3.28} \\
    OmniWeaving & 2.97 & 3.17 & 3.34 & 2.98 & \underline{3.26} & 3.22 & 3.10 \\
    Cosmos-14B & 2.66 & 2.98 & 3.20 & 2.81 & 2.98 & 3.38 & 2.91 \\
    LTX-2.3-22B & 2.58 & 2.74 & 2.72 & 2.62 & 3.07 & 2.98 & 2.69 \\
    Wan2.2-TI2V-5B & 2.44 & 2.68 & 2.78 & 2.58 & 2.71 & 2.99 & 2.63 \\
    Cosmos-2B & 2.33 & 2.56 & 2.77 & 2.53 & 2.77 & 3.35 & 2.58 \\
    LTX-2-19B & 2.56 & 2.64 & 2.48 & 2.46 & 3.04 & 2.78 & 2.56 \\
\bottomrule
\end{tabular}%
}
\end{table}

%% file: tables/humaneval_tables.tex
\begin{table}[t]
\centering
\begin{minipage}[t]{0.46\textwidth}
\centering
\caption{Split-half ranking reliability of the leaderboard. The 352-annotator post-QC pool is randomly partitioned into two equal halves; each cell reports mean $\pm$ standard deviation over 100 random partitions.}
\label{tab:robustness_split_half}
\setlength{\tabcolsep}{6pt}
\renewcommand{\arraystretch}{1.15}
\begin{tabular}{@{}lcc@{}}
\toprule
Score & Spearman's $\rho$ & Kendall $\tau$-b \\
\midrule
General & 0.930 $\pm$ 0.045 & 0.828 $\pm$ 0.071 \\
Physics & 0.907 $\pm$ 0.058 & 0.801 $\pm$ 0.092 \\
Overall & 0.928 $\pm$ 0.051 & 0.830 $\pm$ 0.088 \\
\bottomrule
\end{tabular}
\end{minipage}\hfill
\begin{minipage}[t]{0.50\textwidth}
\centering
\caption{Pairwise preference agreement on the Total score, with WorldModelBench and VideoPhy-2 shown as reference points. Ties are excluded; $N$ is the number of duplicate-vote comparisons after excluding ties.}
\label{tab:pairwise_preference_agreement}
\setlength{\tabcolsep}{6pt}
\renewcommand{\arraystretch}{1.15}
\begin{tabular}{@{}lcr@{}}
\toprule
Benchmark & \makecell{Pairwise\\agreement (\%) $\uparrow$} & $N$ \\
\midrule
Total score & 73.0 & 58,984 \\
\midrule
\textbf{WorldModelBench} & 70.0 & --- \\
\textbf{VideoPhy-2} & 75--80 & --- \\
\bottomrule
\end{tabular}
\end{minipage}
\end{table}

%% file: tables/TABLE_pred_error_subtable.tex
\begin{table}[t]
\centering
\caption{Per-dimension aggregate relative bias (\%) using prompts that provide sub-questions for each physical law; non-CoT rows are score-only, while CoT rows use reasoning-augmented outputs (the judge first answers each sub-question with brief evidence and a written justification before emitting the final integer score). The first three columns report absolute grouped mean-bias, $E = |\bar J - \bar H| / \bar H$, computed per dimension/law and then macro-averaged. Physics is macro-averaged over domains. Overall $=$ 0.5$\cdot$Macro Avg (General) $+$ 0.5$\cdot$Physics. Lower is better; \textbf{bold} marks the column-wise minimum. The last column reports the WorldModelBench~\cite{li2025worldmodelbench} signed relative bias on the per-video overall score, $(\bar J - \bar H) / \bar H$, where positive values indicate judge overestimation and negative values indicate underestimation. For this signed column, values closer to zero are better; \textbf{bold} marks the row with the smallest $|\cdot|$.}
\label{tab:pred_error_subq_subtable}
\resizebox{\textwidth}{!}{%
\begin{tabular}{l c c c c}
\toprule
Model & \textbf{General (\%) $\downarrow$} & \textbf{Physics (\%) $\downarrow$} & \textbf{Overall (\%) $\downarrow$} & \textbf{Signed (\%) $\to 0$} \\
\midrule
    Claude (Opus 4.7) & 16.2 & 10.0 & 13.1 & +14.5 \\
    Claude (Opus 4.7, CoT) & 25.2 & 34.4 & 29.8 & +34.3 \\
    Gemini-3.1-Pro & 17.3 & 15.9 & 16.6 & -6.3 \\
    Gemini-3.1-Pro (CoT) & 17.2 & 17.4 & 17.3 & -6.5 \\
    Qwen9B-base & 28.2 & 35.0 & 31.6 & +37.7 \\
    Qwen9B-base (CoT) & 41.8 & 41.6 & 41.7 & +47.8 \\
    Qwen27B-base & 21.3 & 40.9 & 31.1 & +35.7 \\
    Qwen27B-base (CoT) & 37.3 & 46.9 & 42.1 & +48.3 \\
    PhyJudge-9B & \textbf{1.0} & \textbf{5.5} & \textbf{3.3} & \textbf{-1.1} \\
\bottomrule
\end{tabular}%
}
\end{table}

%% file: tables/gemini_per_law.tex
\begin{table}[t]
\centering
\caption{Raw Gemini-3.1-Pro judge means under prompt-schema ablations on the human-eval set. Each cell reports $\bar J$ with absolute relative bias $|\bar J - \bar H|/\bar H$ in parentheses, pooled over law-video pairs in the domain. Adding SubQ closes large gaps to human means on Fluid and Optical; CoT on top of SubQ is not consistently beneficial. \textbf{Bold} marks the schema closest to $\bar H$ in each row.}
\label{tab:gemini_per_law}
\resizebox{\textwidth}{!}{%
\begin{tabular}{l c c c c c}
\toprule
Domain & \textbf{$\bar H$} & \textbf{Gemini-3.1-Pro}\,$\downarrow$ & \textbf{+ CoT}\,$\downarrow$ & \textbf{+ SubQ}\,$\downarrow$ & \textbf{+ SubQ + CoT}\,$\downarrow$ \\
\midrule
    Solid-Body & 2.95 & 1.76 (40\%) & 2.02 (31\%) & 2.29 (22\%) & \textbf{2.65 (10\%)} \\
    Fluid & 3.57 & 1.86 (48\%) & 1.99 (44\%) & \textbf{3.29 (8\%)} & 2.84 (21\%) \\
    Optical & 3.62 & 2.64 (27\%) & 2.68 (26\%) & 3.28 (9\%) & \textbf{3.68 (2\%)} \\
\bottomrule
\end{tabular}%
}
\end{table}

%% file: sections/6_conclusion.tex
\section{Conclusion}
\label{sec:conclusion}

We introduce PhyGround, a criteria-grounded benchmark for evaluating physical reasoning in generative world models at the level of individual physical laws. It reveals that video generation models can exhibit distinct strengths and weaknesses across different physical constraints. We further provide a large-scale, quality-controlled human annotation set and an open physics-specialized automatic judge to support scalable and reproducible evaluation. Overall, this work advances our understanding of current video generation models as world models and provides a diagnostic foundation for future improvements in physical reasoning.

%% file: appendix/law_selection.tex
\section{Prompt Suite and Physics Law Selection Rationale}
\label{sec:appendix_prompt_law_selection}

This appendix provides the full design rationale behind the prompt suite (\Cref{sec:appendix_prompt_selection}) and the physics-law taxonomy (\Cref{sec:appendix_law_selection}).
The prompt selection determines \emph{whether} the generated video triggers an evaluable physical event; the law selection determines \emph{how} to score the event once triggered.

\subsection{Prompt Selection}
\label{sec:appendix_prompt_selection}

\subsubsection{Source Complementarity}

The prompt suite draws from four complementary sources rather than adopting any single existing benchmark wholesale.
The guiding principle is \emph{coverage of evaluable physics scenarios for the text+image-to-video paradigm}, not exhaustive inheritance of all physics-labeled samples from any one benchmark.

\subsubsection{Suite Composition}

\begin{table}[h]
\centering
\caption{Composition of the 250-prompt benchmark suite by source.}
\label{tab:suite_composition}
\small
\begin{tabular}{lc}
\toprule
Source & Final suite \\
\midrule
VideoPhy-2 & 166 \\
Physics-IQ & 37 \\
OpenVid & 35 \\
WorldModelBench & 12 \\
\midrule
\textbf{Total} & \textbf{250} \\
\bottomrule
\end{tabular}
\end{table}

\subsubsection{Source Licenses}
\label{sec:appendix_source_licenses}

All four prompt sources are released under permissive licenses that allow research use and redistribution.
VideoPhy-2 is released under the MIT License, while WorldModelBench, Physics-IQ, and OpenVid are released under the Creative Commons Attribution 4.0 (CC BY 4.0) license.
Our benchmark redistributes only the prompts and first-frame images derived from these sources, with attribution preserved.

\subsubsection{Filtering Criteria}
\label{sec:appendix_prompt_selection_filtering}

A prompt is retained only if it simultaneously satisfies all of the following:

\begin{enumerate}
    \item \textbf{Unique Physical Outcome.} The expected outcome must be unambiguous, admitting only one physically plausible result. For example, ``spaghetti breaks into pieces'' is excluded because the number of pieces has no correct answer.

    \item \textbf{Visual Verifiability.} Whether the outcome occurred must be immediately apparent from the generated video, without domain expertise or frame-by-frame analysis.

    \item \textbf{Concise Everyday Scenario.} The prompt must describe a concrete situation in under 50 words, set in a familiar context. Prompts requiring specialized laboratory equipment or rare objects (e.g., ``rotary filling machine,'' ``press brake'') are excluded because video models lack sufficient visual reference for these objects.

    \item \textbf{Domain Alignment.} The core phenomenon must genuinely belong to the target physics domain. Prompts that are only superficially associated through keywords are excluded. For example, grinding should not be classified as collision when the underlying mechanism is friction, nor should a penguin entering water be classified as fluid dynamics when the scene primarily depicts animal behavior.

    \item \textbf{Physics as Core Phenomenon.} The scenario must be driven by a natural physical process. Machine and tool operations (hydraulic lifts, excavators, mechanical presses) are excluded because the physical outcome is incidental to the mechanical action. Active human or animal motion (jumping, throwing, running) is excluded because the initial conditions (applied force, launch angle) are unobservable, making the space of correct outcomes unbounded. For example, ``The athlete throws a javelin'' does not specify whether it should travel 30 or 60 meters. This contrasts with passive processes (free fall, fluid flow, collision rebound), where the outcome is fully determined by the observable initial state.
\end{enumerate}

\subsubsection{Expected-Outcome Augmentation}

Many original prompts describe only the initial setup without stating the expected physical result.
For instance, ``The grabber tools let go of the ball and block'' does not mention that the objects should fall.
Such prompts allow the video generation model to avoid the physical event entirely, rendering evaluation vacuous.

We use Gemini 2.5 Flash to augment each implicit prompt by appending a concise description of the expected outcome (typically 10--20 words), without altering the scene semantics; each augmented prompt is then manually checked for consistency with the first frame and filtering criteria.
Examples:
\begin{itemize}
    \item ``A ball rolls across a table'' $\to$ ``...~\emph{reaches the edge, falls off, and bounces on the floor below.}''
    \item ``Water is poured into a cup on a baking tray'' $\to$ ``...~\emph{until the cup overflows, with water spilling onto the baking tray surface.}''
    \item ``The wooden stick hits the first block as it rotates'' $\to$ ``...~\emph{The blocks topple over one by one in a domino chain reaction.}''
\end{itemize}

For OpenVid, whose source data consists of lengthy video captions (often over 100 words of narrative description), captions are rewritten into concise physics-focused prompts (under 50 words) with the physical event made explicit, grounded against the first frame of the source video to ensure visual consistency.

This augmentation makes the implicit physical expectation explicit, ensuring that the model must attempt a scorable physical event and giving the evaluator an unambiguous reference for judgment.
Prompts that already specify the physical outcome are left unchanged.

All prompts use a unified text+image-to-video (ti2v) format, where the first frame serves as an image condition.
This ensures fair cross-model comparison by eliminating variability from different prompt-processing pipelines.

\subsection{Physics Law Selection}
\label{sec:appendix_law_selection}

We curate the 13-law taxonomy from the union of physics indicators across PhyGenBench~\citep{phygenbench2024}, Physics-IQ~\citep{physicsiq2026}, VideoPhy-2~\citep{videphy2_2025}, VideoScience-Bench~\citep{videosciencebench2025}, and WorldModelBench~\citep{li2025worldmodelbench}.
These benchmarks organize physics along different axes: PhyGenBench uses four broad domains, Physics-IQ and VideoPhy group by scenario type, VideoScience-Bench uses a three-level university-course hierarchy, and WorldModelBench lists specific physical laws.
We apply three selection criteria to consolidate them into a unified vocabulary.

\subsubsection{Selection Criteria}

\paragraph{Visual Observability.}
The physical phenomenon must be immediately verifiable from a short generated video.
This criterion excludes quantities that are not directly visible, such as temperature, energy, and electromagnetic fields. It also excludes phenomena that, while observable in principle, cannot be reliably captured within the short durations typical of generated videos. For instance, heat conduction is invisible in video frames, whereas melting and boiling, though visually apparent, rarely unfold within such brief time spans.

\paragraph{VLM Discriminability.}
Each law must be convertible into concrete visual questions that a VLM judge can reliably answer (e.g., ``Do unsupported objects fall downward?'' rather than ``Does the video obey the law of gravity?'').

\paragraph{Paradigm Compatibility.}
The evaluation must work without a ground-truth reference video (ti2v paradigm).
This excludes Physics-IQ's pixel-level metrics (Spatial IoU, Spatiotemporal IoU, Weighted Spatial IoU, MSE), which assume a unique ground-truth continuation and penalize physically plausible but visually different generations.

\subsubsection{Included Laws}

Across the 250 prompt suite, the per-domain prompt-law annotation counts are 693 for Solid Body, 136 for Fluid, and 47 for Optical.
The full per-law breakdown is shown in \Cref{tab:law_counts}, which lists the number of (prompt, law) pairs per law; this is the population used by every per-law and per-domain mean reported in the paper.

\input{appendix/tables/law_counts}

\begin{table}[t]
\centering
\caption{Evaluation criteria taxonomy: 3 general physical adherence criteria applied to every video, plus 13 physics laws organized by domain. The third column shows the question presented to human annotators for each criterion.}
\label{tab:taxonomy}
\resizebox{\linewidth}{!}{%
\begin{tabular}{llp{8.2cm}}
\toprule
Domain & Criterion & Human Rating Question \\
\midrule
\multirow{3}{*}{General} & Semantic Alignment (SA) & To what extent does the video content align with the text prompt? \\
 & Physical Temporal Validity (PTV) & To what extent does the temporal sequence of physical events follow a physically plausible order? \\
 & Object Persistence & To what extent do objects maintain consistent appearance, shape, and existence throughout the video? \\
\midrule
\multirow{6}{*}{Solid-Body} & Gravity & To what extent do objects and liquids move in accordance with gravity? \\
 & Inertia & To what extent do objects maintain their state of motion---remaining stationary or continuing to move---in the absence of a plausible external force? \\
 & Momentum & To what extent are post-collision directions of motion physically plausible? \\
 & Impenetrability & To what extent do solid objects maintain impenetrability, avoiding passage through one another? \\
 & Collision & To what extent do objects exhibit physically plausible responses to impact, proportional to the applied force? \\
 & Material & To what extent do objects respond in ways consistent with their apparent material properties? \\
\midrule
\multirow{5}{*}{Fluid} & Buoyancy & To what extent do objects sink or float in a manner consistent with their apparent density? \\
 & Displacement & When volume is added or an object is submerged, to what extent does the liquid level rise in a physically plausible manner? \\
 & Flow dynamics & To what extent does liquid flow along surfaces, spread, and drain in a physically plausible manner? \\
 & Boundary interaction & To what extent does liquid splash, rebound, or split in a physically plausible manner when hitting a boundary? \\
 & Continuity & Does the liquid avoid disappearing or appearing out of nowhere? \\
\midrule
\multirow{2}{*}{Optical} & Reflection & To what extent do reflective surfaces display content that is physically plausible given the surrounding scene? \\
 & Shadow & To what extent are shadows physically plausible in their direction, placement, and movement relative to light sources and objects? \\
\bottomrule
\end{tabular}
}
\end{table}

\paragraph{Solid-body Mechanics (6 laws).}
Gravity, inertia, momentum conservation, impenetrability, collision response, and material properties.
These represent the most fundamental everyday physical commonsense and produce the clearest visual evaluation criteria.
Solid-body mechanics laws form the main body of the benchmark because nearly every video generation scenario involves rigid objects.

\paragraph{Fluid (5 laws).}
Buoyancy (dense objects sink, lightweight objects float), displacement (liquid level rises upon submersion, containers overflow when full), flow dynamics (bulk liquid motion patterns), boundary interaction (response to obstacles---splashing, deflection), and fluid continuity (mass conservation, no spontaneous gaps or generation).
We merge buoyancy and displacement into the fluid domain because both are governed by liquid behavior and share the same physical substrate as flow dynamics; treating them as a separate top-level domain produced too few samples per video for stable domain-level aggregation.

\paragraph{Optical Physics (2 laws).}
Reflection (mirrored content matches the scene) and shadow (direction consistent with light source, synchronized with objects).
We separate shadow from reflection because their evaluation criteria differ fundamentally: reflection requires matching mirrored content, while shadow requires directional consistency.
This split, originating from Physics-IQ's optics category, has not been adopted by prior benchmarks.

\begin{figure}[t]
\centering
\includegraphics[width=\linewidth]{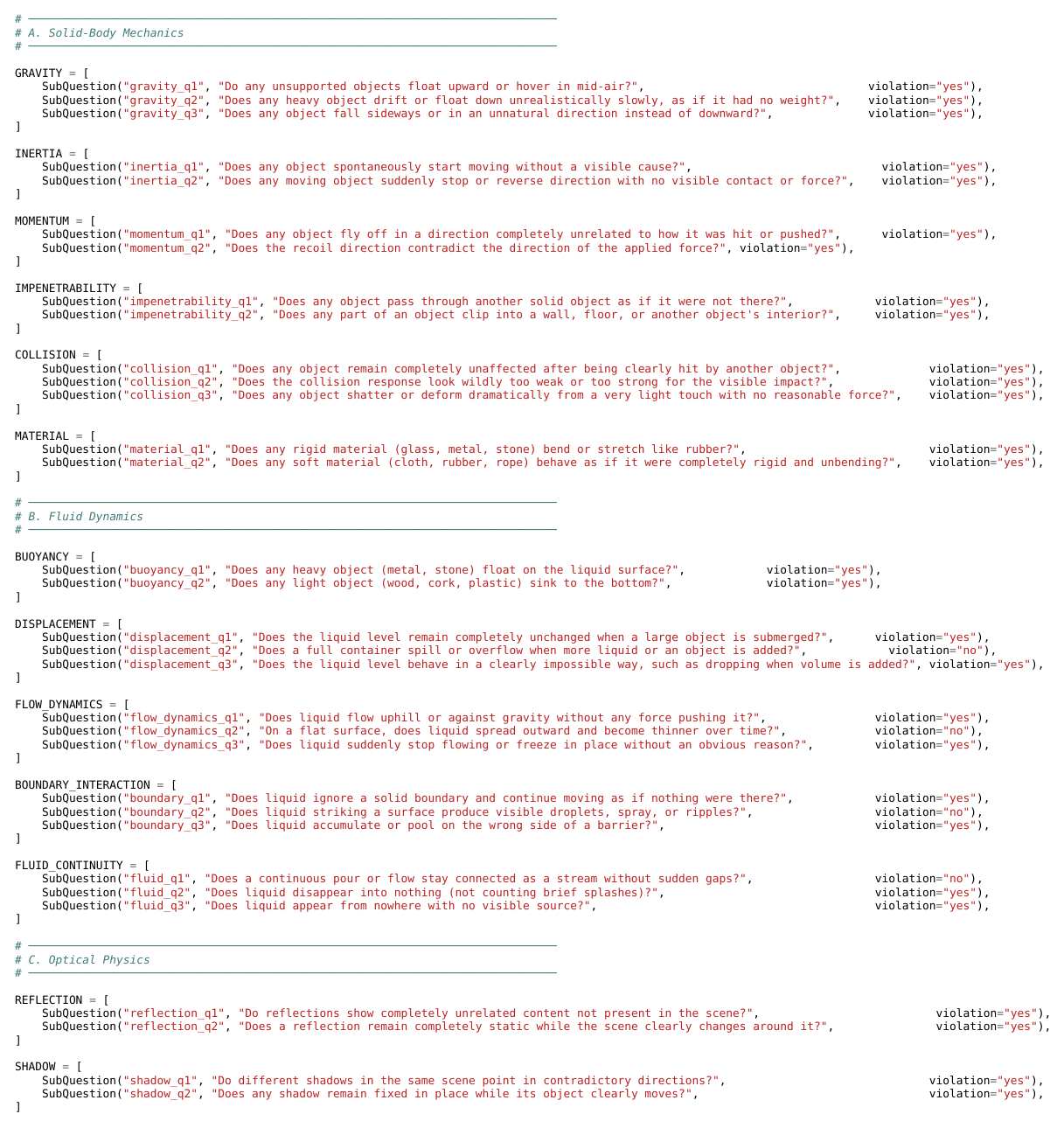}
\caption{Full sub-question inventory across the 13 laws (Solid-Body Mechanics, Fluid Dynamics, Optical Physics). Each law is decomposed into 2--3 binary observational probes phrased as violation checks; a ``yes'' answer indicates that the corresponding physical law has been violated in the video.}
\label{fig:subq_inventory}
\end{figure}

\paragraph{Full Sub-question Inventory.}
\Cref{fig:subq_inventory} lists the complete set of binary sub-questions used by the evaluator for each of the 13 laws across the three domains.
Each sub-question is phrased as a violation probe (a ``yes'' answer indicates a physical violation), giving the VLM judge a concrete mental checklist for each law. Human annotators do not see these sub-questions. Exposing the checklist to humans would increase annotation burden and risk overly mechanical scoring. In contrast, the VLM judge benefits from explicit probes because they structure its reasoning and reduce ambiguity in law-level evaluation.

\subsubsection{Excluded Categories}

\paragraph{Thermodynamics.}
Phase transitions (melting, boiling) and combustion do not reliably occur in short generated videos, yield too few evaluation samples, and have ambiguous visual boundaries (``partially melted'' vs.\ ``unchanged'').

\paragraph{Electromagnetism.}
After filtering, no prompts target electromagnetic effects.
More fundamentally, most electromagnetic phenomena produce visual effects indistinguishable from non-electromagnetic explanations---magnetic repulsion looks identical to collision rebound, making VLM verification of the underlying mechanism impossible.

\paragraph{Chemistry.}
Oxidation-reduction, acid-base reactions, and related phenomena (VideoScience-Bench) are too complex for current video models to produce and too specialized for evaluators to judge from video frames.

\paragraph{Waves, Energy, and Modern physics.}
Covered only by VideoScience-Bench at coarse granularity.
Wave phenomena require precise spatial patterns; energy is not directly observable; modern physics has no visual evaluation standard.

\paragraph{Reference-dependent and Holistic metrics.}
We exclude both pixel-level reference metrics (Physics-IQ's IoU family, which requires ground-truth video incompatible with ti2v) and holistic physics scores (VideoPhy's PC, WorldModelBench's PA, PhyGenBench's binary naturalness), which provide no diagnostic information about \emph{which} laws a model violates.
Our per-law sub-question decomposition with 1--5 scoring addresses both limitations.

%% file: appendix/tables/law_counts.tex
\begin{table}[htbp]
\centering
\caption{Per-law prompt-law annotation counts on PhyGround. Each row reports the number of (prompt, law) pairs annotated for that law; the same set of pairs is evaluated for every model in the leaderboard. The suite contains 876 (prompt, law) pairs in total across the 250 prompts.}
\label{tab:law_counts}
\begin{tabular}{l l r}
\toprule
Domain & Law & N \\
\midrule
    \multirow{6}{*}{Solid-Body} & Gravity & 123 \\
     & Inertia & 55 \\
     & Momentum & 123 \\
     & Impenetrability & 117 \\
     & Collision & 162 \\
     & Material & 113 \\
    \midrule
    \multirow{5}{*}{Fluid} & Buoyancy & 18 \\
     & Displacement & 26 \\
     & Flow dynamics & 32 \\
     & Boundary interaction & 32 \\
     & Continuity & 28 \\
    \midrule
    \multirow{2}{*}{Optical} & Reflection & 17 \\
     & Shadow & 30 \\
\bottomrule
\end{tabular}
\end{table}

%% file: appendix/humaneval.tex
\section{Human Evaluation Design and Quality Control}
\label{sec:appendix_humaneval}

This appendix describes the design and annotation quality-control pipeline, including the criteria for removing/retaining annotators, the analysis of whether difficult-to-annotate items lead to noisy scoring, and the post-cleaning inter-annotator agreement statistics.

\subsection{Annotator Recruitment and Participation Context}
\label{sec:appendix_annotator_recruitment}

The annotation study received IRB approval before data collection. Annotators were recruited from participants (also called as subject or annotator) at selective universities in North America and included both expert and non-expert annotators. The expert group consisted of 35 graduate students who had completed university-level physics coursework and could apply domain knowledge when judging physical plausibility. The non-expert group included more than 400 participants with at least undergraduate-level training, including undergraduate students, graduate students, and individuals who had already received a bachelor's degree. This mixed annotator pool is appropriate for PhyGround because the benchmark evaluates visually observable physical plausibility in everyday scenes rather than expert-only physics derivations. Annotators are not asked to compute forces, estimate physical parameters, or solve equations; instead, they judge observable outcomes, such as whether unsupported objects fall downward, whether liquids remain continuous, and whether shadows are consistent with the light source. The expert group provides domain-informed judgment, while the larger non-expert group provides broader commonsense visual assessment and supports scalable human evaluation.

We use a controlled university-based annotator pool rather than an open crowdsourcing pool. That is because the task requires consistent viewing conditions, training compliance, workload control, and careful application of structured physical criteria. Prior work shows that online labor markets such as Amazon Mechanical Turk can support behavioral research, but they also raise concerns about participant composition, attentiveness, and participant nonna{\"i}vet{\'e} in repeated experimental tasks \citep{paolacci2010running,chandler2014nonnaivete}. In our setting, a controlled annotator pool allows us to require desktop-only participation, provide a common training module, cap annotation workload, randomize video assignment and presentation order, and use hypothesis-blind task framing.

The annotation task was administered as an optional extra-credit activity. To comply with IRB requirements, participation was voluntary, and students could decline without penalty beyond not receiving the optional extra credit. Course-credit records were separated from research annotation records, and individual annotation content was not used to evaluate students' academic performance. Annotation quality was further supported through training examples, randomized assignment and ordering, workload caps, multi-signal quality control, and reliability analyses showing stable aggregate model rankings.

\subsection{Evaluation Design}
\label{sec:appendix_eval_design}

\begin{figure}[t]
\centering
\includegraphics[width=0.8\linewidth]{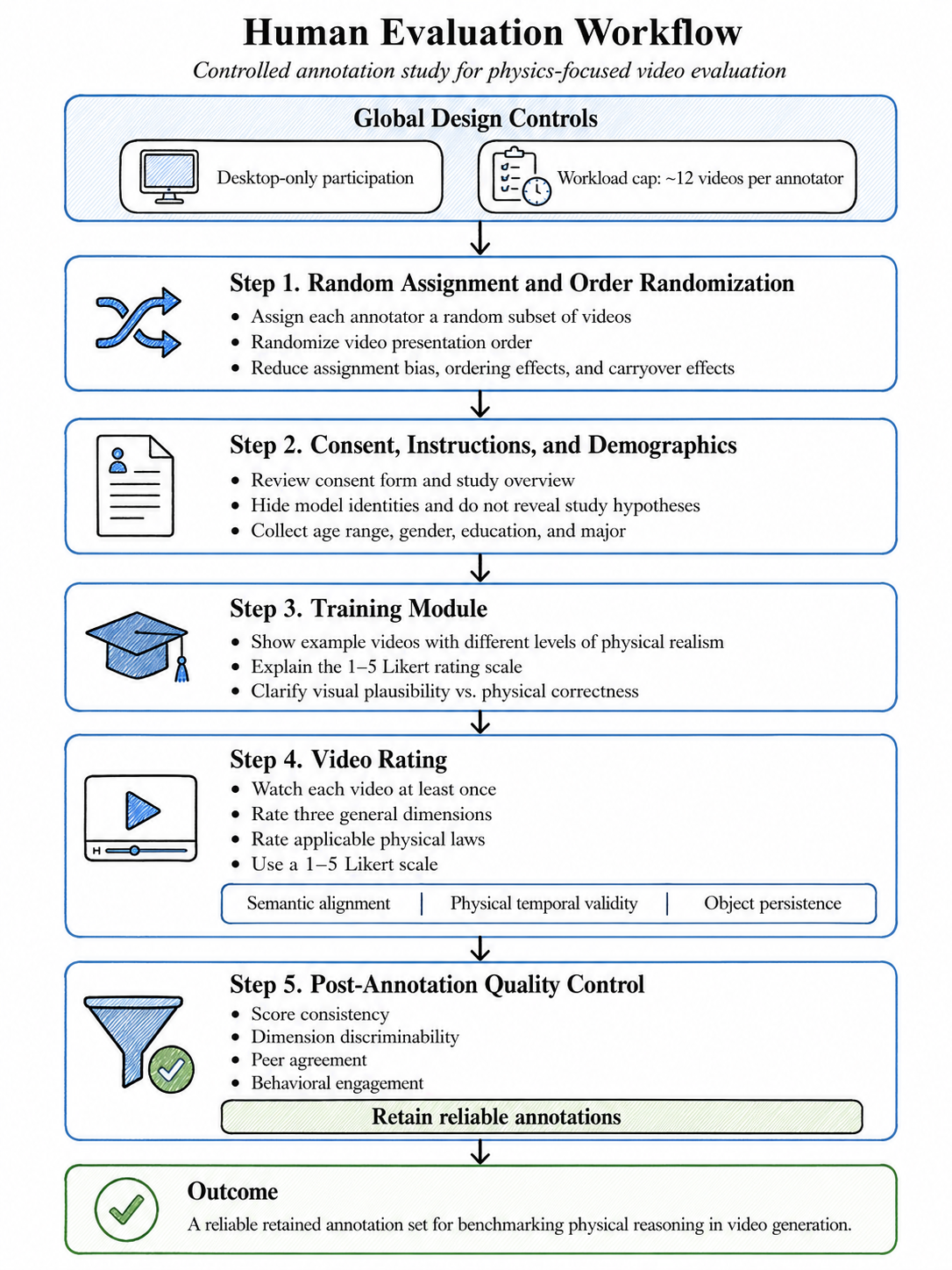}
\caption{Annotation design workflow.}
\label{fig:humanevalworkflow}
\end{figure}

\paragraph{Human Annotation.} We design the annotation process using established principles for valid and reliable human measurement \citep{campbell1963experimental,reis2000handbook,shadish2002experimental}. Annotators complete the task on desktop devices only to ensure consistent viewing conditions and reduce variation in screen size, resolution, and interface layout. To limit fatigue and low-effort responding \citep{krosnick1991response}, each annotator rates roughly 12.6 videos on average. Participants first review the consent form, study overview, voluntary-participation statement, and task instructions. The task is described without revealing our hypotheses, and model identities are hidden to reduce demand effects \citep{orne1962social}. After consenting, participants report basic demographic information and complete a short training module that illustrates different levels of physical realism and explains the five-point Likert scale. During evaluation on our annotation platform, each participant is randomly assigned videos from the full pool, and video order is randomized independently to reduce assignment bias, ordering effects, and carryover effects. Each video must be watched at least once before rating. Annotators score three general dimensions and 2--4 physical laws, each on a 1--5 Likert scale. Finally, we apply a two-round quality-control pipeline based on score consistency, dimension discriminability, peer agreement, and behavioral engagement. \Cref{fig:humanevalworkflow} summarizes the overall workflow.

We design the annotation process as a controlled human-measurement study rather than a simple labeling task. Following established research-design principles for valid and reliable measurement \citep{reis2000handbook,campbell1963experimental,shadish2002experimental}, our goal is to reduce sources of annotation noise that are unrelated to the physical quality of the generated videos. These sources include variation in viewing devices, annotator fatigue, ordering effects, response shortcuts, and demand effects.

\paragraph{Viewing Environment and Workload Control.}
To ensure consistent viewing conditions, annotators are required to complete the task on desktop devices only. This restriction reduces systematic variation in screen size, resolution, interface layout, and mobile-specific viewing behavior. Because physics-focused video evaluation requires careful inspection of motion, object behavior, and visual consistency, mobile viewing could introduce avoidable measurement noise. To reduce fatigue and low-effort responding in this cognitively demanding task \citep{krosnick1991response}, we also cap each annotator's workload at roughly 12.6 videos on average. This workload limit is intended to keep annotation sessions short enough for annotators to apply the rating criteria carefully rather than rely on shortcuts.

The experimental steps are as follows:

\begin{enumerate}
\item \textbf{Random Assignment and Presentation Order.}
When a subject enters our self-developed annotation platform, they are randomly assigned a subset of videos from the full video pool. The presentation order of videos is randomized independently for each annotator. These procedures reduce assignment bias, ordering effects, and carryover effects. Random assignment prevents particular annotators from being systematically matched with particular prompts, models, or difficulty levels. Randomized presentation order reduces the risk that earlier videos influence how later videos are judged.

\item \textbf{Consent, Task Framing, and Demographic Information.}
Participants first review the consent form, study overview, voluntary-participation statement, and other required information. The task is described without revealing our hypotheses, and model identities are hidden from annotators. This design reduces the possibility that annotators adjust their ratings based on what they believe the researchers expect, a concern commonly discussed as demand characteristics in behavioral research \citep{orne1962social}. After consenting, participants report demographic information relevant to evaluation, including age range, gender, educational background, and academic major. These variables are used only for aggregate characterization of the annotator pool and are separated from any identifying course-credit records.

\item \textbf{Training Module.}
Before entering the main evaluation task, annotators complete a short training module. The module presents example videos that illustrate different levels of physical realism and explains how to apply the five-point Likert scale. This step helps annotators understand the distinction between visual plausibility and physical correctness. It also clarifies how minor imperfections, moderate inconsistencies, and severe physical violations should be reflected in the rating scale.

\item \textbf{Rating Procedure.}
Each video must be watched at least once before rating. Annotators evaluate three general dimensions, semantic alignment, physical temporal validity, and object persistence, as well as the applicable physical laws for that video. The rating design asks annotators to assess observable visual evidence rather than solve physics equations or estimate physical parameters.

\end{enumerate}

\paragraph{Post-Annotation Quality Control.}
After annotation, we apply a two-round quality-control pipeline to remove inattentive or non-discriminating responses. The pipeline uses four complementary signals: score consistency, dimension discriminability, peer agreement, and behavioral engagement. Score consistency identifies annotators who provide nearly constant ratings across videos or dimensions. Dimension discriminability identifies annotators who repeatedly assign identical scores to distinct criteria within the same video. Peer agreement compares an annotator's ratings with other annotators who evaluated the same videos. Behavioral engagement uses platform-side signals, such as video play behavior and page engagement, to detect insufficient attention. This multi-signal design avoids relying on a single heuristic and supports a more reliable retained annotation set.

\subsection{Annotation Setup}

\begin{figure}[htbp]
    \centering
    \includegraphics[height=0.92\textheight,keepaspectratio]{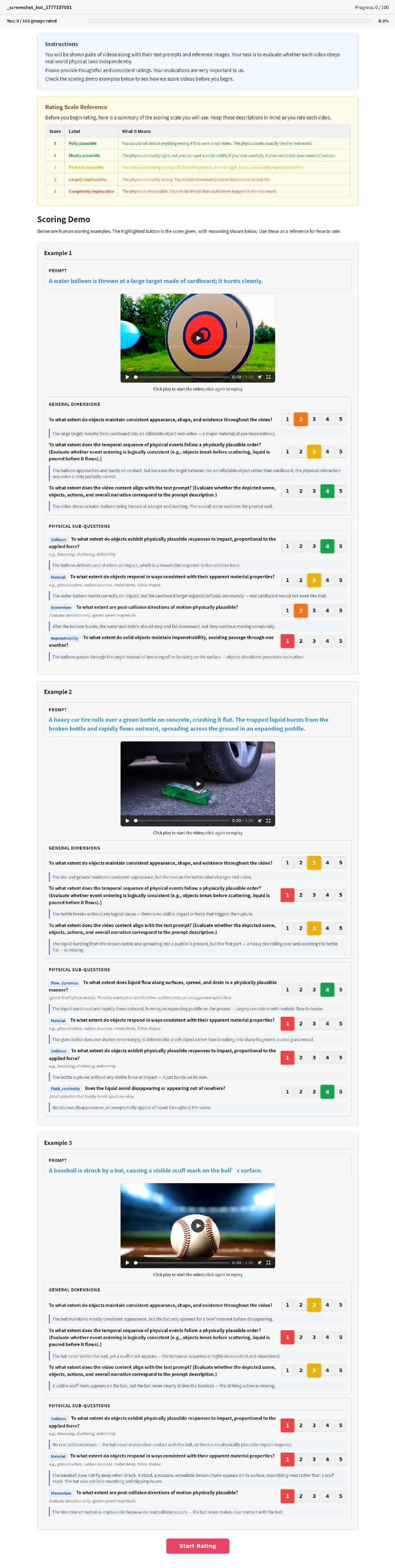}
    \caption{Screenshot for Training Session. Scoring demo shown to annotators before they begin rating. The page lists the 1-5 rating scale and three worked examples (water balloon hitting a cardboard target, tire crushing a glass bottle with liquid spill, baseball struck by a bat), each with per-dimension scores highlighted and a short justification, so annotators internalize the scoring rubric before evaluating real videos.}
    \label{fig:humaneval_demo}
\end{figure}

Annotators watched generated videos on a web-based annotation platform and assigned a score from 1-5 to each video on multiple scoring dimensions (general dimensions: SA, PTV, persistence; physical dimensions: 2-4 laws). Each video was independently annotated by multiple annotators, and the final score is the mean across annotators (the per-video aggregation rule used by the leaderboard pipeline). Before rating, every annotator went through a scoring demo (\Cref{fig:humaneval_demo}) that walks through three worked examples with the rating scale and per-dimension reasoning shown for each example.

Data scale before annotator filtering:

\begin{itemize}
    \item Annotators before filtering: 459 contributed annotations.
    \item Videos: 2{,}000 ($\geq$2 annotators: 2{,}000; $\geq$3: 1{,}410).
    \item Completed annotations: 5{,}796; individual score labels: 37{,}408.
    \item After quality control, 352 annotators contributed retained annotations.
\end{itemize}

\Cref{tab:vote_statistics} summarizes the basic vote statistics (annotation, annotator, label counts and the average annotators per video) on the unfiltered set.
\input{tables/TABLE_vote_statistics.tex}

\subsection{Quality Control Pipeline}

The raw annotation data went through two rounds of quality control. Filtering uses four mutually independent signals; an annotator is removed only when several signals are simultaneously abnormal:

\begin{enumerate}
    \item \textbf{Score Constancy}: The standard deviation of an annotator's scores. $\text{std}=0$ means assigning the same score to every dimension of every video and provides no discriminating information whatsoever; $\text{std}<0.3$ is treated as near-constant.
    \item \textbf{Per-video Copy-paste Rate}: Whether the annotator gives identical scores to all dimensions within each video. A 100\% copy-paste rate means the annotator does not differentiate evaluation dimensions (e.g., gravity vs.\ collision), so the labels should be treated as invalid even when the annotator's cross-video overall std is greater than zero. This signal is complementary to score constancy: copy-paste annotators can have an overall std greater than zero and thus escape constancy detection.
    \item \textbf{Peer MAE}: The mean absolute deviation between an annotator's scores and those of other annotators on the same video-dimension cells, averaged over all pairwise comparisons in which the annotator participated. On the $1$--$5$ scale, peer MAE values above $1.8$ are flagged as extreme, corresponding to disagreeing with peers by nearly two scale points on average; this threshold is roughly twice the population's typical pairwise disagreement and indicates either systematic bias or essentially random scoring. Because honest annotators occasionally disagree (e.g., stricter or more lenient calibration), peer MAE alone is not used as a removal criterion, it must co-occur with a behavioral anomaly or a high copy-paste rate.
    \item \textbf{Behavioral Signals}: Two platform-side signals are recorded for every annotation submission. \emph{Page stay time} is the elapsed seconds between page load and rating submission; the median across an annotator's submissions below $30$ seconds is treated as short stay (the videos themselves are typically several seconds long, and rating $3$ general dimensions plus $2$--$4$ physical laws on a Likert scale realistically requires at least half a minute), and below $10$ seconds is treated as too short for any deliberate judgment. \emph{Video play count} is the number of times the annotator triggered playback before submitting; a maximum play count of zero across all of an annotator's submissions means the rating was entered without ever playing the video, which is incompatible with the instruction that each video must be watched at least once before rating. Either signal in isolation is noisy: a fast but careful annotator can submit quickly, and the play-count counter occasionally fails to register, so behavioral anomalies are only used in conjunction with copy-paste or peer-MAE evidence.
\end{enumerate}

In total, 23.3\% of annotators and their associated ratings were removed by this pipeline (459 raw annotators to 352 retained annotators).

\subsection{Removed Annotators}

Removed annotators were identified using several quality-control rules rather than mutually exclusive categories. 
The largest source of removals was constant or near-constant scoring, where annotators assigned identical scores across all dimensions and videos ($\text{std}=0$) or exhibited very low score variation ($\text{std}<0.3$). 
We additionally flagged a distinct per-video copy-paste pattern: some annotators assigned identical scores to all dimensions within each video, while varying their scores across different videos. 
Because these annotators can have nonzero overall variance, they are not captured by the constancy rule alone. 
This pattern is methodologically important because it indicates that the annotator did not differentiate evaluation dimensions, making the labels uninformative for per-law diagnostic scoring.
Finally, we removed annotators when multiple weaker signals jointly indicated unreliable behavior: (i) high copy-paste rates ($\geq 75\%$) combined with short stay time or unplayed videos, (ii) extremely low annotation volume or stay time, and (iii) extreme peer MAE paired with behavioral abnormalities such as short stay, unplayed videos, or high copy-paste rates.

\subsection{Difficult-Item Analysis}

To assess whether annotation difficulty leads to unreliable or skipped judgments, we group videos into four buckets (easy / medium / hard / very\_hard) by inter-annotator pairwise MAE; roughly 30\% of videos fall in the very\_hard bucket. The key findings are:

\begin{itemize}
    \item \textbf{Skip rate is low}: the total skip rate is small and shows no clear correlation with difficulty---there is no systematic ``skip when hard'' pattern.
    \item \textbf{Stay time grows with difficulty}: very\_hard items have the longest average stay time, indicating that annotators invest more time on harder items rather than rushing through them.
    \item \textbf{Copy-paste rate rises slightly on the very\_hard group}: a small number of annotators tend to assign identical dimension scores when uncertain, but they are few in number and contribute few labels on hard items, so their effect on the aggregated result (mean) is limited.
    \item \textbf{Cross-dimension difficulty is similar}: no physical dimension is markedly harder to annotate than the others.
\end{itemize}

The key evidence is that very\_hard items receive the longest average stay time, which is inconsistent with systematic skipping. Our cleaning procedure already removes annotators with clear copy-paste patterns; the remaining low-effort cases are sparse and isolated, rather than systematic.

\subsection{Ranking Robustness}

The aggregate model ranking is reproducible across complementary resampling protocols.
\Cref{tab:robustness_legacy} reports, for each of the eight models, the mean-aggregated Overall score (matching the published leaderboard pipeline: per-video annotator-mean, then model-level mean over videos for General and sample-weighted mean over (vid, law) units for Physics), two $95\%$ bootstrap confidence intervals (CI), and a median-aggregation comparison. The prompt CI resamples prompts with replacement, probing sensitivity to the sampled prompt set, while the annotator CI resamples annotator scores within each fixed video, probing sensitivity to which annotators rated each video. The final two columns report the rank and score obtained by replacing the per-video annotator mean with an annotator median, while keeping the rest of the model-level pipeline unchanged.

Both bootstrap CIs are tight and mostly non-overlapping between adjacent models, indicating that prompt and annotator sampling noise alone is unlikely to change the ranking. The mean-vs.-median comparison is a stronger perturbation because it changes the aggregation rule itself: mean aggregation averages annotator scores within each video before propagating to the model level, whereas median aggregation takes the median annotator score for each video and dimension. Even under this change, none of the eight model ranks move, and the Overall score shifts by at most $\approx\!0.02$ on the $1$--$5$ scale; since adjacent-model gaps remain larger than this perturbation, the published mean ordering is robust to the aggregation choice.

\input{appendix/tables/robustness_analysis.tex}

\paragraph{Split-half rank-correlation intervals.}
In each of 100 random trials, the 352 post-QC annotators are split into two equal halves; every video model is scored independently in each half using the same leaderboard aggregation pipeline, and the two resulting eight-model score vectors are compared with Spearman $\rho$ and Kendall $\tau$-b. We compute the mean, standard deviation, and empirical 95\% interval (the 2.5th and 97.5th percentiles over the 100 splits). The intervals stay high across all three score views: Spearman $\rho$ is $0.930 \pm 0.045$ for General, $0.907 \pm 0.058$ for Physics, and $0.928 \pm 0.051$ for Overall, while Kendall $\tau$-b is $0.828 \pm 0.071$, $0.801 \pm 0.092$, and $0.830 \pm 0.088$, respectively. Even the lower ends of the empirical intervals remain positive and large (Spearman $\rho \geq 0.773$, Kendall $\tau$-b $\geq 0.605$), indicating that independently sampled annotator halves recover highly similar model orderings.

\paragraph{Annotator-pool sub-sampling.}
We additionally probe how the leaderboard responds to shrinking the annotator pool. For each setting, we draw 200 random sub-samples of annotators and recompute every model's mean over the SA / PTV / persistence general dimensions, then report the root-mean-square deviation (RMS $\Delta$, in 1--5 score units) from the full-pool baseline. Dropping a random $30\%$ of annotators perturbs per-model scores by only $0.033$ RMS; sub-sampling down to $20$ annotators inflates this to $0.204$ RMS---roughly a sixfold increase in score noise. \Cref{tab:robustness_loo} reports the full sweep.

\input{appendix/tables/robustness_loo.tex}

\subsection{Annotator Diversity}

To quantify how broadly the annotator pool is distributed across demographic facets, we compute an Annotator Diversity Index (ADI): for each facet (gender, education, major, age, cohort) we report the effective number of categories $\exp(H)$ of the annotator distribution, and take the geometric mean across the five facets as the ADI. \Cref{tab:annotator_diversity_index} reports the per-facet numbers alongside a hypothetical 12-CS-PhD expert panel for contrast: our ADI is $2.01\times$ that panel, with the largest gap on \emph{major} (eight coarse fields of study) and \emph{cohort} (four recruitment cohorts).

\input{appendix/tables/diversity_index.tex}

%% file: tables/TABLE_vote_statistics.tex
\begin{table}[t]
\centering
\caption{Vote statistics of PhyGround before annotator filtering. An \emph{annotation} is one annotator's complete submission of all dimension scores for a single video; a \emph{label} is one such per-dimension score, covering SA, PTV, persistence, and each physics law applicable to that video, so each annotation contributes multiple labels.}
\label{tab:vote_statistics}
\small
\begin{tabular}{@{}lc@{}}
\toprule
\multicolumn{2}{c}{\textbf{Basic Statistics}} \\
\midrule
\# annotations & 5,796 \\
\# annotators & 459 \\
\# annotations per video & 2.90 \\
\# labels & 37.4K \\
\bottomrule
\end{tabular}
\end{table}

%% file: appendix/tables/robustness_analysis.tex
\begin{table}[htbp]
\centering
\caption{Robustness checks under the published mean-aggregation pipeline. Prompt CI resamples prompts with replacement; annotator CI resamples annotator scores within each fixed video; the last two columns swap the per-video annotator mean for an annotator median.}
\label{tab:robustness_legacy}
\resizebox{\textwidth}{!}{%
\begin{tabular}{l l c c c c c}
\toprule
Rank & Model & Mean score & Prompt 95\% CI & Annotator 95\% CI & Median rank & Median--mean \\
\midrule
    1 & Wan2.2-27B-A14B & 3.282 & 3.197--3.371 & 3.234--3.343 & 1 & +0.019 \\
    2 & Veo-3.1 & 3.280 & 3.184--3.373 & 3.228--3.339 & 2 & +0.006 \\
    3 & OmniWeaving & 3.101 & 3.021--3.185 & 3.060--3.156 & 3 & +0.007 \\
    4 & Cosmos-14B & 2.904 & 2.824--3.009 & 2.852--2.947 & 4 & +0.004 \\
    5 & LTX-2.3-22B & 2.698 & 2.600--2.785 & 2.644--2.743 & 5 & -0.012 \\
    6 & Wan2.2-TI2V-5B & 2.628 & 2.528--2.708 & 2.585--2.682 & 6 & +0.001 \\
    7 & Cosmos-2B & 2.585 & 2.488--2.697 & 2.546--2.629 & 7 & +0.010 \\
    8 & LTX-2-19B & 2.564 & 2.475--2.645 & 2.520--2.611 & 8 & -0.021 \\
\bottomrule
\end{tabular}%
}
\end{table}

%% file: appendix/tables/robustness_loo.tex
\begin{table}[htbp]
\centering
\caption{Annotator-pool sub-sampling stability of per-model leaderboard scores. \emph{Top}: drop a random fraction of annotators (simulated). \emph{Middle}: the actual quality-control drop, single-shot, comparing the pre-QC pool of 459 annotators against the 352-annotator post-QC baseline (459$\to$352, a 23.3\% targeted drop). \emph{Bottom}: keep only $N$ uniformly sampled annotators. Random rows average over 200 sub-samples (seed 0); the QC row is a single observation. RMS $\Delta$ is the root-mean-square deviation of a per-model mean (pooled over the SA / PTV / persistence general dimensions, in 1--5 score units) from the full-pool post-QC baseline (352 annotators, 29,541 scores); \emph{mean} and \emph{max} are taken across the 8 video models on the leaderboard. Smaller is more stable. The bottom block approximates the sampling instability one would expect from a much smaller annotator pool, without modelling expertise-specific scoring behaviour.}
\label{tab:robustness_loo}
\begin{tabular}{l r r r}
\toprule
Sub-sample & mean RMS $\Delta$ & max RMS $\Delta$ & $n_{\text{models}}$ \\
\midrule
\multicolumn{4}{l}{\emph{Drop a random fraction of annotators (simulated)}} \\
\quad drop 5 \% & 0.0115 & 0.0122 & 8 \\
\quad drop 10 \% & 0.0172 & 0.0197 & 8 \\
\quad drop 20 \% & 0.0247 & 0.0290 & 8 \\
\quad drop 30 \% & 0.0333 & 0.0375 & 8 \\
\midrule
\multicolumn{4}{l}{\emph{Observed quality-control drop (real, single-shot)}} \\
\quad QC drop 23.3\% ($459\!\to\!352$) & 0.0585 & 0.1065 & 8 \\
\midrule
\multicolumn{4}{l}{\emph{Keep only $N$ uniformly sampled annotators}} \\
\quad $N = 10$ & 0.2944 & 0.3209 & 8 \\
\quad $N = 20$ & 0.2037 & 0.2391 & 8 \\
\quad $N = 50$ & 0.1228 & 0.1458 & 8 \\
\quad $N = 100$ & 0.0794 & 0.0894 & 8 \\
\bottomrule
\end{tabular}
\end{table}

%% file: appendix/tables/diversity_index.tex
\begin{table}[htbp]
\centering
\caption{Annotator Diversity Index. For each demographic facet we report effective categories $\exp(H)$, where $H = -\sum_i p_i \log p_i$ is the Shannon entropy of the annotator distribution: a panel uniform across $k$ buckets has $\exp(H)=k$, a panel concentrated in one bucket has $\exp(H)=1$. ADI (last row) is the geometric mean of $\exp(H)$ across the five facets. The expert column is a hypothetical 12-CS-PhD panel (gender 6/6, education=PhD, major=CS, age=\{26--30:6, 31--40:6\}, cohort=single\_lab) shown purely for contrast; our ADI is 2.01$\times$ that panel.}
\label{tab:annotator_diversity_index}
\begin{tabular}{l r r r}
\toprule
Facet & Ours $\exp(H)$ & Ours buckets & Expert $\exp(H)$ \\
\midrule
    Gender & \textbf{2.15} & 4 & 2.00 \\
    Education & \textbf{1.90} & 3 & 1.00 \\
    Major & \textbf{3.17} & 8 & 1.00 \\
    Age & \textbf{3.47} & 6 & 2.00 \\
    Cohort & \textbf{2.95} & 4 & 1.00 \\
    \midrule
    \textbf{ADI} & \textbf{2.66} &  & \textbf{1.32} \\
\bottomrule
\end{tabular}
\end{table}

%% file: appendix/aggregation_sensitivity.tex
\section{Aggregation Sensitivity}
\label{sec:appendix_aggregation_sensitivity}

The Physics score averages per-(model, video, law) scores within each domain and weights domains by the number of scored prompt-video-law instances per model: 693 for Solid-Body, 136 for Fluid, and 47 for Optical. We test this choice by recomputing Overall scores with equal-law aggregation schemes that reduce prompt-count weighting. The rankings remain nearly unchanged, suggesting that high-frequency laws do not drive the headline leaderboard.

\subsection{Schemes}

\begin{itemize}
    \item \textbf{Sample-weighted} (current): the domain mean is the mean of all per-video law scores in that domain, so laws with more prompts pull harder; the Physics aggregate is then $(693 \cdot \text{Solid} + 136 \cdot \text{Fluid} + 47 \cdot \text{Optical}) / 876$.
    \item \textbf{Equal-law-within-domain}: the domain mean is the unweighted average of the per-law means inside that domain, so within Solid-Body the six laws (gravity, inertia, momentum, impenetrability, collision, material) each contribute $1/6$ regardless of prompt count, while Physics still weights domains by $n$.
    \item \textbf{Equal-law-global}: the Physics aggregate is the unweighted mean of all 13 per-law means (6 Solid-Body, 5 Fluid, 2 Optical), ignoring domain grouping; each law contributes $1/13 \approx 7.7\%$ regardless of prompt count or domain size, so the implied total domain weights become Solid $6/13 = 46.2\%$, Fluid $5/13 = 38.5\%$, Optical $2/13 = 15.4\%$.
\end{itemize}

In all cases, General is the unweighted mean of (SA, PTV, Persistence), and Overall $= 0.5 \cdot \text{General} + 0.5 \cdot \text{Physics}$.

\subsection{Per-Domain $\Delta$ from Removing Within-Domain Law-Count Imbalance}

\Cref{tab:agg_per_domain_delta} reports, per domain, the absolute gap between the equal-law-within-domain domain mean and the sample-weighted domain mean. The mean and max are taken across the eight comparison models, and the third column names the model that attains the maximum gap. The table lets one read off how much within-domain law-count imbalance moves each domain mean before any cross-domain reweighting is applied.

\begin{table}[htbp]
    \centering
    \caption{Per-domain absolute gap $\Delta = |\text{equal-law-within-domain domain mean} - \text{sample-weighted domain mean}|$, summarized across the 8 comparison models.}
    \label{tab:agg_per_domain_delta}
    \begin{tabular}{lccc}
        \toprule
        Domain & Mean abs $\Delta$ & Max abs $\Delta$ & Model with max abs $\Delta$ \\
        \midrule
        Solid-Body & 0.014 & 0.03 & Cosmos-2B / Veo-3.1 \\
        Fluid      & 0.014 & 0.03 & Cosmos-14B \\
        Optical    & 0.030 & 0.07 & Cosmos-2B \\
        \bottomrule
    \end{tabular}
\end{table}

\subsection{Overall Sensitivity Under Each Alternative Scheme}

\Cref{tab:agg_overall_sensitivity} reports, for each alternative scheme, the absolute Overall shift relative to the sample-weighted baseline together with the induced rank changes. The table lets one read off how much each weighting choice moves absolute Overall scores and how many of the eight models change rank as a result.

\begin{table}[htbp]
    \centering
    \caption{Sensitivity of the Overall score and ranking to alternative Physics aggregation schemes. Rank changes are computed relative to the sample-weighted baseline.}
    \label{tab:agg_overall_sensitivity}
    \begin{tabular}{lcccc}
        \toprule
        Scheme & Max abs $\Delta$ & Mean abs $\Delta$ & Rank changes & Max rank shift \\
        \midrule
        Equal-law-within-domain & 0.015 & 0.007 & 2/8 & 1 \\
        Equal-law-global        & 0.099 & 0.063 & 2/8 & 1 \\
        \bottomrule
    \end{tabular}
\end{table}

\subsection{Takeaway}

Within-domain law imbalance contributes essentially nothing to the Overall score (mean abs $\Delta \approx 0.007$). Equal-law-global shifts absolute Overall scores by tens of milli-points, but rankings remain stable up to a one-position shift. The only persistent swap is the Wan2.2-27B-A14B $\leftrightarrow$ Veo-3.1 exchange at the top of the leaderboard, which is already a near-tie under sample-weighting ($\Delta$ Overall $= 0.002$). The headline ordering appears broadly consistent across weighting schemes, with Wan2.2-27B-A14B / Veo-3.1 leading, then OmniWeaving, then Cosmos-14B.

%% file: appendix/scoreperlaw.tex
\section{Per-Law Score Breakdown}
\label{sec:appendix_scoreperlaw}

This appendix drills one level finer, reporting the human-evaluation mean for every individual law in \Cref{tab:law_breakdown}.
The four observations below are scoped to per-law signals that the three-domain aggregate averages over.

\input{tables/TABLE_law_breakdown.tex}

\paragraph{Observation 1: Within solid-body, the per-law leaderboard does not match the domain leaderboard.}
Even though Wan2.2-27B-A14B leads the solid-body domain aggregate, two of the six solid-body laws (Inertia, Material) are won by other models, and Material is in fact the only law in the table where neither the leading open-source model nor the closed-source model is best.
Treating the solid-body domain as a single capability would mask these within-domain leadership reversals; per-law decomposition exposes them as candidate axes for targeted improvement.

\paragraph{Observation 2: The solid-body bottleneck is concentrated in Momentum and Collision.}
Aggregating the table column-wise, Momentum and Collision are the two hardest laws across the entire table---both noticeably below Impenetrability (the easiest solid-body axis) and well below Optical.
What separates the easy and hard solid-body laws is whether contact dynamics are involved: free-motion laws (Gravity, Inertia) and the static non-overlap constraint (Impenetrability) score uniformly higher than the dynamic-contact laws (Momentum, Collision).
The bottleneck for current models is therefore \emph{contact resolution}---rebound direction, restitution, fracture pattern---rather than ballistic motion or geometric non-overlap.

\paragraph{Observation 3: Veo-3.1's fluid advantage is concentrated in flow patterns, not conservation.}
Veo-3.1 leads all three fluid groups, but the size of that lead varies sharply with which fluid axis is being measured: it is decisive on Liq-Solid and FlowDyn, but on Conserv.\ (fluid mass/volume conservation) the lead over OmniWeaving collapses to a near-tie.
The closed-source fluid advantage is therefore primarily about the realism of bulk-flow patterns and liquid--solid interaction, not about preserving mass and volume---open-source models are already nearly competitive on the latter axis, and ``fluid'' as a domain-level label conflates two qualitatively different capability gaps.

\paragraph{Observation 4: Per-law within-model spread separates balanced from spiky physics profiles.}
The spread between a model's best and worst law differs substantially across models: Wan2.2-27B-A14B is the most balanced (a small spread, with no obviously weak law), while Cosmos-Predict2.5-2B is the most spiky (a wide spread that crosses nearly a full rating level between its weakest and strongest laws).
LTX-2-19B is the clearest example of within-model rank flipping---it is last on Impenetrability yet competitive on the fluid laws, so its leaderboard rank changes as one moves between physics axes.
A balanced profile indicates that a single underlying capability lifts every law together, while a spiky profile indicates partially specialized capabilities decoupled across physics axes and is the most informative starting point for targeted post-training---the law that should be fixed first is different for each model.

\section{Per-Law Qualitative Examples of Judge--Human Alignment}
\label{sec:appendix_judge_examples}

\begin{figure}[htbp]
    \centering
    \includegraphics[width=\linewidth]{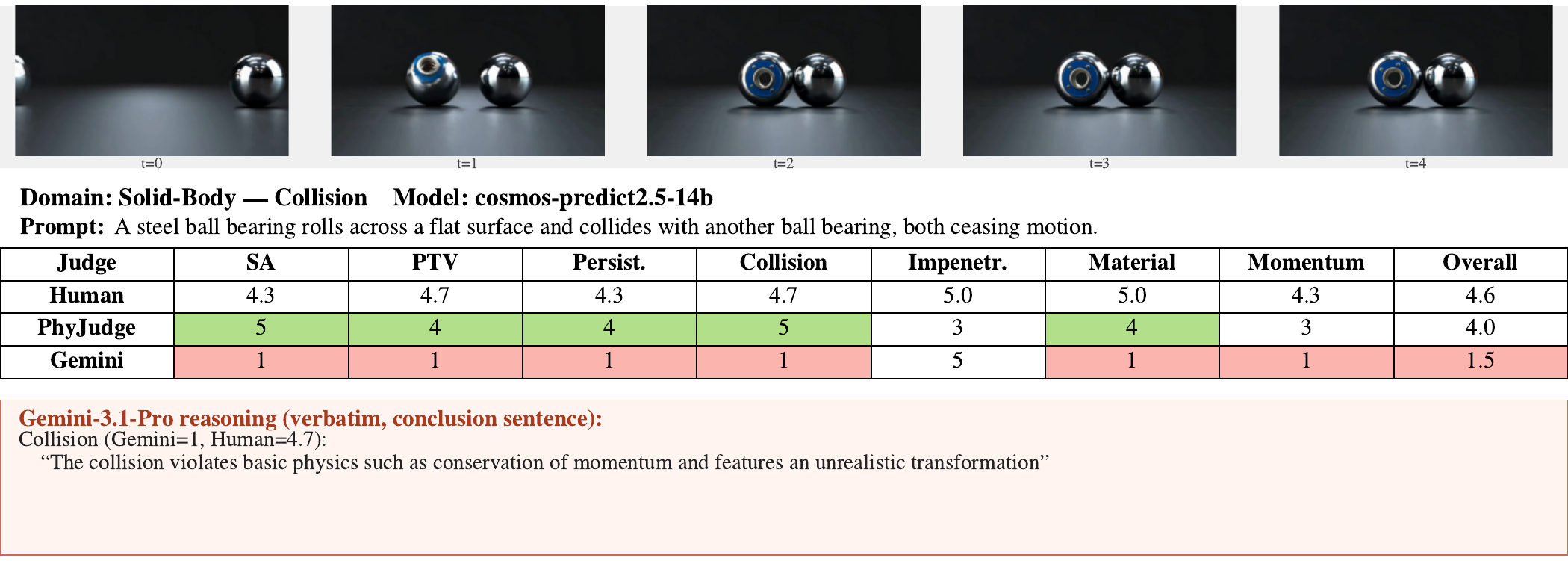}
    \caption{\textbf{Collision} on a Cosmos-Predict2.5-14B generation: two steel ball bearings collide and come to rest. Humans rate the collision realistic (4.7); PhyJudge-9B agrees (5.0), while Gemini-3.1-Pro reads the same outcome as a near-total physics violation (1.0). The example isolates a contact-resolution case where the closed-source judge over-penalizes plausible post-impact rest.}
    \label{fig:judge_example_collision}
\end{figure}

\begin{figure}[htbp]
    \centering
    \includegraphics[width=\linewidth]{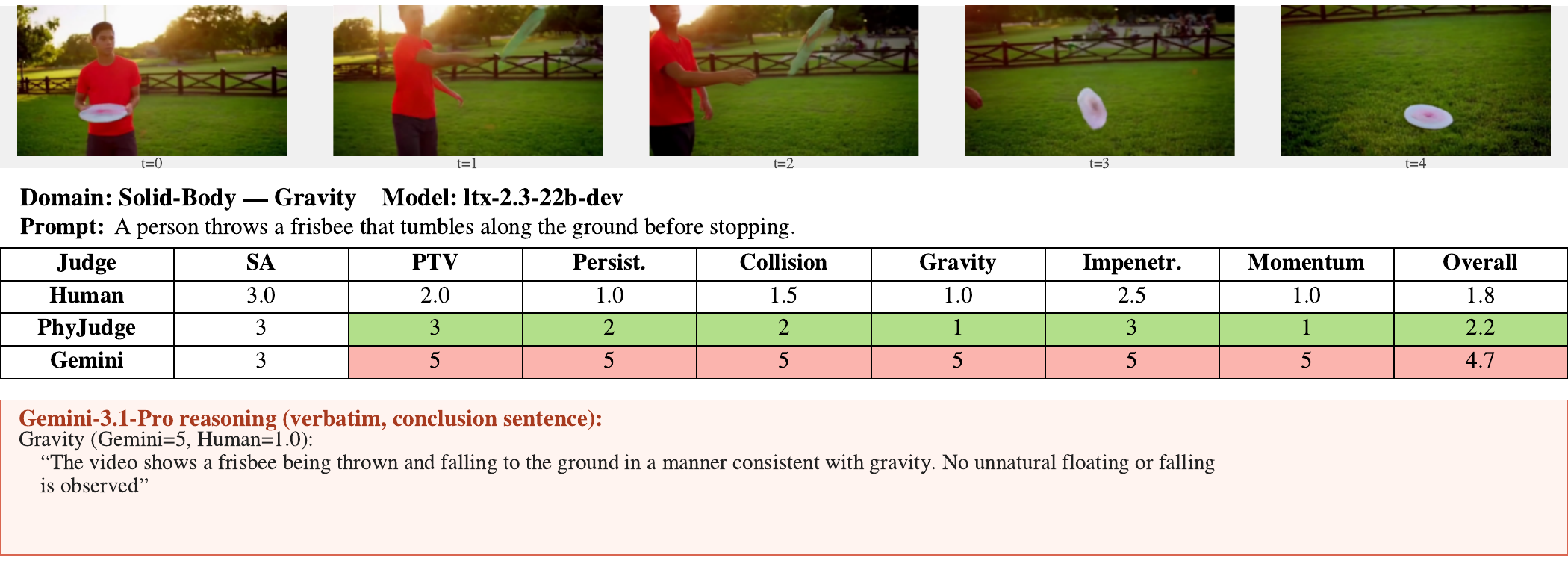}
    \caption{\textbf{Gravity} on an LTX-2.3-22B generation: a thrown frisbee tumbles along the ground. Humans rate gravity broken (1.0) and PhyJudge-9B agrees (1.0), while Gemini-3.1-Pro scores 5.0, illustrating an over-score where Gemini misses an obvious free-fall failure that humans flag.}
    \label{fig:judge_example_gravity}
\end{figure}

\begin{figure}[htbp]
    \centering
    \includegraphics[width=\linewidth]{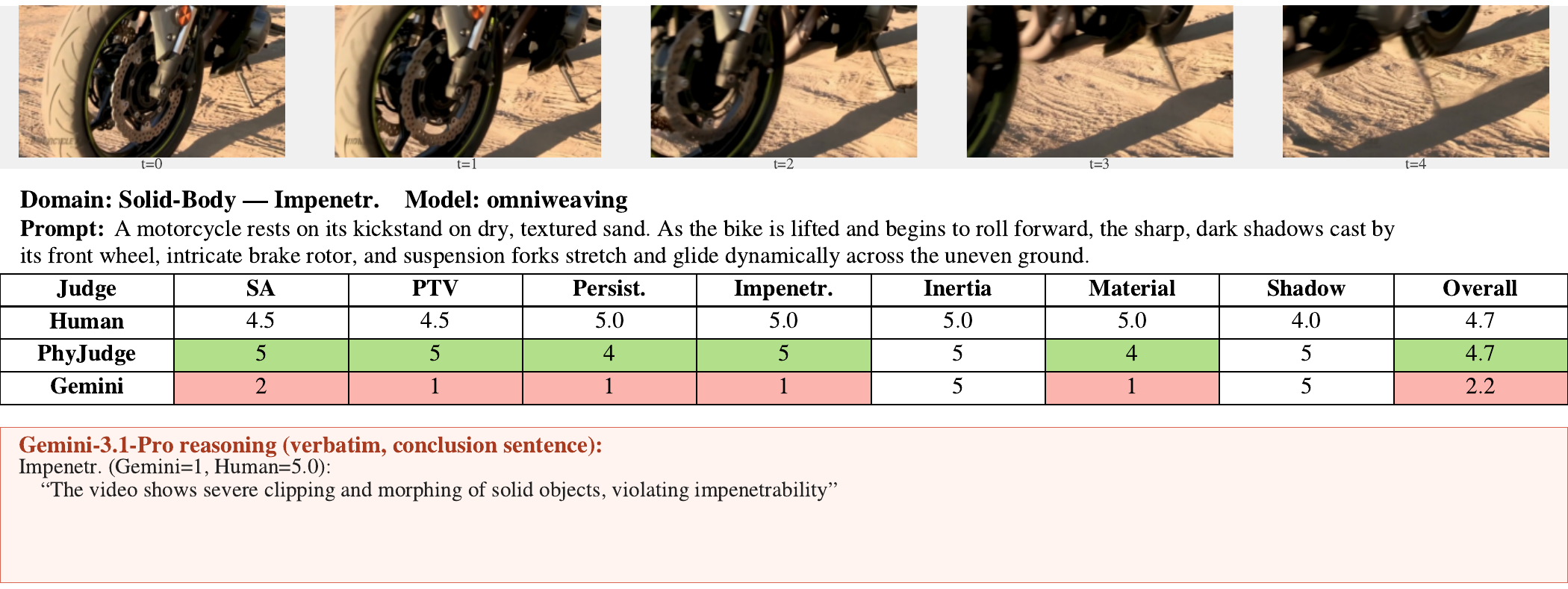}
    \caption{\textbf{Impenetrability} on an OmniWeaving generation: a motorcycle is lifted off its kickstand and begins to roll on textured sand. Humans see no inter-penetration (5.0) and PhyJudge-9B agrees (5.0), while Gemini-3.1-Pro scores 1.0 despite the absence of any visible solid--solid overlap.}
    \label{fig:judge_example_impenetrability}
\end{figure}

\begin{figure}[htbp]
    \centering
    \includegraphics[width=\linewidth]{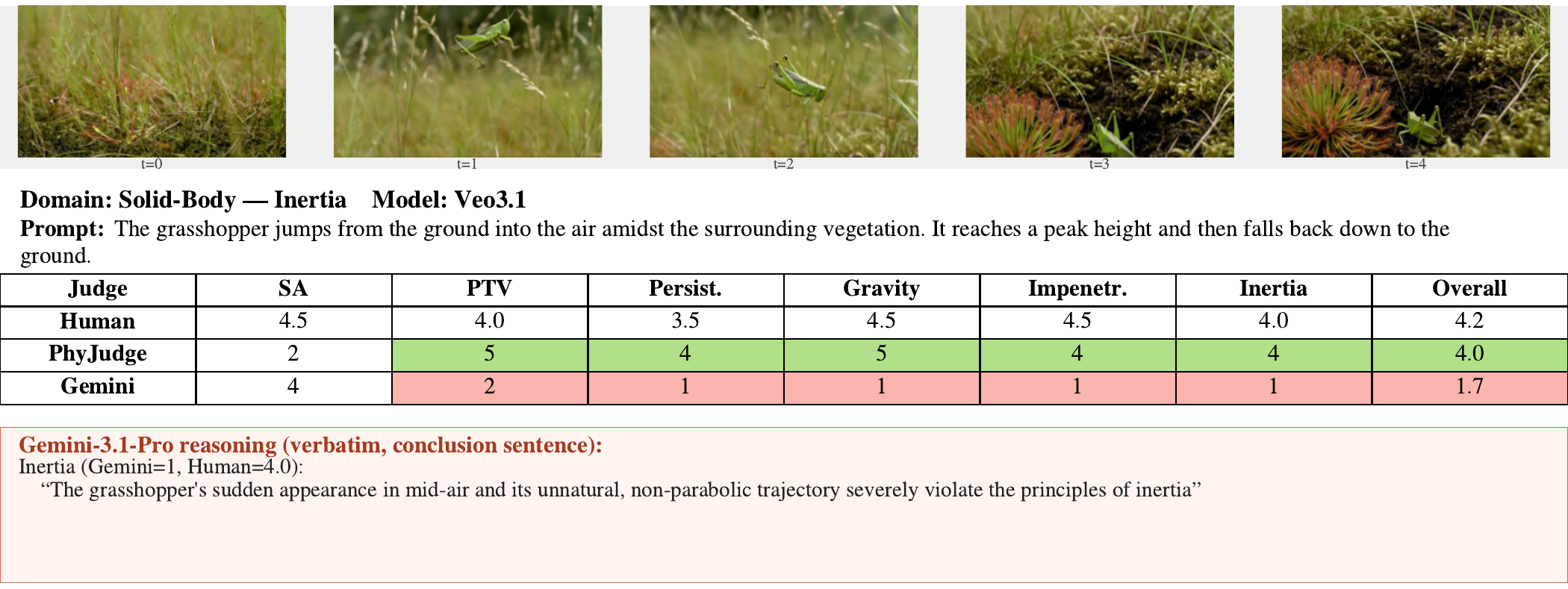}
    \caption{\textbf{Inertia} on a Veo-3.1 generation: a grasshopper jumps from the ground, reaches a peak, and falls back. Humans rate the inertial trajectory plausible (4.0) and PhyJudge-9B agrees (4.0), while Gemini-3.1-Pro scores 1.0, treating the same ballistic motion as a near-total inertia violation.}
    \label{fig:judge_example_inertia}
\end{figure}

\begin{figure}[htbp]
    \centering
    \includegraphics[width=\linewidth]{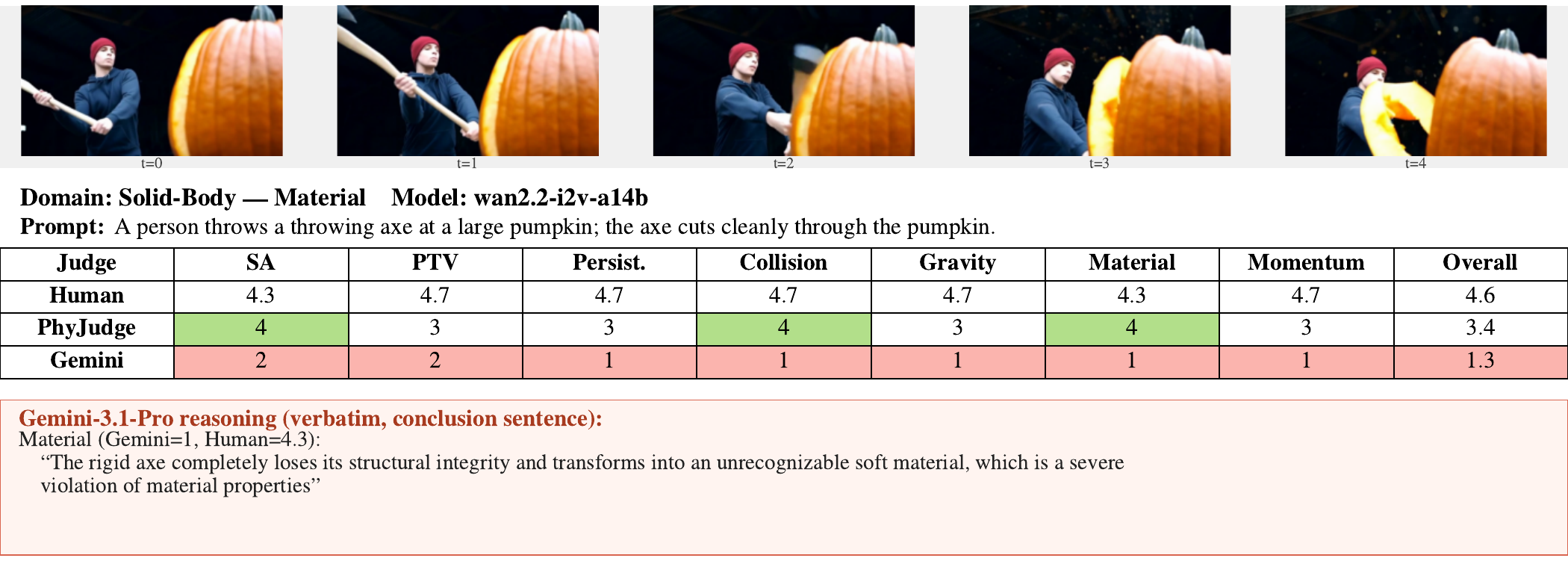}
    \caption{\textbf{Material} on a Wan2.2-27B-A14B generation: a thrown axe cuts cleanly through a pumpkin. Humans rate the material response plausible (4.3) and PhyJudge-9B tracks them (4.0), while Gemini-3.1-Pro scores 1.0, an over-penalization on a case where humans accept the cut behavior.}
    \label{fig:judge_example_material}
\end{figure}

\begin{figure}[htbp]
    \centering
    \includegraphics[width=\linewidth]{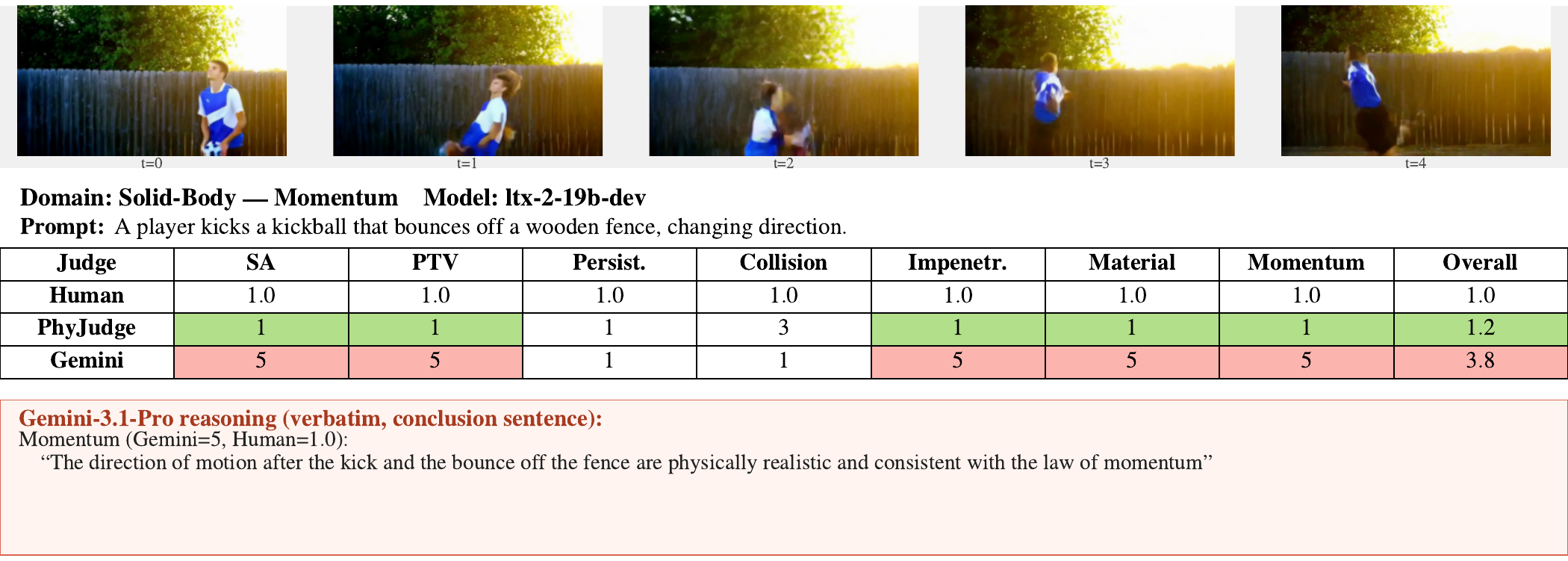}
    \caption{\textbf{Momentum} on an LTX-2-19B generation: a player is supposed to kick a kickball that bounces off a wooden fence, but the generated clip contains no kick and no bounce, so no momentum transfer is depicted. Humans rate momentum broken (1.0) and PhyJudge-9B agrees (1.0), while Gemini-3.1-Pro hallucinates near-perfect momentum behavior (5.0) on the same clip.}
    \label{fig:judge_example_momentum}
\end{figure}

\begin{figure}[htbp]
    \centering
    \includegraphics[width=\linewidth]{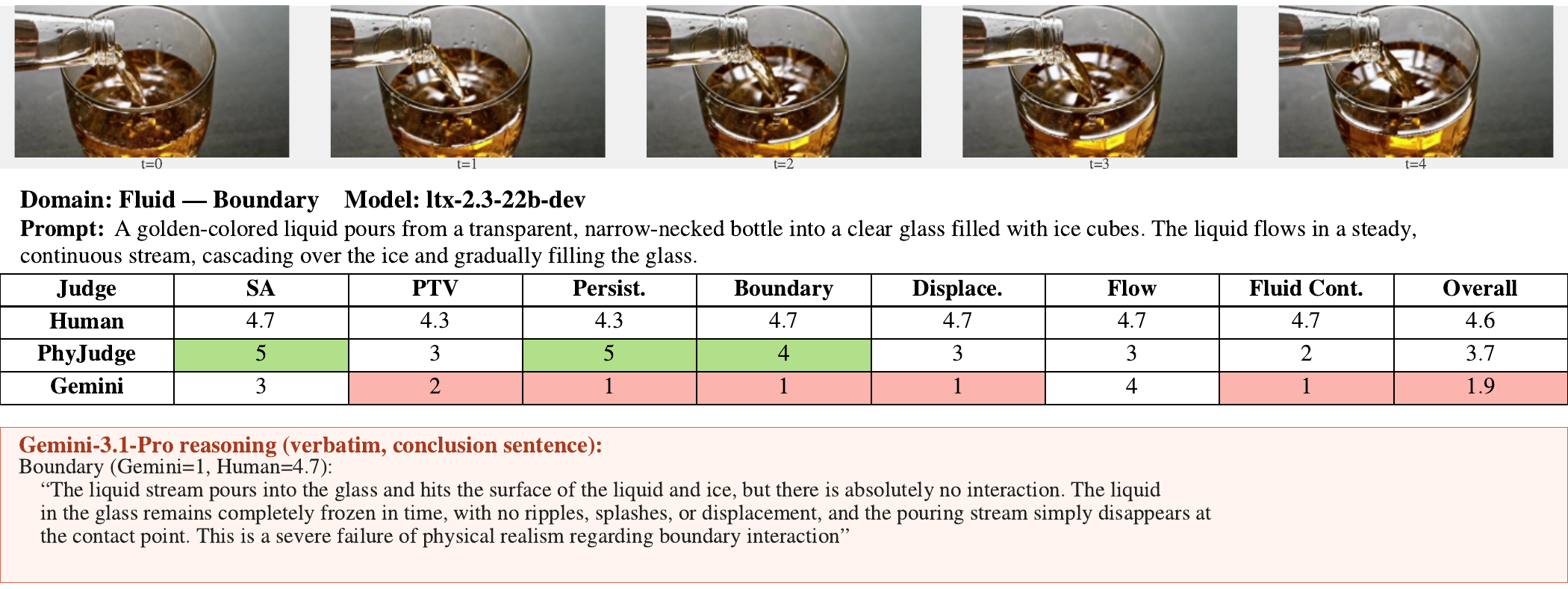}
    \caption{\textbf{Boundary interaction} on an LTX-2.3-22B generation: a golden liquid pours over ice cubes inside a glass. Humans see realistic deflection and splash on the ice (4.7), and PhyJudge-9B follows them (4.0), while Gemini-3.1-Pro reads the same liquid--solid interaction as broken (1.0). This is a representative fluid case where Gemini-3.1-Pro is substantially more pessimistic than human raters.}
    \label{fig:judge_example_boundary_interaction}
\end{figure}

\begin{figure}[htbp]
    \centering
    \includegraphics[width=\linewidth]{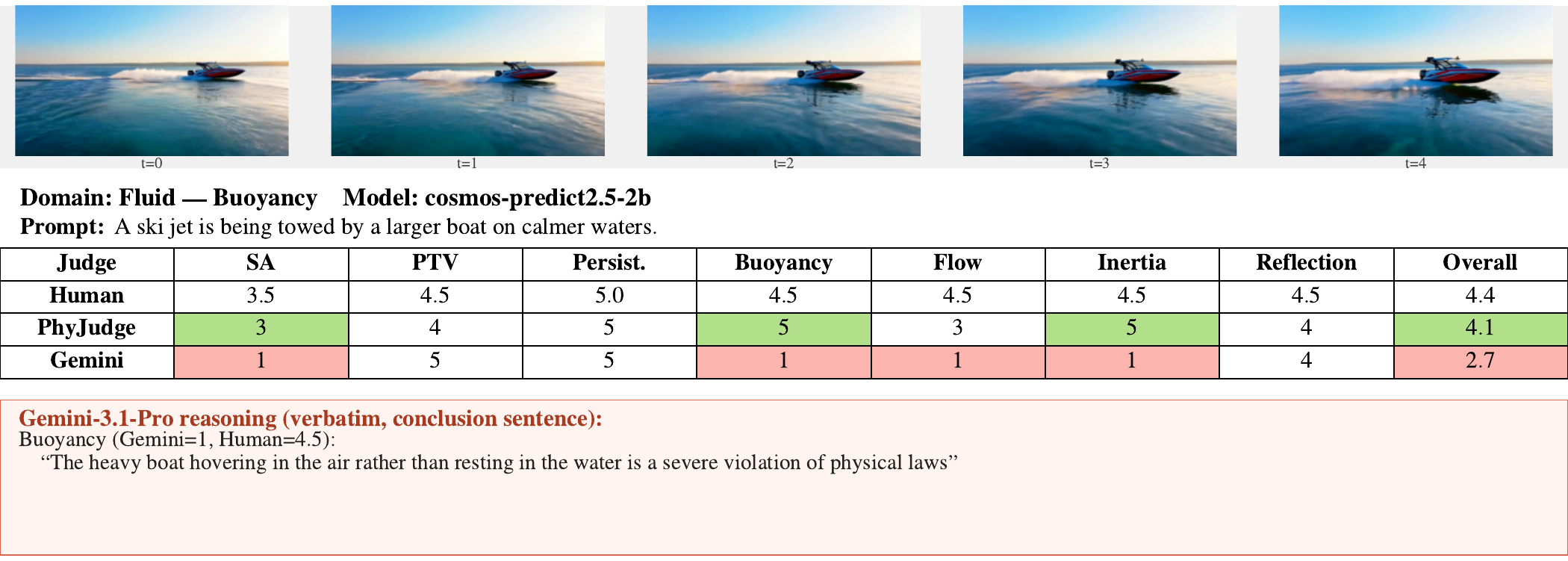}
    \caption{\textbf{Buoyancy} on a Cosmos-Predict2.5-2B generation: a ski jet is towed across calm water. Humans rate buoyancy plausible (4.5); PhyJudge-9B agrees (5.0), while Gemini-3.1-Pro scores 1.0, treating the floating motion as a violation that humans accept.}
    \label{fig:judge_example_buoyancy}
\end{figure}

\begin{figure}[htbp]
    \centering
    \includegraphics[width=\linewidth]{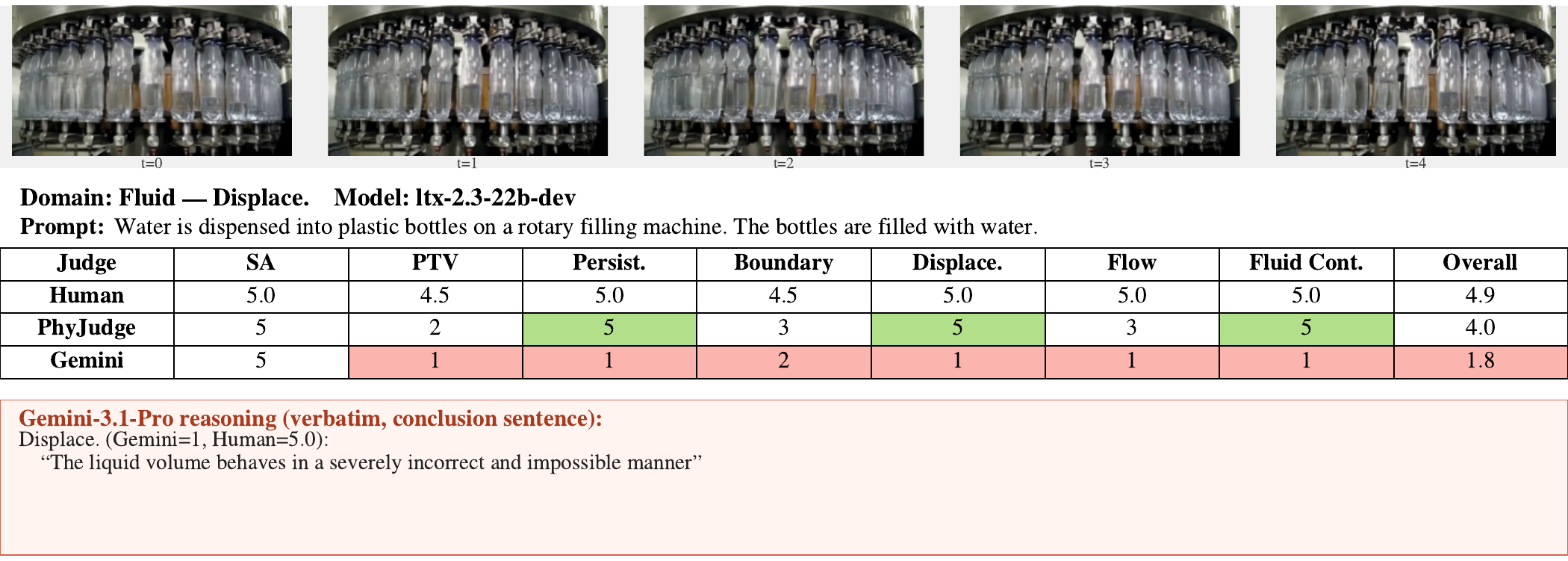}
    \caption{\textbf{Displacement} on an LTX-2.3-22B generation: water is dispensed into plastic bottles on a rotary filling machine, with sustained flow and steady volume transfer into each bottle. Humans rate the volume transfer plausible (5.0) and PhyJudge-9B matches them exactly (5.0), while Gemini-3.1-Pro scores 1.0, illustrating the fluid-pessimism mode in which it reads continuous filling as a violation.}
    \label{fig:judge_example_displacement}
\end{figure}

\begin{figure}[htbp]
    \centering
    \includegraphics[width=\linewidth]{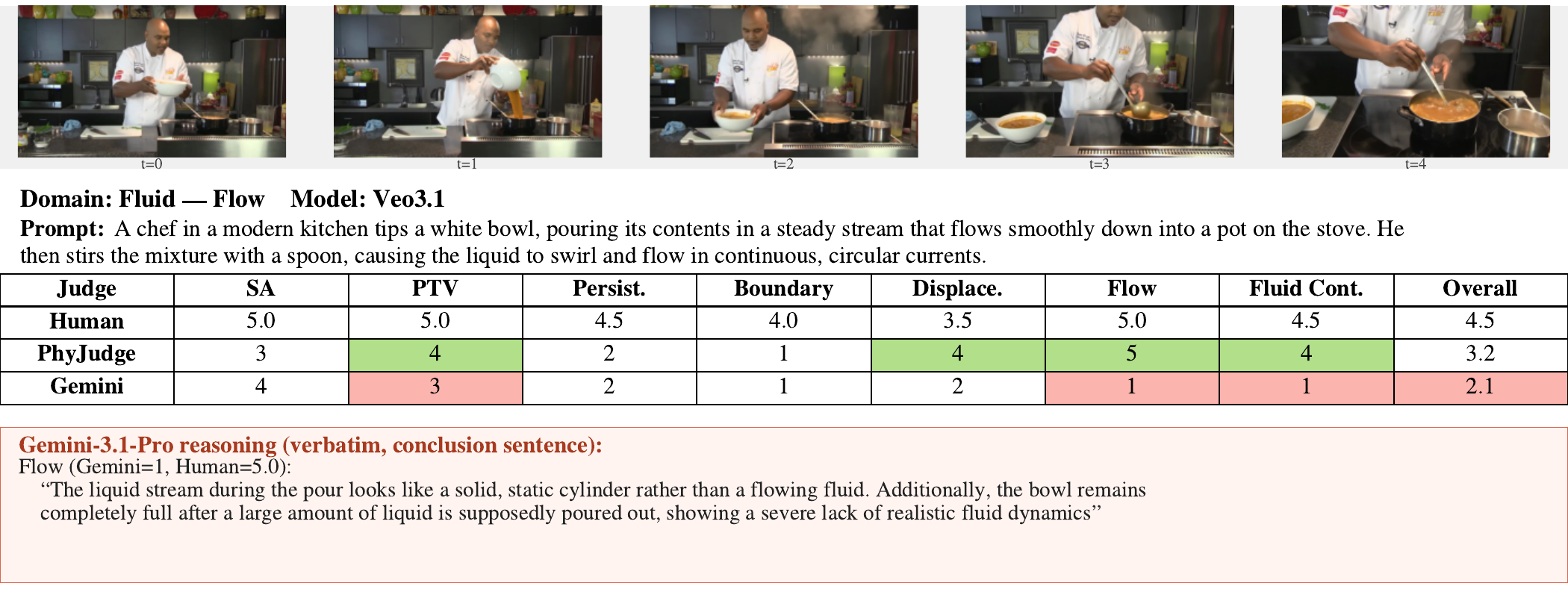}
    \caption{\textbf{Flow dynamics} on a Veo-3.1 generation: a chef pours from a bowl into a pot and stirs the resulting mixture. Humans rate bulk-flow patterns realistic (5.0) and PhyJudge-9B matches them (5.0), while Gemini-3.1-Pro collapses to 1.0 on the same clip.}
    \label{fig:judge_example_flow_dynamics}
\end{figure}

\begin{figure}[htbp]
    \centering
    \includegraphics[width=\linewidth]{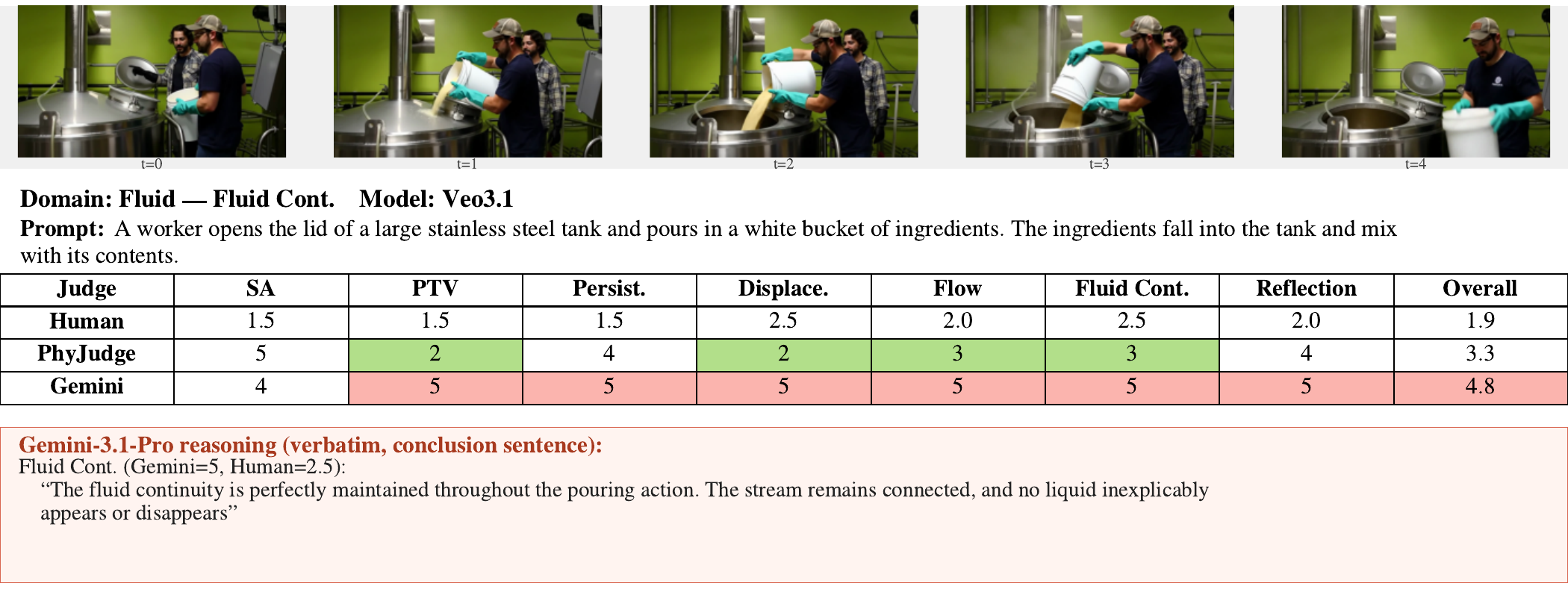}
    \caption{\textbf{Fluid continuity} on a Veo-3.1 generation: a worker pours a bucket of ingredients into a stainless tank. Humans see partial continuity violations (2.5) and PhyJudge-9B agrees (3.0), while Gemini-3.1-Pro scores 5.0---an over-score in the opposite direction from the typical fluid-pessimism mode.}
    \label{fig:judge_example_fluid_continuity}
\end{figure}

\begin{figure}[htbp]
    \centering
    \includegraphics[width=\linewidth]{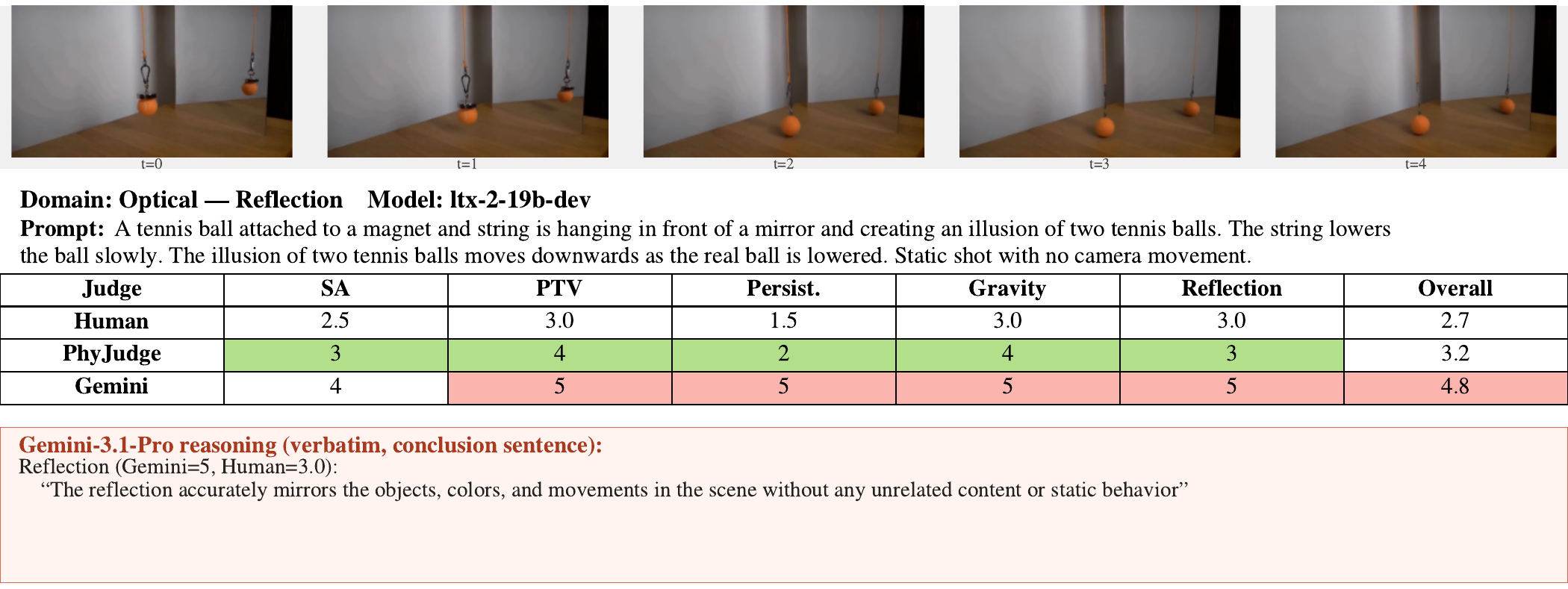}
    \caption{\textbf{Reflection} on an LTX-2-19B generation: a tennis ball is supposed to be lowered in front of a mirror, but the generated object morphs through several distinct shapes during the clip. Humans penalize this failure accordingly (3.0), with PhyJudge-9B matching the human score (3.0). In contrast, Gemini-3.1-Pro assigns a perfect score of 5.0, failing to notice that the reflected object changes identity across frames.}
    \label{fig:judge_example_reflection}
\end{figure}

\begin{figure}[htbp]
    \centering
    \includegraphics[width=\linewidth]{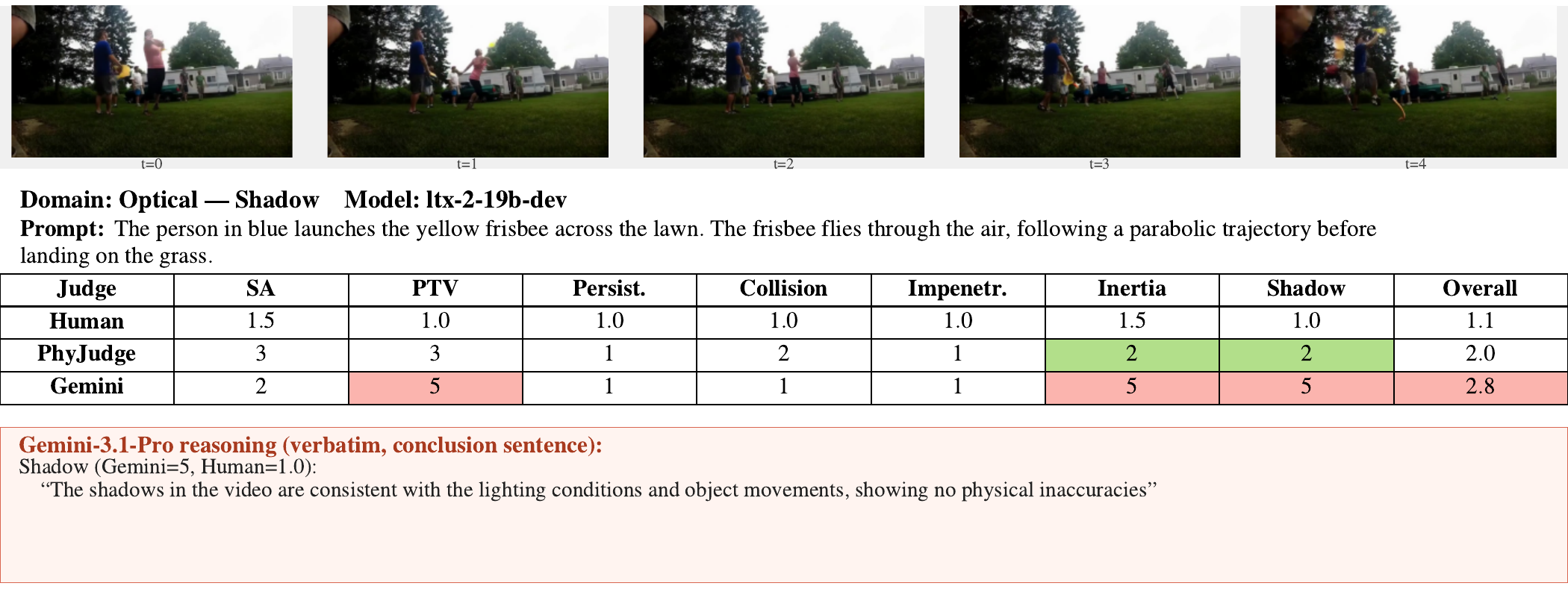}
    \caption{\textbf{Shadow} on an LTX-2-19B generation: a yellow frisbee flies across a lawn and lands on the grass. Humans rate the cast shadow inconsistent with the frisbee's trajectory (1.0); PhyJudge-9B is close (2.0), while Gemini-3.1-Pro scores 5.0, missing the optical mismatch that humans see.}
    \label{fig:judge_example_shadow}
\end{figure}

To complement the aggregate calibration numbers reported in the main text, we walk through one qualitative case per physical law in the benchmark, in \crefrange{fig:judge_example_collision}{fig:judge_example_shadow}.
Each case is selected so that PhyJudge-9B closely tracks the human rating on the focal law while Gemini-3.1-Pro disagrees by at least two rating levels, and so that the same alignment pattern also holds on the overall score---the disagreement is not driven by a single noisy axis.
For every case we show five evenly spaced frames from the generated clip, the prompt, the per-dimension score table comparing Human, PhyJudge-9B, and Gemini-3.1-Pro, and the verbatim conclusion sentence Gemini emits about the focal law. This exposes both the numerical disagreement and the closed-source judge's rationale for the assessment.
Across all 13 cases the two judges are run with the same sub-question prompt schema as in the main results: Gemini-3.1-Pro with chain-of-thought (CoT) reasoning on top of the sub-questions, and PhyJudge-9B with the score-only sub-question prompt that matches its training format.
The 13 cases span six video generators and the full law inventory across solid-body, fluid, and optical domains, illustrating that the calibration gap is broad rather than concentrated in any single source model or physics group.

\section{Judge Bias without Per-Law Sub-Questions}
\label{sec:appendix_nosubq}

We ablate the per-law sub-question scaffold by training an \textit{Ours (no SubQ)} 9B checkpoint on the same data but with a baseline template that asks for a single dimension score per video instead of giving per-law sub-questions as hint.

\input{appendix/tables/pred_error_subtable.tex}

%% file: tables/TABLE_law_breakdown.tex
\begin{table}[t]
\centering
\caption{Per-law human evaluation breakdown on PhyGround (1--5 scale, higher is better). Each column reports the per-model mean over all videos annotated for that law (or merged group). Laws with $<30$ prompts per model are merged with a semantically and empirically related law: Liq-Solid (liquid--solid interaction) = boundary\_interaction + buoyancy; Conserv.\ (fluid mass/volume conservation) = displacement + fluid\_continuity; Optical = reflection + shadow. \textbf{Bold} marks the best score in each column and \underline{underline} marks the second-best. $^\dagger$: closed-source.}
\label{tab:law_breakdown}
\resizebox{\textwidth}{!}{%
\begin{tabular}{l c c c c c c c c c c}
\toprule
& \multicolumn{6}{c}{Solid-Body} & \multicolumn{3}{c}{Fluid} & \multicolumn{1}{c}{Optical} \\
\cmidrule(lr){2-7} \cmidrule(lr){8-10} \cmidrule(lr){11-11}
Model & Gravity & Inertia & Momentum & Impen. & Collision & Material & Liq-Solid & FlowDyn & Conserv. & Optical \\
\midrule
    Wan2.2-27B-A14B & \textbf{3.23} & \underline{3.23} & \textbf{3.15} & \textbf{3.49} & \textbf{3.13} & \underline{3.17} & \underline{3.15} & 3.17 & 3.21 & \underline{3.55} \\
    Veo-3.1$^\dagger$ & \underline{3.21} & \textbf{3.32} & \underline{2.94} & \underline{3.30} & \underline{2.99} & 3.14 & \textbf{3.70} & \textbf{3.82} & \textbf{3.51} & \textbf{3.69} \\
    OmniWeaving & 2.97 & 2.90 & 2.89 & 3.15 & 2.82 & \textbf{3.18} & 3.05 & \underline{3.39} & \underline{3.38} & 3.22 \\
    Cosmos-14B & 2.77 & 2.83 & 2.75 & 3.05 & 2.69 & 2.82 & 2.92 & 2.82 & 3.14 & 3.36 \\
    LTX-2.3-22B & 2.62 & 2.56 & 2.54 & 2.66 & 2.55 & 2.77 & 3.02 & 2.93 & 3.18 & 2.98 \\
    Wan2.2-TI2V-5B & 2.66 & 2.58 & 2.46 & 2.75 & 2.47 & 2.59 & 2.71 & 2.62 & 2.76 & 2.98 \\
    Cosmos-2B & 2.50 & 2.75 & 2.28 & 2.87 & 2.36 & 2.59 & 2.62 & 2.86 & 2.84 & 3.35 \\
    LTX-2-19B & 2.55 & 2.62 & 2.35 & 2.45 & 2.42 & 2.49 & 2.72 & 3.23 & 3.23 & 2.78 \\
\bottomrule
\end{tabular}%
}
\end{table}

%% file: appendix/tables/pred_error_subtable.tex
\begin{table}[htbp]
\centering
\caption{Per-dimension aggregate relative bias (\%) using baseline prompts without per-law sub-questions. The first three columns report absolute grouped mean-bias, $E = |\bar J - \bar H| / \bar H$, computed per dimension/law and then macro-averaged. Physics is macro-averaged over domains. Overall $=$ 0.5$\cdot$Macro Avg (General) $+$ 0.5$\cdot$Physics. Lower is better; \textbf{bold} marks the column-wise minimum. The last column reports the WorldModelBench~\cite{li2025worldmodelbench} signed relative bias on the per-video overall score, $(\bar J - \bar H) / \bar H$, where positive values indicate judge overestimation and negative values indicate underestimation. For this signed column, values closer to zero are better; \textbf{bold} marks the row with the smallest $|\cdot|$. \textit{Ours (no SubQ)} denotes our 9B checkpoint trained with the no-sub-question template, ablating PhyJudge-9B's sub-question supervision.}
\label{tab:pred_error_nosubq_subtable}
\resizebox{\textwidth}{!}{%
\begin{tabular}{l c c c c}
\toprule
Model & \textbf{General (\%) $\downarrow$} & \textbf{Physics (\%) $\downarrow$} & \textbf{Overall (\%) $\downarrow$} & \textbf{Signed (\%) $\to 0$} \\
\midrule
    Claude (Opus 4.7) & 19.7 & \textbf{7.8} & 13.7 & +10.7 \\
    Gemini-3.1-Pro & 20.9 & 37.1 & 29.0 & -22.7 \\
    Qwen9B-base & 35.8 & 29.8 & 32.8 & +38.5 \\
    Qwen27B-base & 36.8 & 36.2 & 36.5 & +40.4 \\
    Ours (no SubQ) & \textbf{7.5} & 10.2 & \textbf{8.9} & \textbf{-6.6} \\
\bottomrule
\end{tabular}%
}
\end{table}

%% file: appendix/trainingdetail.tex
\section{VLM Judge Training Details}
\label{sec:appendix_judge_training}

LoRA, optimization, and compute configuration for the judge fine-tune are listed in \Cref{tab:training_config}, which also reports hardware, peak per-GPU memory, and wall-clock training time.
\input{appendix/tables/training_config_table.tex}

\subsection{Training Data}

Each training sample contains three elements:

\begin{itemize}
    \item \textbf{Video frames}: sampled at 4fps (consistent with inference), each frame resized so that the short side is 360 to fit the visual encoder; during training the cap is \texttt{FPS\_MAX\_FRAMES=12}, the per-frame token cap is \texttt{IMAGE\_MAX\_TOKEN\_NUM=1024}, and the whole-video token cap is \texttt{VIDEO\_MAX\_TOKEN\_NUM=1024}
    \item \textbf{Augmented prompt with expected outcome}: completely identical to the prompt used at benchmark evaluation time, ensuring training-inference distribution alignment
    \item \textbf{Structured score supervision}: a JSON object containing only the target evaluation key and the corresponding 1--5 human score
\end{itemize}

%% file: appendix/tables/training_config_table.tex
\begin{table}[htbp]
\centering
\small
\caption{LoRA, optimization, and compute configuration for the VLM judge. Hardware, peak per-GPU memory, and wall-clock time are reported in the Compute group.}
\label{tab:training_config}
\begin{tabular}{@{}llp{0.55\linewidth}@{}}
\toprule
Group & Parameter & Value \\
\midrule
LoRA & Adapter rank / alpha / dropout & 32 / 64 / 0.05 \\
 & Target modules & all linear layers (q, k, v, o, MLP gate/up/down) \\
 & ViT & frozen \\
 & Vision--language aligner & trainable, learning rate $2\!\times\!10^{-6}$ \\
\addlinespace[2pt]
Optimization & Optimizer & AdamW ($\beta_1{=}0.9$, $\beta_2{=}0.95$, weight decay $0.1$) \\
 & Learning rate & $1\!\times\!10^{-4}$ (LoRA), $2\!\times\!10^{-6}$ (aligner) \\
 & Schedule & cosine decay with $5\%$ linear warmup \\
 & Gradient clipping & max-norm $1.0$ \\
\addlinespace[2pt]
Batching & Per-device batch size & 1 \\
 & Gradient accumulation & 8 steps \\
 & Epochs & 1 \\
 & Maximum sequence length & 8192 tokens \\
\addlinespace[2pt]
Systems & Distributed strategy & DeepSpeed ZeRO-2 \\
 & Numerical precision & bfloat16 \\
\addlinespace[2pt]
Selection & Validation & evaluated every 100 optimizer steps; best checkpoint by validation loss \\
\addlinespace[2pt]
Compute & Hardware & 4$\times$ NVIDIA A100-SXM4-80\,GB \\
 & Peak memory & ${\sim}48$\,GiB / GPU \\
 & Wall-clock & ${\sim}1$\,h $42$\,min (294 optimizer steps, ${\sim}20.2$\,s/step) \\
 & Throughput & $1.53$ samples/s \\
\bottomrule
\end{tabular}
\end{table}

%% file: appendix/vlmevalsetting.tex
\section{Evaluator Configuration Details}
\label{sec:appendix_vlm_eval_setting}

This appendix describes the prompt structure used by the VLM scorer in the
automatic evaluation, ablates the schema choice against alternative
designs, and ablates the inference-time frame sampling rate.

\subsection{System Prompt}

\begin{quote}
You are a strict video evaluation model.
\end{quote}

\subsection{Scoring Dimensions and Prompt Structure}

Each dimension (general or physical law) is a single, independent VLM call, and the VLM directly outputs a JSON object with a 1--5 score. The sub-questions are provided to the VLM as a mental checklist (hint) before scoring, but the VLM is not required to answer each one individually.

\subsubsection{General Dimensions}

The three general dimensions (SA, PTV, Object Persistence; defined in \Cref{sec:scoring}) are scored independently. The prompt contains a textual description of the video, the reference sub-questions, and the scale anchors. The judge returns a single integer score for each dimension.

\subsection{Schema Ablation}
\label{sec:subq-schema}

\Cref{tab:gemini_per_domain} reports absolute grouped mean-bias against human labels per general dimension (SA/PTV/Persist.) and per physics domain (Solid-Body/Fluid/Optical) for Gemini-3.1-Pro, Qwen9B-base, and Qwen27B-base judges under the bare rubric, \textsc{+SubQ}, \textsc{+SubQ +CoT}, and (for Qwen) free-form \textsc{+CoT}.
\input{appendix/tables/gemini_per_domain.tex}

A central design choice in our pipeline is the prompt schema that elicits
per-law scores from a frontier VLM judge. We ablate three schemas at the
same physics-domain granularity used in the main paper: the bare rubric
(model name only), per-law sub-questions (\textsc{+SubQ}), and per-law
sub-questions with chain-of-thought (\textsc{+SubQ +CoT}). The same
schemas plus a free-form \textsc{+CoT} row are reported for the untuned
Qwen9B-base judge as a 9B-parameter open-weights reference.

Our sub-question schema yields the strongest agreement with human labels
on the physics dimensions across both judges. For Gemini-3.1-Pro, the
mean-bias drops from $39.2\%/47.9\%/27.1\%$ on Persist./Fluid/Optical
under the bare rubric to $17.3\%/7.8\%/9.4\%$ under \textsc{+SubQ};
for the open 9B reference the same direction holds on the Persist.\ row. We adopt \textsc{+SubQ} as the default judge
configuration on the basis of these numbers.

\paragraph{Why sub-questions help, and when CoT does not stack.}
The \textsc{+SubQ} schema binds the judge to specific physical violations
through narrow yes/no items rather than to a single holistic reading of
the scene. This decouples per-law scoring from any one perceptual
summary of the video and, in our ablation, yields agreement that is
both higher and more stable across repeated samples. Stacking
chain-of-thought on top is non-monotone: Optical drops further to
$1.8\%$, but Fluid moves from $7.8\%$ back up to $20.6\%$. The
free-form chain can re-introduce a holistic claim that the sub-questions
were designed to bypass. We therefore keep \textsc{+SubQ} without CoT as
the default schema and treat \textsc{+SubQ +CoT} as a configuration to
enable on a per-domain basis.

\paragraph{Design choices that make the schema ablation legible.}
The ablation is informative because the surrounding evaluation retains
four ingredients that are often dropped in aggregate VLM-as-judge
benchmarks. (i)~We report per-law and per-domain errors, so each
schema's effect is visible at law granularity rather than collapsed into
a single physical-commonsense score. (ii)~We collect per-law human
labels, so disagreement is attributed to specific laws rather than to a
whole-prompt rubric. (iii)~We evaluate VLM-as-judge behavior directly
rather than substituting pixel-level reference metrics that bypass the
judge entirely. (iv)~We compare more than one prompt schema, so the
contribution of rubric form, sub-questions, and reasoning style can each
be measured independently.

\subsection{Frame Sampling Rate}

We ablate the inference frame-sampling rate at fps $\in \{1, 2, 4, 8\}$ for three judges---PhyJudge-9B (fine-tuned), and Qwen3.5-9B-base and Qwen3.5-27B-base as off-the-shelf references (controlled by frame resampling at inference; the training-time frame upper bound is 12).
\Cref{tab:fps_pred_error} reports per-dimension aggregate relative bias and the WMBench signed bias for every (judge, fps) pair, with the best fps within each judge bolded per column.
Two qualitative findings come out of the table.
First, within each judge the choice of fps moves the General, Physics, Overall, and signed columns by only a small amount and does not produce a consistent monotone trend, so the choice is robust within the range we swept; we adopt 4fps as the main setting because it sits at or near the per-judge optimum on most columns and, relative to 8fps, halves the visual-token input and the associated inference cost.
Second, across judges PhyJudge-9B is markedly closer to human labels than either base judge at every fps, while the two base judges are roughly comparable to each other---scaling the off-the-shelf base from 9B to 27B does not close the gap to the fine-tuned 9B judge, indicating that fine-tuning on our supervision is far more impactful than base size at this scale.

\input{appendix/tables/fps_pred_error.tex}

Note: high-speed, transient physical events may still miss key details under 4fps. This limitation applies to all VLM judges.

\subsection{Evaluation Cost}
\label{sec:appendix_vlm_eval_cost}

\Cref{tab:eval_cost_summary} summarises the approximate API and compute cost of the \evalname{} automatic evaluation pipeline. The Claude Opus 4.7 chain-of-thought judge dominates the total because of its image-token pricing.

\input{appendix/tables/cost_summary.tex}

\subsection{Reproducibility.}
We release the full evaluation pipeline, including the base model, LoRA weights, evaluation code, physics sub-question definitions, training configuration, and score-supervision records. 
Reproduction does not require any closed-source API call or closed-source reasoning traces, and runs on a single GPU.
Because the released judge weights are version-pinned, evaluation results do not drift with provider-side updates.

%% file: appendix/tables/gemini_per_domain.tex
\begin{table}[htbp]
\centering
\caption{Absolute grouped mean-bias $E = |\bar J - \bar H|/\bar H$ on the human-eval set for Gemini-3.1-Pro, Qwen9B-base, and Qwen27B-base judges, broken down by general dimension (SA: Semantic Alignment; PTV: Physical Temporal Validity; Persist.: Object Persistence) and by physics domain (Solid-Body, Fluid, Optical), with each domain column sample-pooled across all law-video pairs in that domain. Prompt schemas: a single direct rubric (no SubQ), per-law sub-questions (SubQ), per-law sub-questions with chain-of-thought (SubQ + CoT), and free-form chain-of-thought without sub-questions (CoT, no SubQ). Lower is better; \textbf{bold} marks the column-wise minimum within each model block. The general (SA/PTV/Persist.) and physics (Solid-Body/Fluid/Optical) columns score independent sub-rubrics, not regroupings of the same items. From Qwen9B-base to Qwen27B-base, SA and PTV improve but Persist.\ regresses, and physics-domain bias rises across Solid-Body, Fluid, and Optical. On Gemini, SubQ reduces mean-bias on every physics domain; on Qwen, the same swap raises mean-bias on every physics domain for both 9B and 27B. On Gemini, free-form CoT alone (no SubQ) lowers bias on every physics domain over the base rubric, but layering CoT on top of SubQ is non-monotone (Solid-Body and Optical improve, Fluid regresses), whereas on Qwen, layering CoT onto either prompt schema raises mean-bias on essentially every physics domain. These domain pooled biases are not directly comparable to the Physics column in the main-text per-dimension aggregate-bias subtable, which macro-averages absolute per-law biases instead of pooling samples within a domain.}
\label{tab:gemini_per_domain}
\resizebox{\textwidth}{!}{%
\begin{tabular}{l c c c c c c}
\toprule
Model & \textbf{SA (\%) $\downarrow$} & \textbf{PTV (\%) $\downarrow$} & \textbf{Persist. (\%) $\downarrow$} & \textbf{Solid-Body (\%) $\downarrow$} & \textbf{Fluid (\%) $\downarrow$} & \textbf{Optical (\%) $\downarrow$} \\
\midrule
    Gemini-3.1-Pro & \textbf{7.4} & 15.9 & 39.2 & 40.2 & 47.9 & 27.1 \\
    \quad + CoT & 15.1 & \textbf{0.9} & 34.4 & 31.4 & 44.2 & 25.9 \\
    \quad + SubQ & 25.4 & 9.4 & \textbf{17.3} & 22.4 & \textbf{7.8} & 9.4 \\
    \quad + SubQ + CoT & 12.6 & 10.4 & 28.5 & \textbf{9.9} & 20.6 & \textbf{1.8} \\
\midrule
    Qwen9B-base & 29.1 & 44.1 & 34.0 & \textbf{47.8} & \textbf{22.8} & \textbf{20.0} \\
    \quad + CoT & \textbf{23.9} & 42.8 & 40.4 & 53.8 & 28.2 & 27.1 \\
    \quad + SubQ & 27.1 & \textbf{26.8} & \textbf{30.9} & 55.6 & 27.0 & 24.7 \\
    \quad + SubQ + CoT & 28.1 & 55.9 & 41.1 & 62.2 & 30.5 & 35.3 \\
\midrule
    Qwen27B-base & 30.7 & 39.6 & 40.1 & \textbf{48.4} & 33.3 & \textbf{27.6} \\
    \quad + CoT & \textbf{21.0} & 33.8 & 22.8 & 57.2 & \textbf{32.7} & 32.9 \\
    \quad + SubQ & 22.0 & \textbf{24.6} & \textbf{17.1} & 56.7 & 34.8 & 32.9 \\
    \quad + SubQ + CoT & 33.7 & 49.2 & 37.1 & 68.5 & 37.4 & 36.5 \\
\bottomrule
\end{tabular}%
}
\end{table}

%% file: appendix/tables/fps_pred_error.tex
\begin{table}[htbp]
\centering
\caption{FPS ablation for the PhyJudge-9B (fine-tuned, local serve), Qwen3.5-9B-base, and Qwen3.5-27B-base (no fine-tune) judges under the sub-question prompt schema. Rows vary the inference frame-sampling rate; columns report absolute grouped mean-bias $E = |\bar J - \bar H| / \bar H$. General is the macro-average of $E$ across the 3 general dimensions (SA/PTV/Persist.). Physics is the macro-average across the 3 physics domains (Solid-Body/Fluid/Optical), where each domain is itself the macro-average of its laws' $E$. Overall $=$ 0.5$\cdot$General $+$ 0.5$\cdot$Physics. The last column is the WorldModelBench~\cite{li2025worldmodelbench} signed relative bias $(\bar J - \bar H) / \bar H$ on per-video overall score; closer to 0 is better. \textbf{Bold} marks the best FPS within each judge (per column).}
\label{tab:fps_pred_error}
\resizebox{\textwidth}{!}{%
\begin{tabular}{l l c c c c}
\toprule
Judge & FPS & \textbf{General (\%) $\downarrow$} & \textbf{Physics (\%) $\downarrow$} & \textbf{Overall (\%) $\downarrow$} & \textbf{Signed (\%) $\to 0$} \\
\midrule
    PhyJudge-9B & fps=1 & 3.0 & \textbf{4.8} & 3.9 & -2.8 \\
     & fps=2 & 2.5 & 8.2 & 5.4 & -1.4 \\
     & fps=4 & \textbf{1.1} & 5.5 & \textbf{3.3} & \textbf{-1.1} \\
     & fps=8 & 3.8 & 9.1 & 6.4 & -4.9 \\
    \midrule
    Qwen3.5-9B-base & fps=1 & \textbf{27.5} & 35.5 & 31.5 & +38.1 \\
     & fps=2 & 28.4 & \textbf{31.4} & \textbf{29.9} & \textbf{+36.9} \\
     & fps=4 & 28.2 & 35.1 & 31.6 & +37.7 \\
     & fps=8 & 30.0 & 31.8 & 30.9 & +37.8 \\
    \midrule
    Qwen3.5-27B-base & fps=1 & 21.5 & 41.2 & 31.3 & +36.5 \\
     & fps=2 & 23.2 & \textbf{40.7} & 32.0 & +36.1 \\
     & fps=4 & \textbf{21.4} & 40.8 & \textbf{31.1} & \textbf{+35.7} \\
     & fps=8 & 24.0 & 41.0 & 32.5 & +38.0 \\
\bottomrule
\end{tabular}%
}
\end{table}

%% file: appendix/tables/cost_summary.tex
\begin{table}[htbp]
\centering
\caption{Approximate cost of running the \evalname{} automatic evaluation pipeline. \textit{VLM judges} are the per-video scoring calls made by cloud judges; \textit{video generation} is the cost of producing reference outputs. Numbers are estimates from token counts and call volumes.}
\label{tab:eval_cost_summary}
\begin{tabular}{l r}
\toprule
Item & Cost (USD) \\
\midrule
\textit{VLM judges} & \\
\quad Gemini 3.1 Pro Preview & $\sim$420 \\
\quad Claude Opus 4.7 & $\sim$972 \\
\textit{Video generation} & \\
\quad Veo-3.1 & $\sim$100 \\
\midrule
\textbf{Total} & \textbf{$\sim$1,492} \\
\bottomrule
\end{tabular}
\end{table}

%% file: appendix/videogen.tex
\section{Video Generation Inference Settings}
\label{sec:appendix_videogen}

The inference settings of the eight video generation models in our
benchmark are listed in \Cref{tab:inference_settings}. We adopt a
single ``upsampled caption'' protocol so that text-conditioning quality
is held constant across the benchmark and any score gaps reflect
generator capability rather than prompt richness: the original dataset
captions are rewritten by Gemini 2.5 Flash to make implicit physical
phenomena, object motion, and lighting cues explicit, and the same
rewritten prompt is sent to every model. For Veo-3.1, FPS, clip length,
frame count, and resolution are recovered by ffprobing the returned
video files; the remaining knobs are unavailable through the API.

\input{appendix/tables/videogen_setting.tex}

The eight models split cleanly along two axes orthogonal to the
per-row settings already in the table. By scheduler family, we distinguish
FlowMatchEulerDiscrete (Cosmos-Predict2.5,
LTX-2-19B, LTX-2.3-22B, OmniWeaving) and FlowUniPCMultistep
(Wan2.2-TI2V-5B, Wan2.2-27B-A14B). By output sizing, three models
write a fixed resolution regardless of input (
Cosmos-Predict2.5, Veo-3.1) while the remaining five resize to track
the input aspect ratio (LTX-2-19B, LTX-2.3-22B, OmniWeaving,
Wan2.2-TI2V-5B, Wan2.2-27B-A14B). For the LTX-2 family we explicitly call
\texttt{TI2VidOneStagePipeline}, because generating the full benchmark
under the recommended two-stage (Stage-1 base + Stage-2 distilled-LoRA
refiner) configuration exceeded our video-generation compute budget. Our
LTX-2 numbers therefore correspond to a deliberately weaker single-stage
baseline rather than the configuration recommended by the model authors,
and should be read as a lower bound on what the LTX-2 family can achieve
on this benchmark.

%% file: appendix/tables/videogen_setting.tex
\begin{table}[htbp]
\centering
\caption{Inference settings for every video model in the benchmark. FPS, length, frames, classifier-free guidance scale (CFG), and number of sampling steps follow each model's released defaults; Veo-3.1 is a closed API and exposes none of these knobs (entries marked --). For models that adapt to input aspect ratio, the listed resolution is the most common output we observed across our datasets. For LTX-2-19B and LTX-2.3-22B we run only the non-distilled Stage-1 pass and do not apply the official Stage-2 distilled-LoRA refinement. All open models run in \texttt{bf16}; the Veo-3.1 endpoint does not expose its numerical precision. All models are conditioned on the same physics-enhanced prompts produced by Gemini 2.5 Flash from the original dataset captions.}
\label{tab:inference_settings}
\resizebox{\textwidth}{!}{%
\begin{tabular}{l r r r r r r l}
\toprule
Model & FPS & Length (s) & Resolution & Frames & CFG & Steps & Scheduler \\
\midrule
Cosmos-Predict2.5-2B & 16 & 5.8 & 1280x704 & 93 & 7 & 36 & FlowMatchEulerDiscreteScheduler \\
Cosmos-Predict2.5-14B & 16 & 5.8 & 1280x704 & 93 & 7 & 36 & FlowMatchEulerDiscreteScheduler \\
LTX-2-19B & 25 & 3.9 & 832x448 & 97 & 3 & 30 & FlowMatchEulerDiscreteScheduler \\
LTX-2.3-22B & 25 & 3.9 & 832x448 & 97 & 3 & 30 & FlowMatchEulerDiscreteScheduler \\
OmniWeaving & 16 & 5.1 & 848x480 & 81 & 6 & 50 & FlowMatchEulerDiscreteScheduler \\
Wan2.2-TI2V-5B & 16 & 5.1 & 1248x704 & 81 & 5.5 & 50 & FlowUniPCMultistepScheduler \\
Wan2.2-27B-A14B & 16 & 5.1 & 832x464 & 81 & 3.5 & 40 & FlowUniPCMultistepScheduler \\
Veo-3.1 & 24 & 8.0 & 1280x720 & 192 & -- & -- & -- \\
\bottomrule
\end{tabular}%
}
\end{table}

%% file: appendix/limitation.tex
\section{Limitations}
\label{sec:appendix_limitation}

\subsection{Prompt Suite Size}

PhyGround contains 250 curated prompts, which is considerably smaller than other video benchmarks. We deliberately did not scale the prompt suite further, for two reasons.
First, we spent substantial effort checking each prompt and its conditioning first frame to ensure an unambiguous expected physical outcome. This curation trades raw scale for higher per-sample evaluability and enables reliable law-level diagnostic scoring.
Second, contemporary video generation models have substantial inference costs. We therefore keep the suite at a size that is practical for both generation and multi-annotator scoring.

\subsection{Physics Law Coverage}

PhyGround organizes its evaluation around a thirteen-law taxonomy. The taxonomy is nevertheless finite by design: each included law was required to be (i) visually verifiable from a short generated video and (ii) decomposable into 2-3 observable sub-questions.
A number of physics categories, including thermodynamic phase transitions and combustion, electromagnetism, chemistry, and waves, energy, and modern physics, were therefore deliberately left out of the current taxonomy. We discuss the reasoning behind each of these intentional exclusions in \Cref{sec:appendix_law_selection}. 

%% file: appendix/poorvideobylaw.tex
\section{Qualitative Examples of Physical Law Violations}
\label{sec:appendix_poorvideobylaw}

\begin{figure}[htbp]
    \centering
    \includegraphics[width=\linewidth]{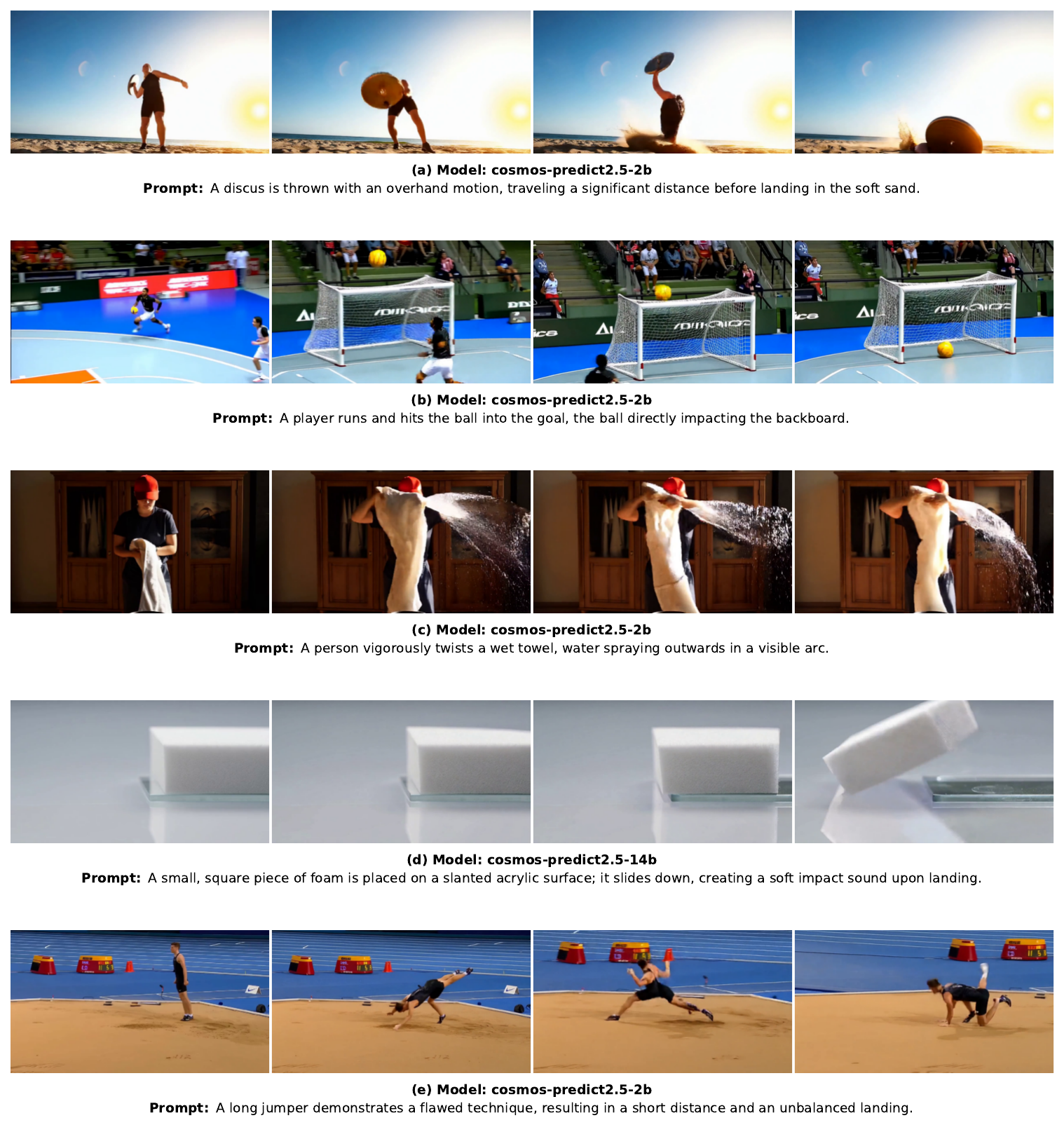}
    \caption{Qualitative examples of \textbf{gravity} violations: unsupported objects fail to fall, hover in mid-air, or accelerate inconsistently with free fall.}
    \label{fig:poor_video_gravity}
\end{figure}

\begin{figure}[htbp]
    \centering
    \includegraphics[width=\linewidth]{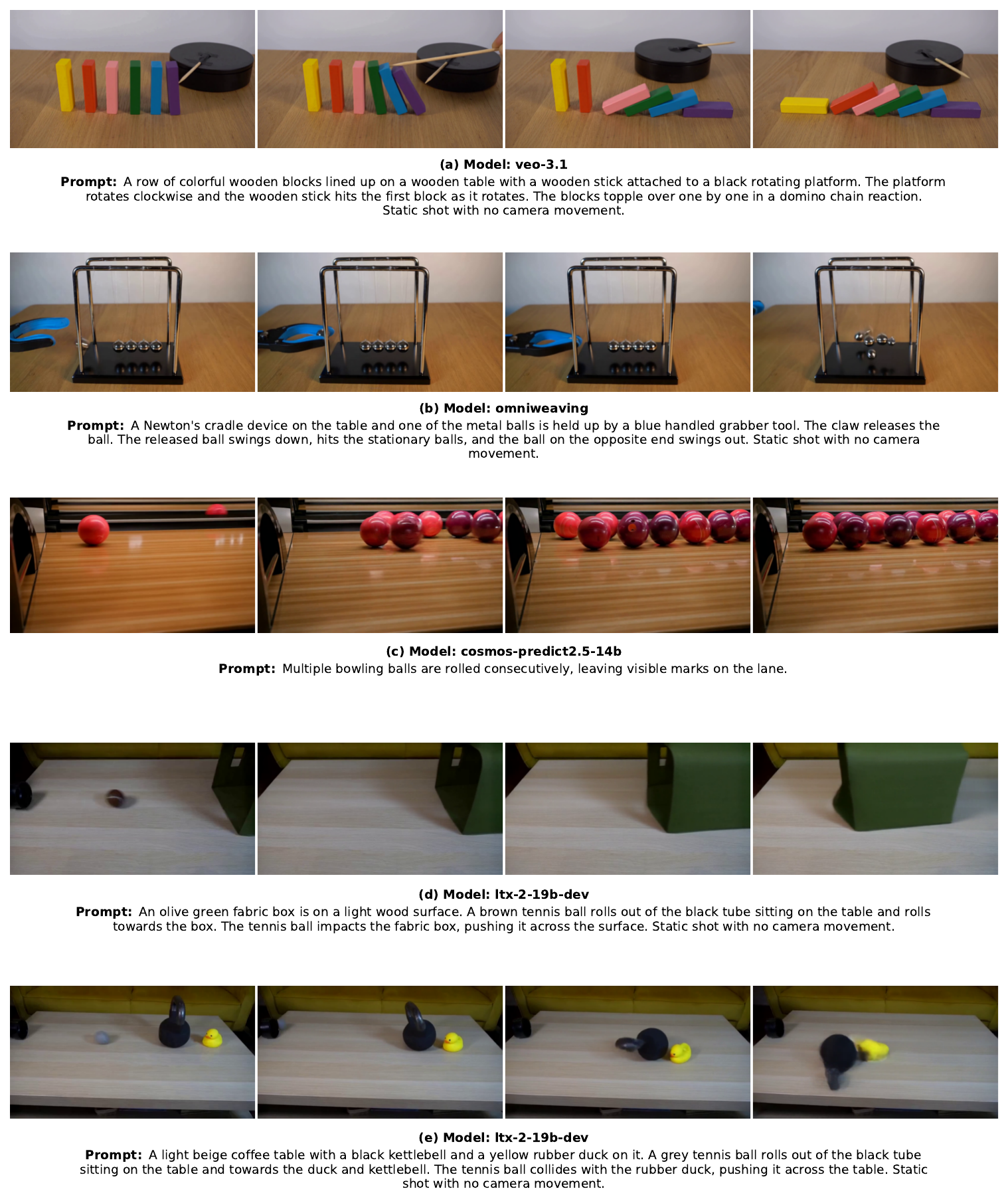}
    \caption{Qualitative examples of \textbf{inertia} violations: stationary objects spontaneously start moving, or moving objects abruptly stop without any external force.}
    \label{fig:poor_video_inertia}
\end{figure}

\begin{figure}[htbp]
    \centering
    \includegraphics[width=\linewidth]{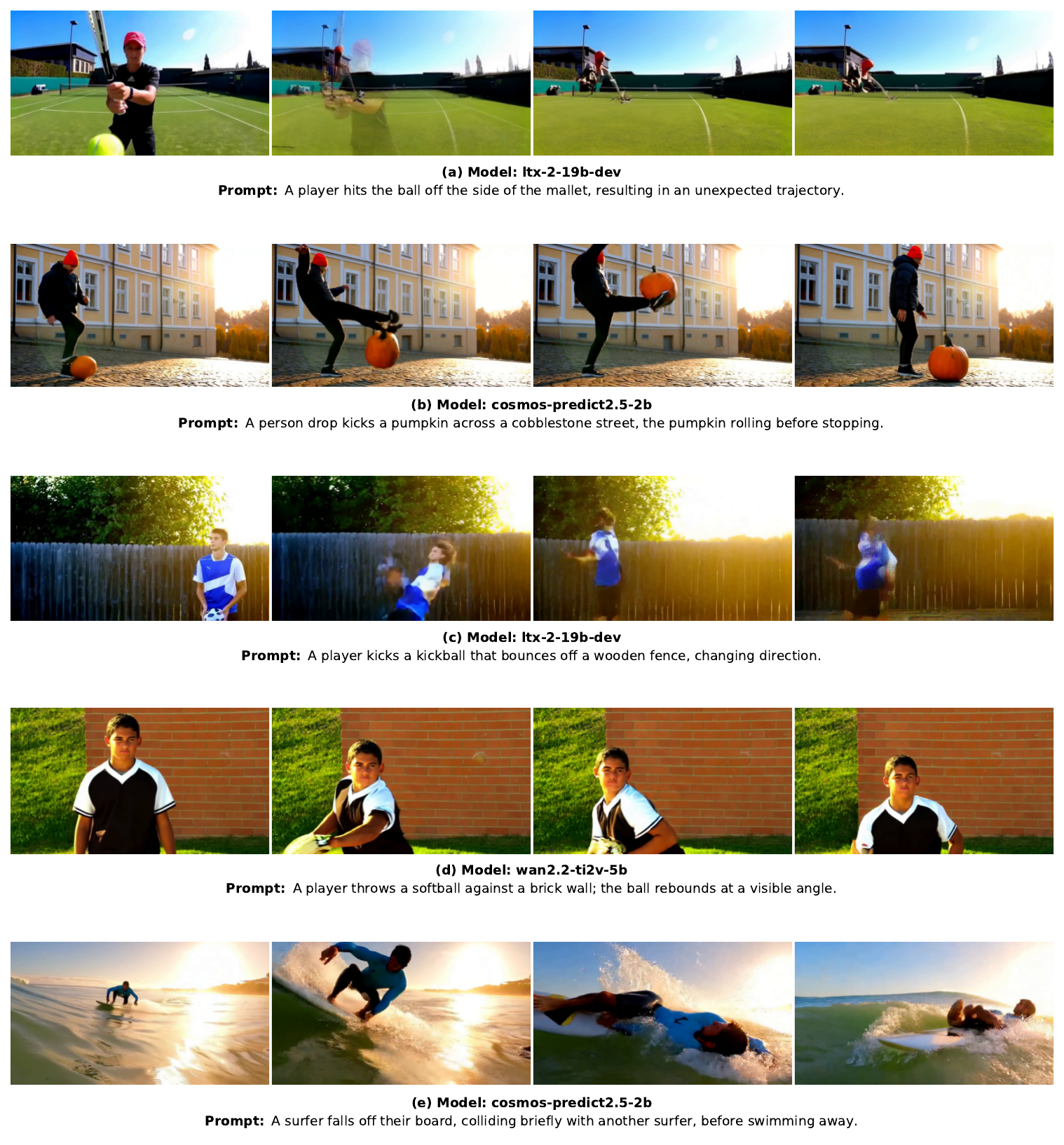}
    \caption{Qualitative examples of \textbf{momentum} violations: post-collision motion directions or speeds are inconsistent with the incoming momentum.}
    \label{fig:poor_video_momentum}
\end{figure}

\begin{figure}[htbp]
    \centering
    \includegraphics[width=\linewidth]{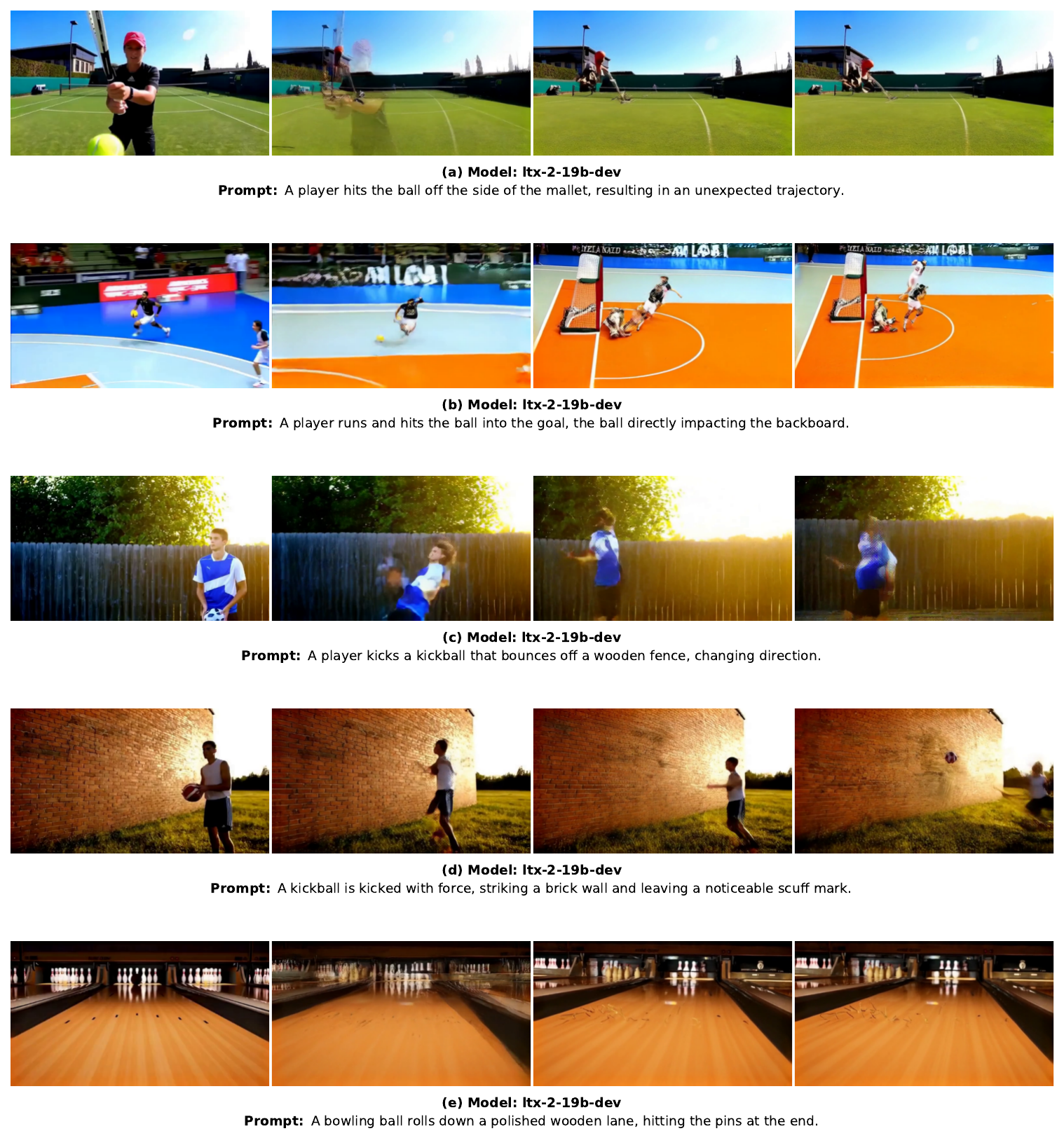}
    \caption{Qualitative examples of \textbf{impenetrability} violations: solid objects pass through one another or merge into the same region of space.}
    \label{fig:poor_video_impenetrability}
\end{figure}

\begin{figure}[htbp]
    \centering
    \includegraphics[width=\linewidth]{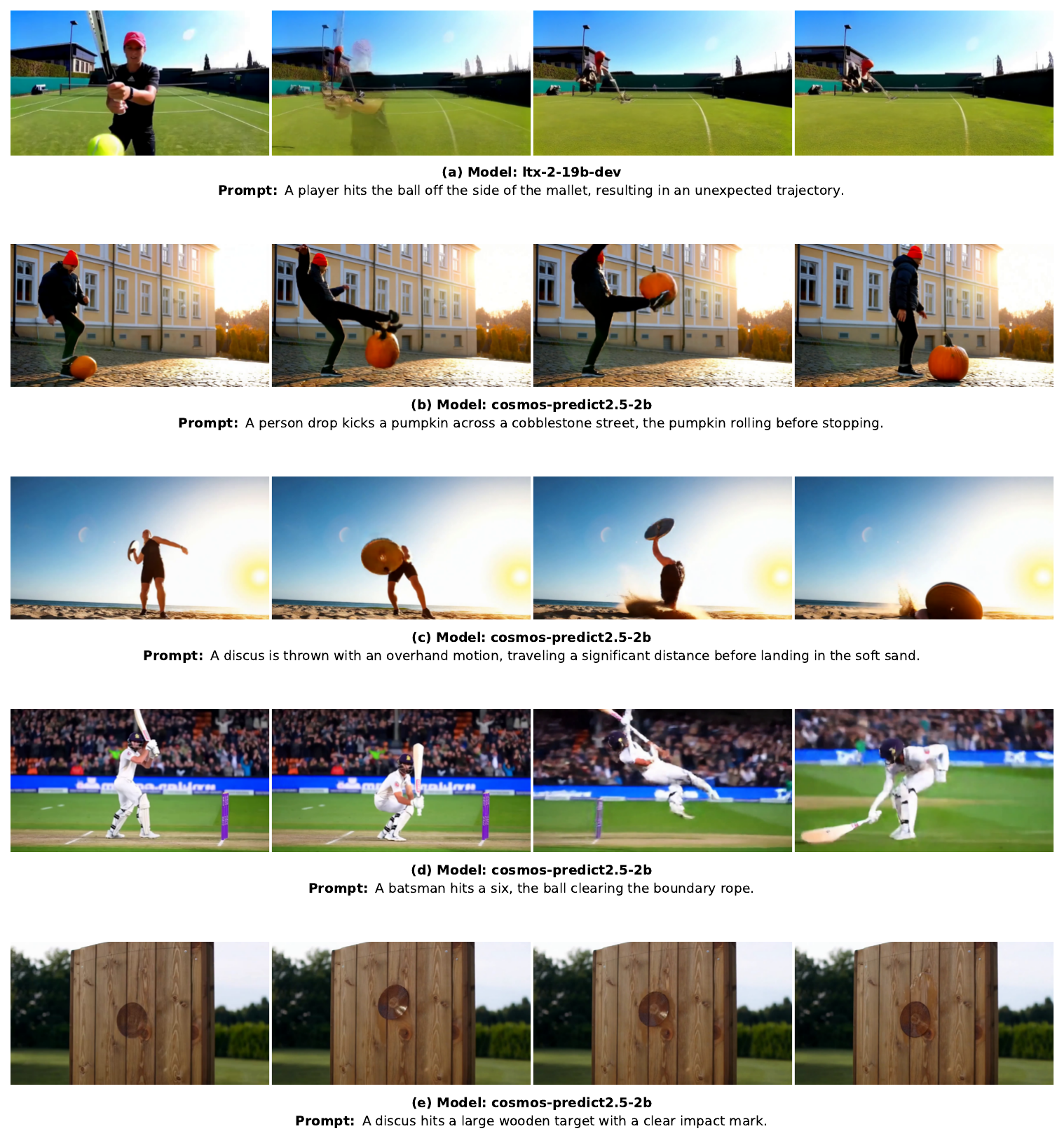}
    \caption{Qualitative examples of \textbf{collision} violations: objects fail to rebound, fracture, or deform appropriately upon impact.}
    \label{fig:poor_video_collision}
\end{figure}

\begin{figure}[htbp]
    \centering
    \includegraphics[width=\linewidth]{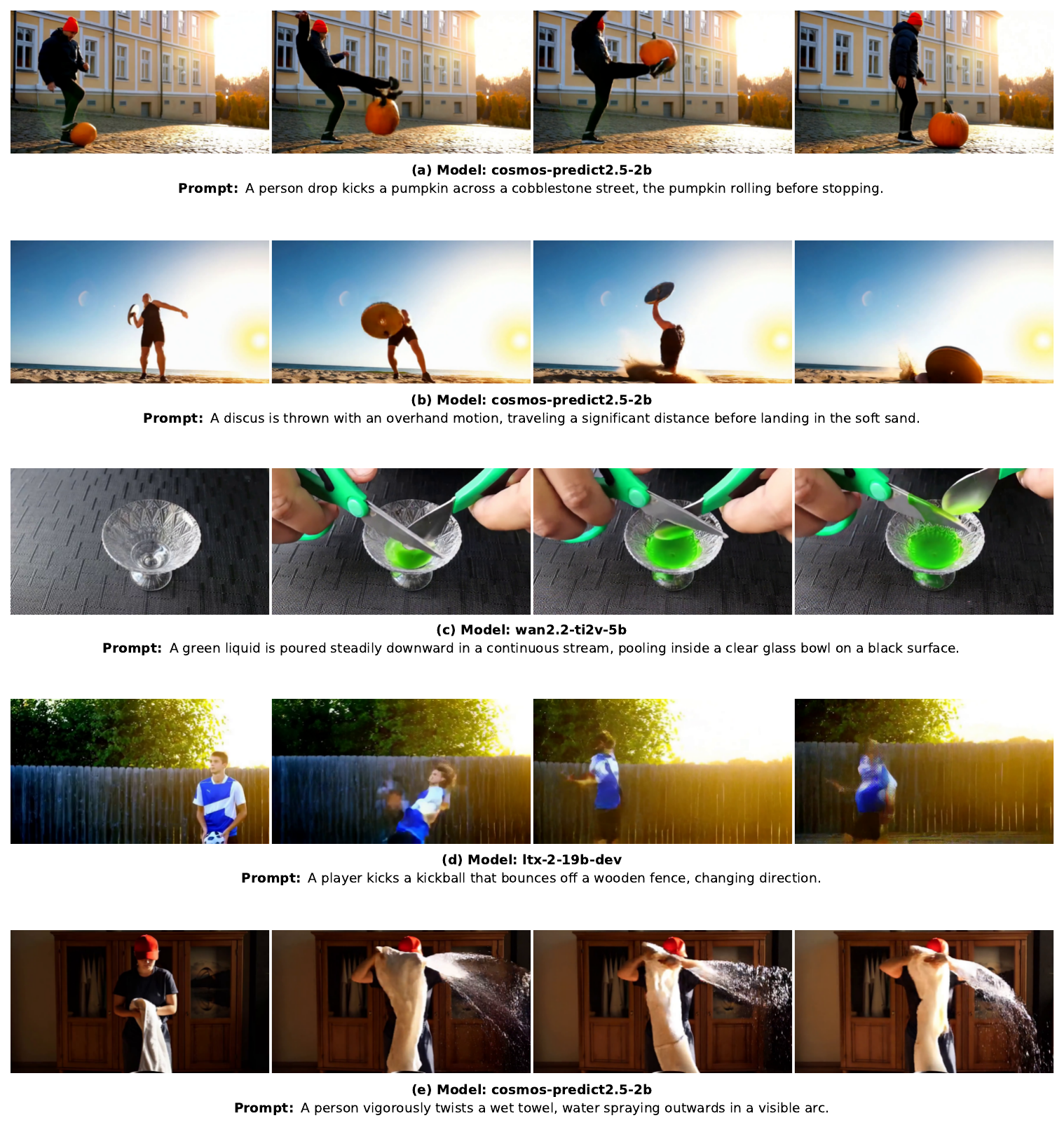}
    \caption{Qualitative examples of \textbf{material} violations: material-specific responses do not match physical properties (e.g., glass that does not shatter, rubber that does not bounce).}
    \label{fig:poor_video_material}
\end{figure}

\begin{figure}[htbp]
    \centering
    \includegraphics[width=\linewidth]{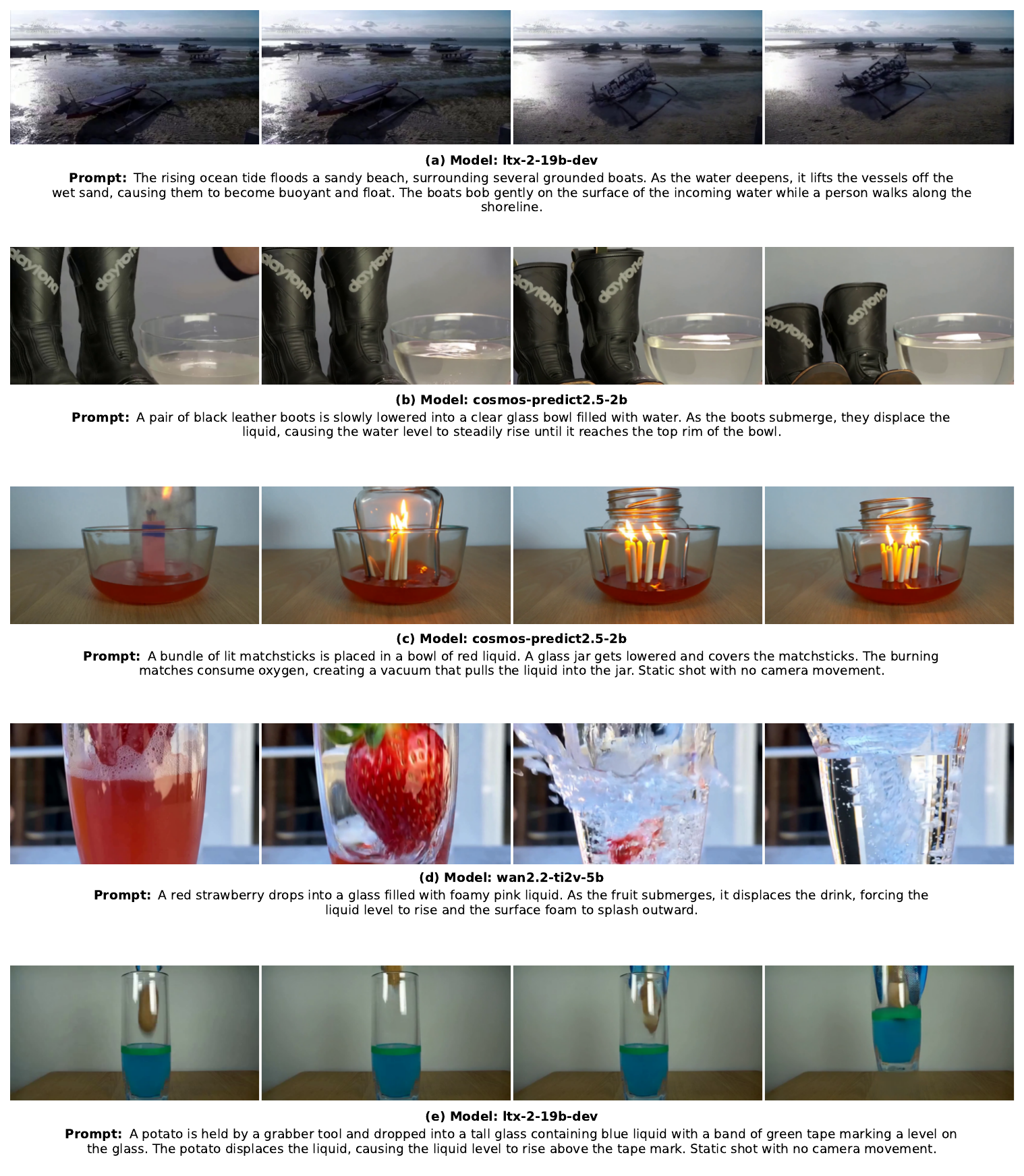}
    \caption{Qualitative examples of \textbf{buoyancy} violations: dense objects fail to sink or lightweight objects fail to float as expected.}
    \label{fig:poor_video_buoyancy}
\end{figure}

\begin{figure}[htbp]
    \centering
    \includegraphics[width=\linewidth]{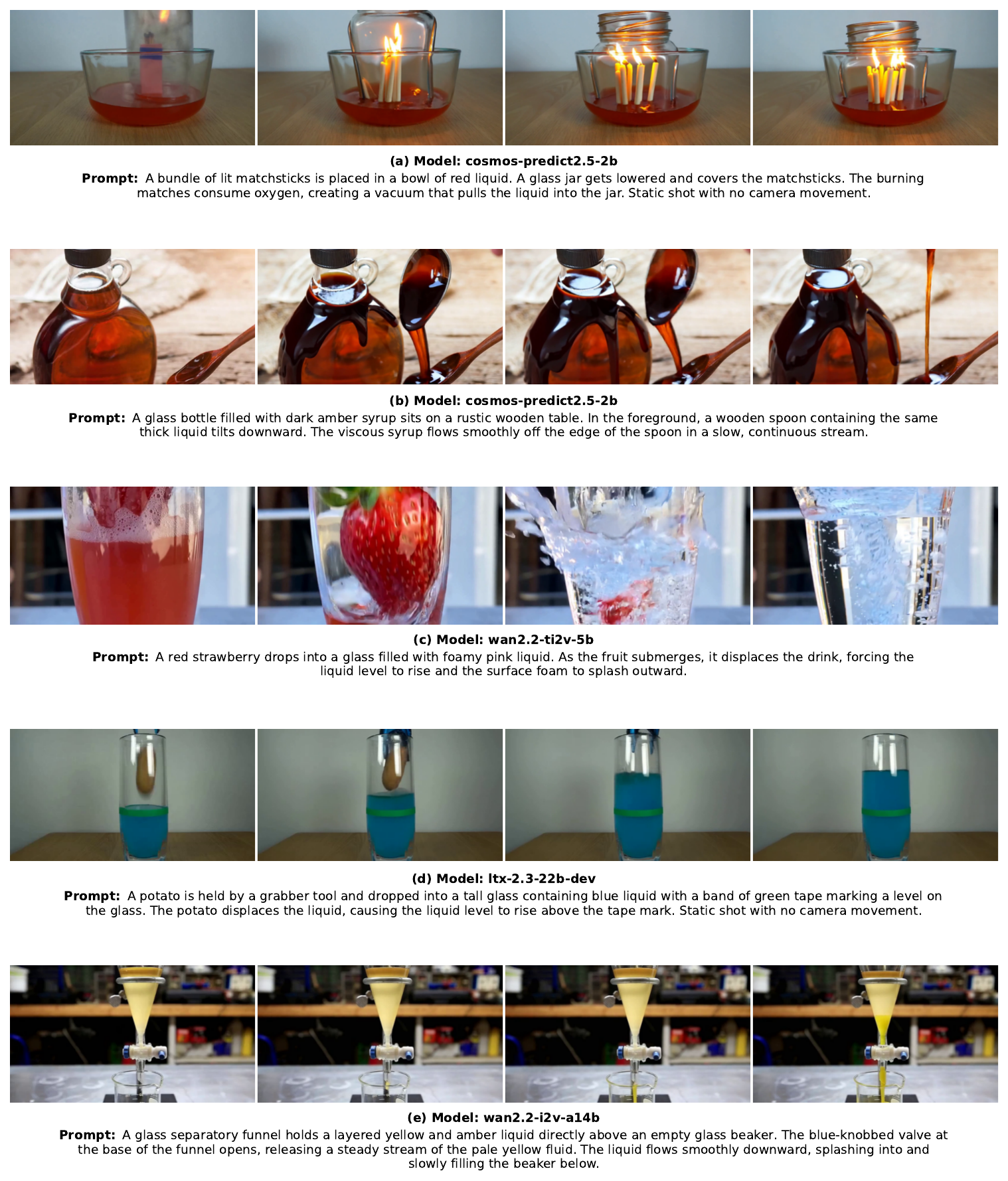}
    \caption{Qualitative examples of \textbf{displacement} violations: liquid level fails to rise upon submersion, or containers do not overflow when full.}
    \label{fig:poor_video_displacement}
\end{figure}

\begin{figure}[htbp]
    \centering
    \includegraphics[width=\linewidth]{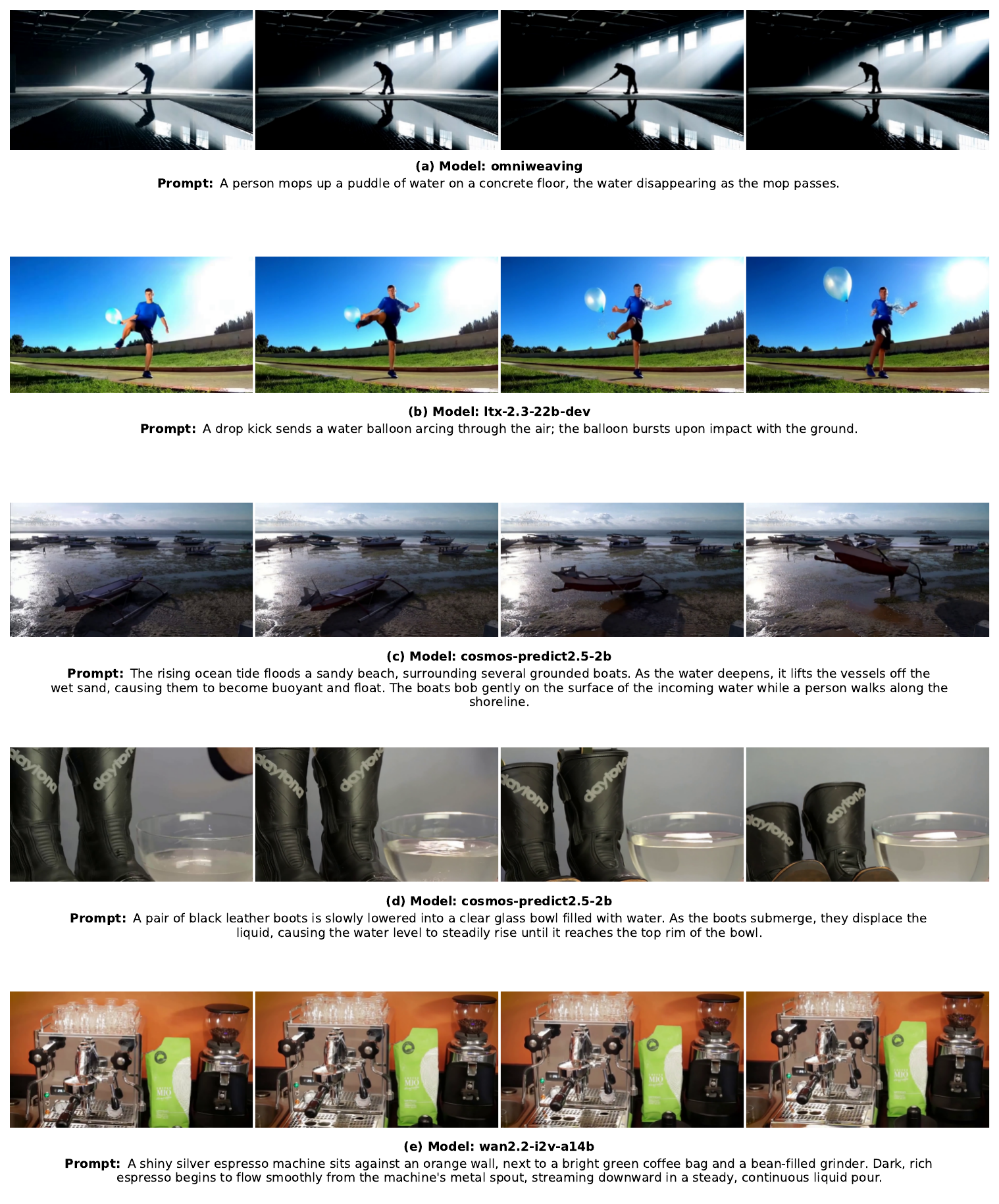}
    \caption{Qualitative examples of \textbf{boundary interaction} violations: liquid does not splash, deflect, or split realistically upon hitting obstacles.}
    \label{fig:poor_video_boundary_interaction}
\end{figure}

\begin{figure}[htbp]
    \centering
    \includegraphics[width=\linewidth]{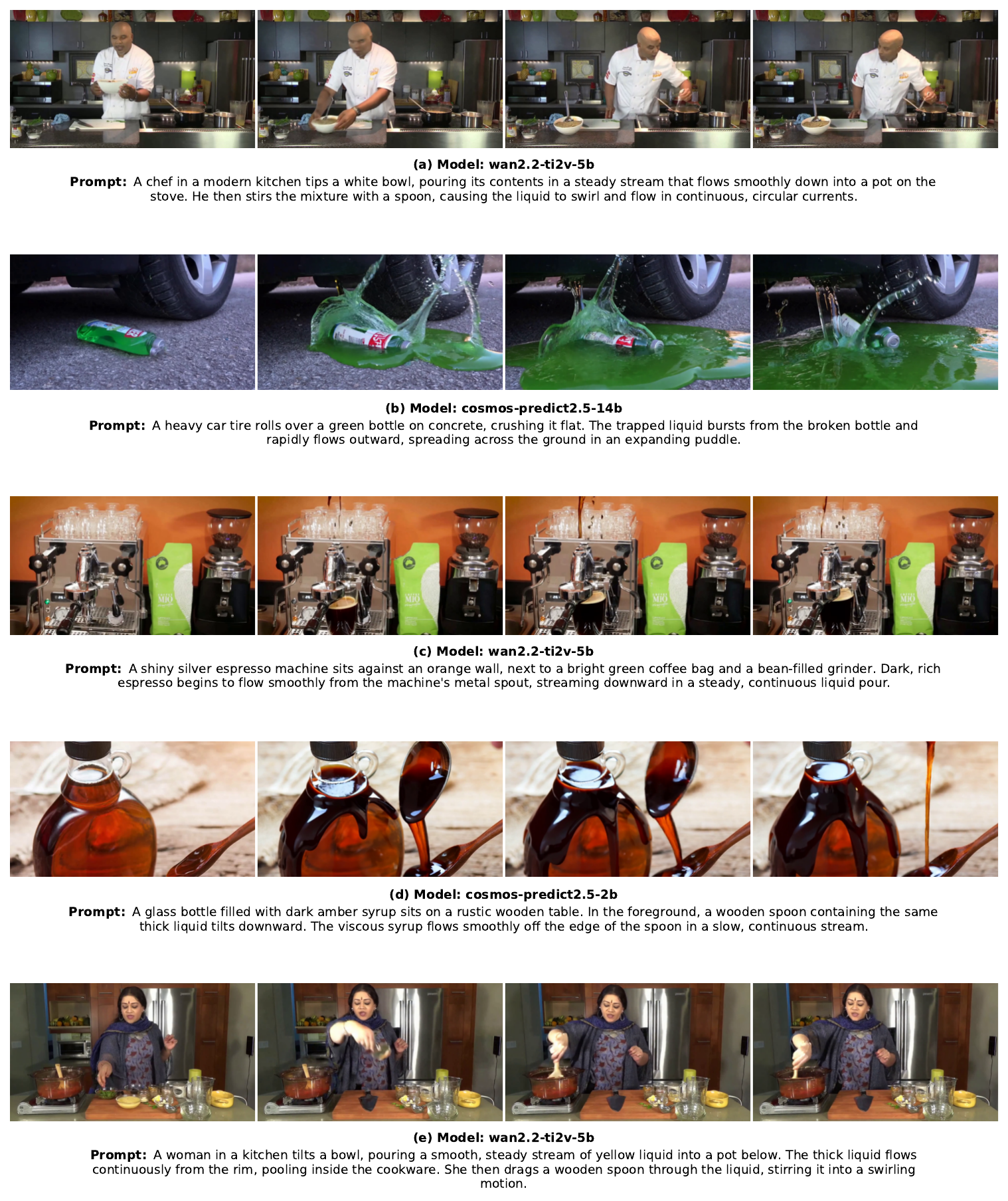}
    \caption{Qualitative examples of \textbf{fluid continuity} violations: liquid loses mass, develops spontaneous gaps, or appears out of nowhere.}
    \label{fig:poor_video_fluid_continuity}
\end{figure}

\begin{figure}[htbp]
    \centering
    \includegraphics[width=\linewidth]{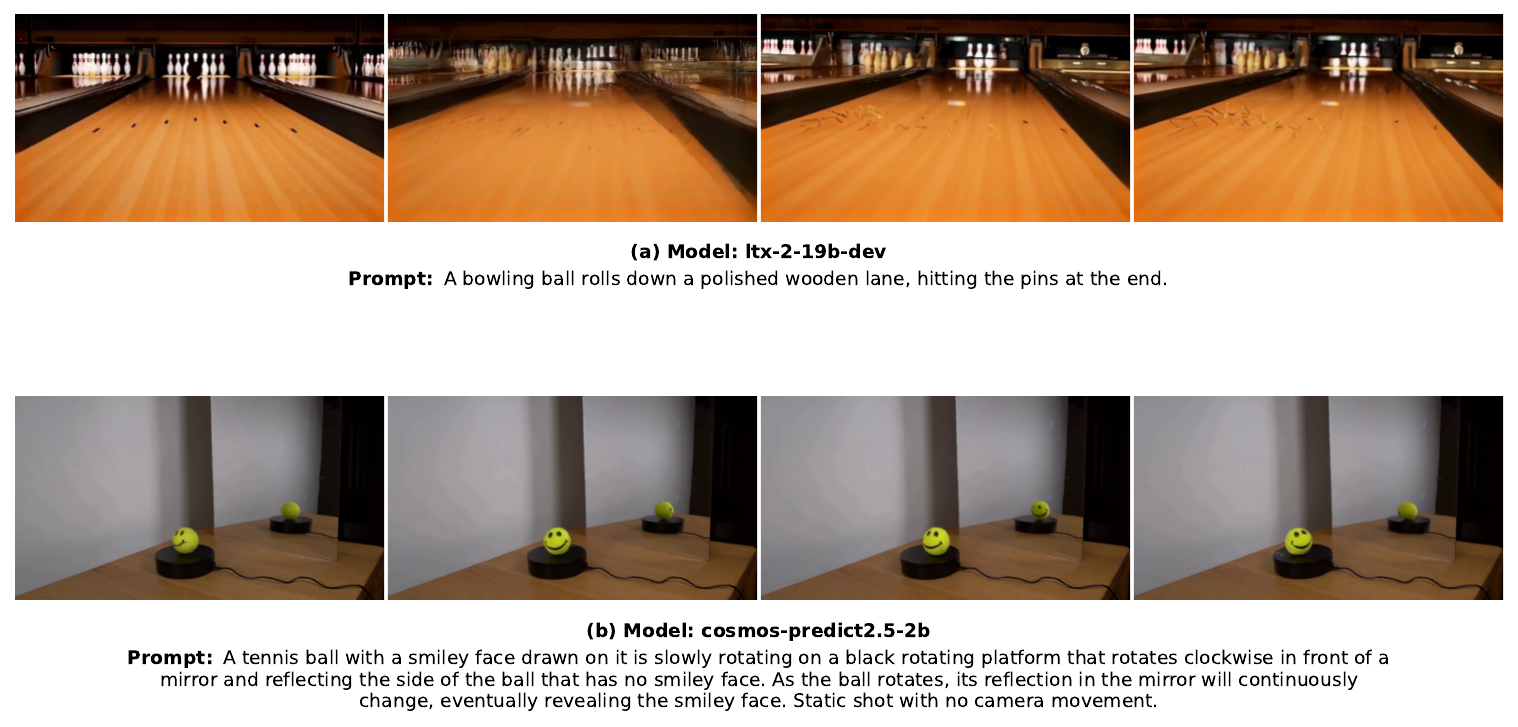}
    \caption{Qualitative examples of \textbf{reflection} violations: reflected content does not approximately match the scene in front of the reflective surface.}
    \label{fig:poor_video_reflection}
\end{figure}

\begin{figure}[htbp]
    \centering
    \includegraphics[width=\linewidth]{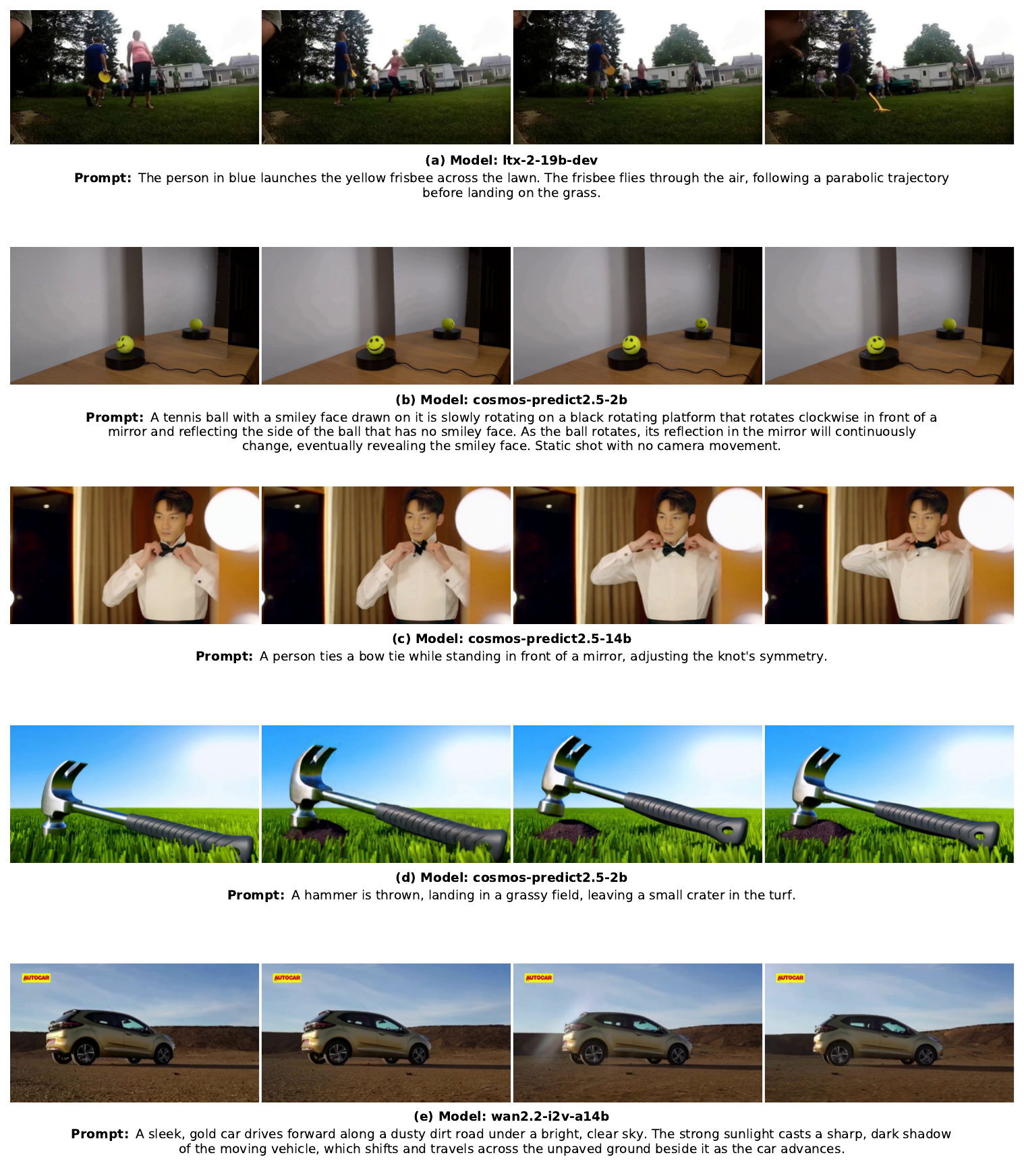}
    \caption{Qualitative examples of \textbf{shadow} violations: shadow directions are inconsistent with the light source, or shadows fail to track the casting object's motion.}
    \label{fig:poor_video_shadow}
\end{figure}

We present a few qualitative examples highlighting instances where specific physical laws are violated in \crefrange{fig:poor_video_gravity}{fig:poor_video_shadow}.

\begin{figure}[htbp]
    \centering
    \includegraphics[width=\linewidth]{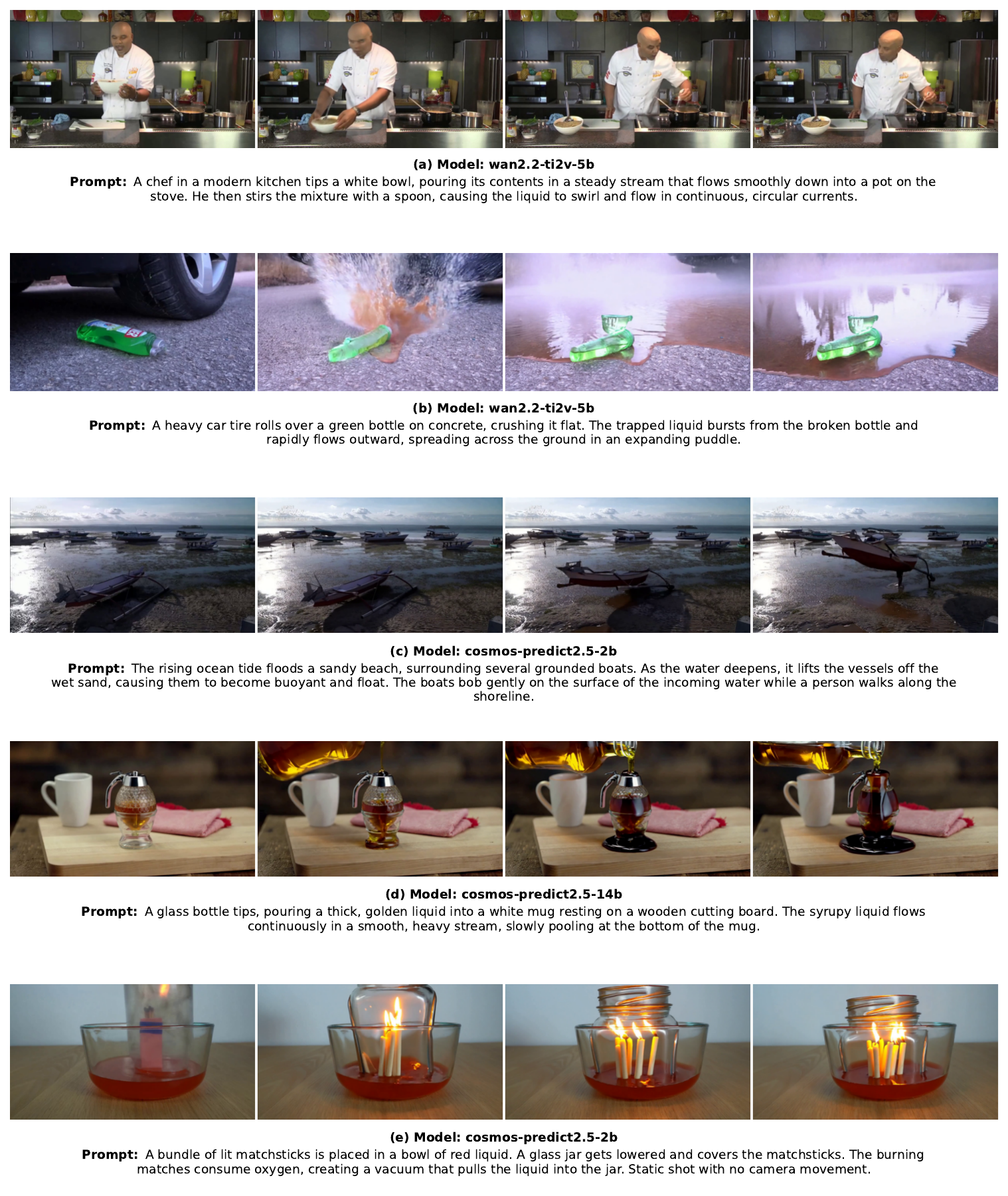}
    \caption{Qualitative examples of \textbf{flow dynamics} violations: bulk liquid motion patterns (spreading, converging, draining) appear visually unrealistic.}
    \label{fig:poor_video_flow_dynamics}
\end{figure}

%% file: appendix/poorvideobymodel.tex
\section{Qualitative Examples of Physical Law Violations by Model}
\label{sec:appendix_poorvideobymodel}

\begin{figure}[htbp]
    \centering
    \includegraphics[width=\linewidth]{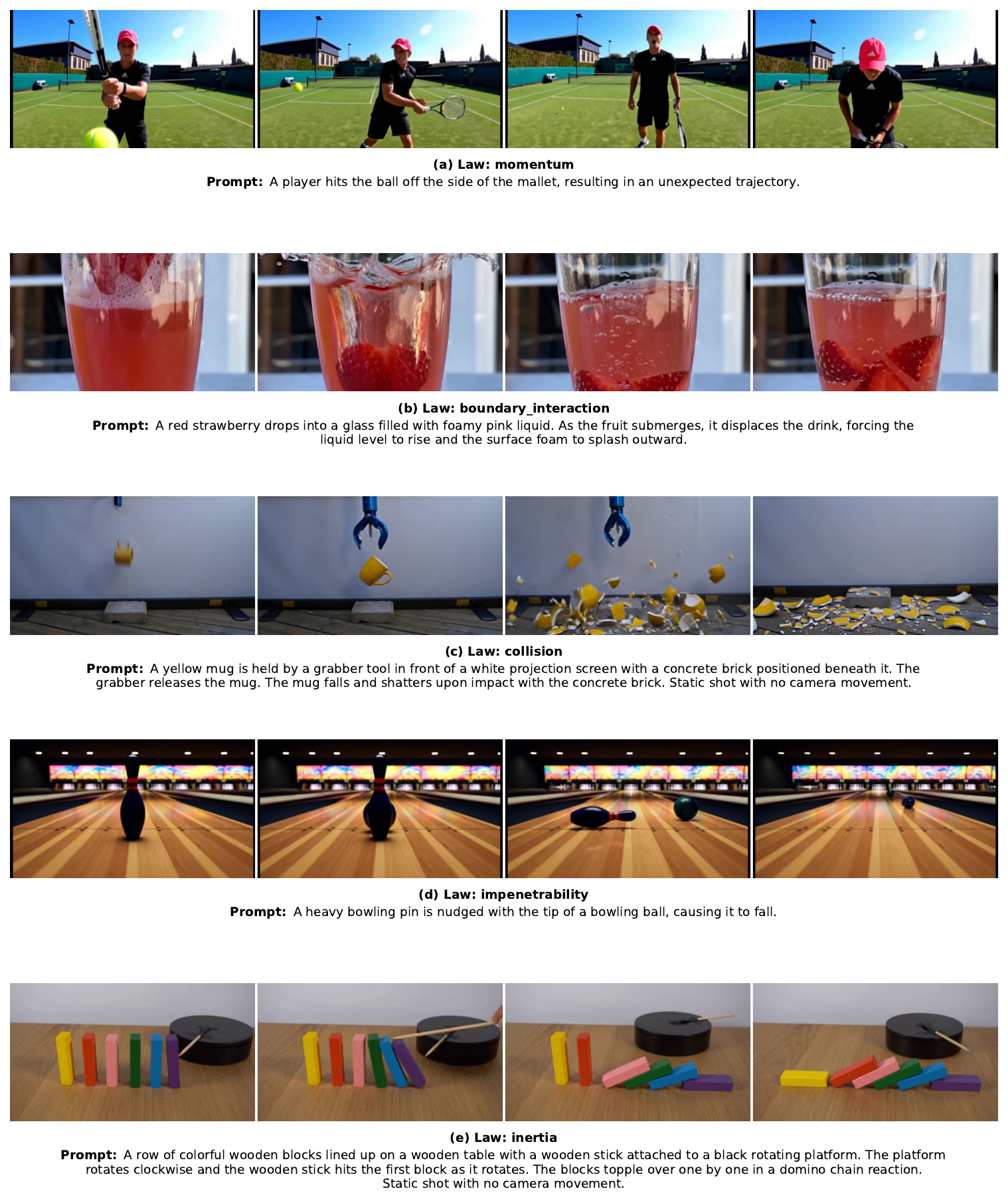}
    \caption{Qualitative examples of physical law violations from \textbf{Veo-3.1}.}
    \label{fig:poor_video_model_veo}
\end{figure}

\begin{figure}[htbp]
    \centering
    \includegraphics[width=\linewidth]{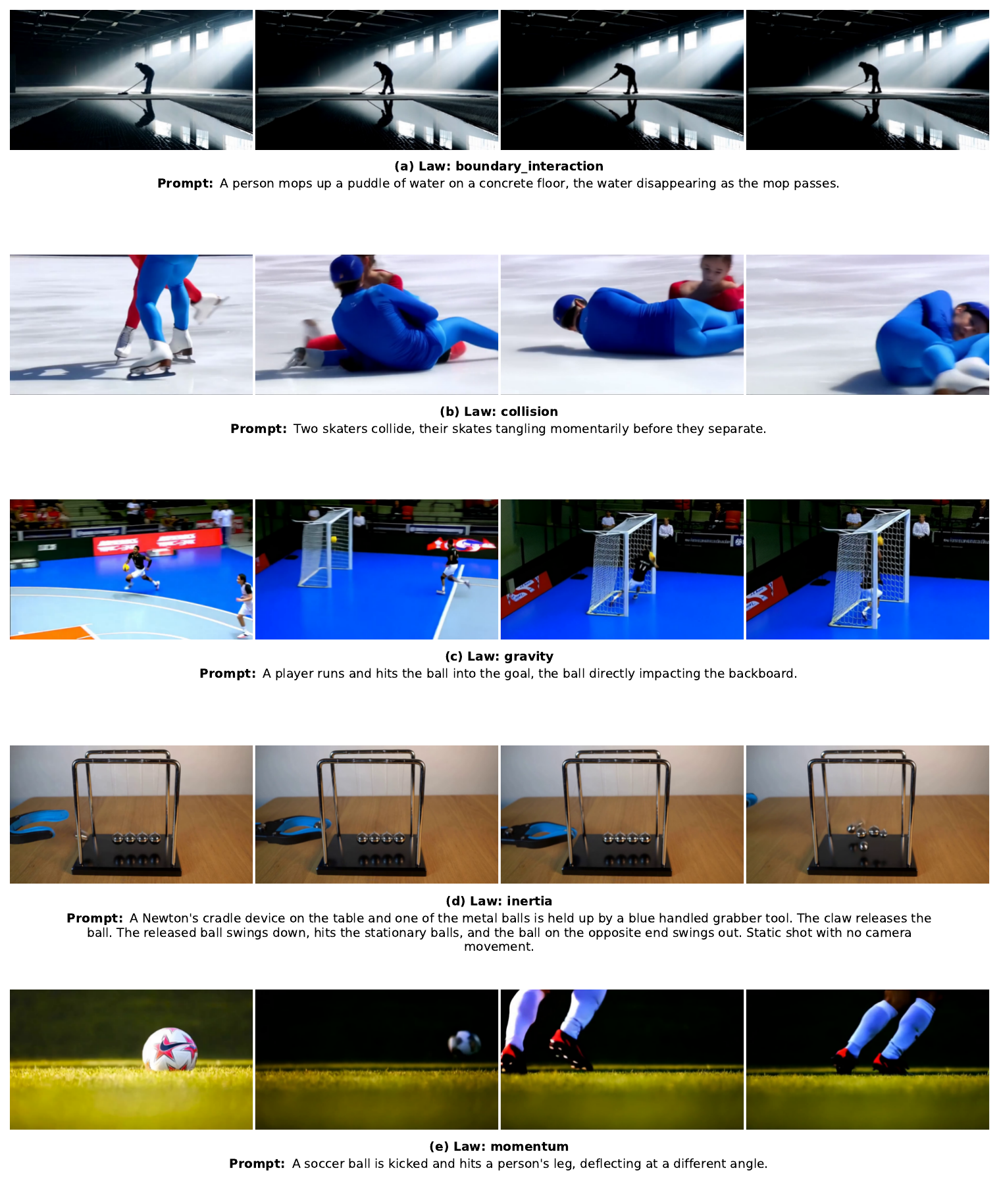}
    \caption{Qualitative examples of physical law violations from \textbf{OmniWeaving}.}
    \label{fig:poor_video_model_omniweaving}
\end{figure}

\begin{figure}[htbp]
    \centering
    \includegraphics[width=\linewidth]{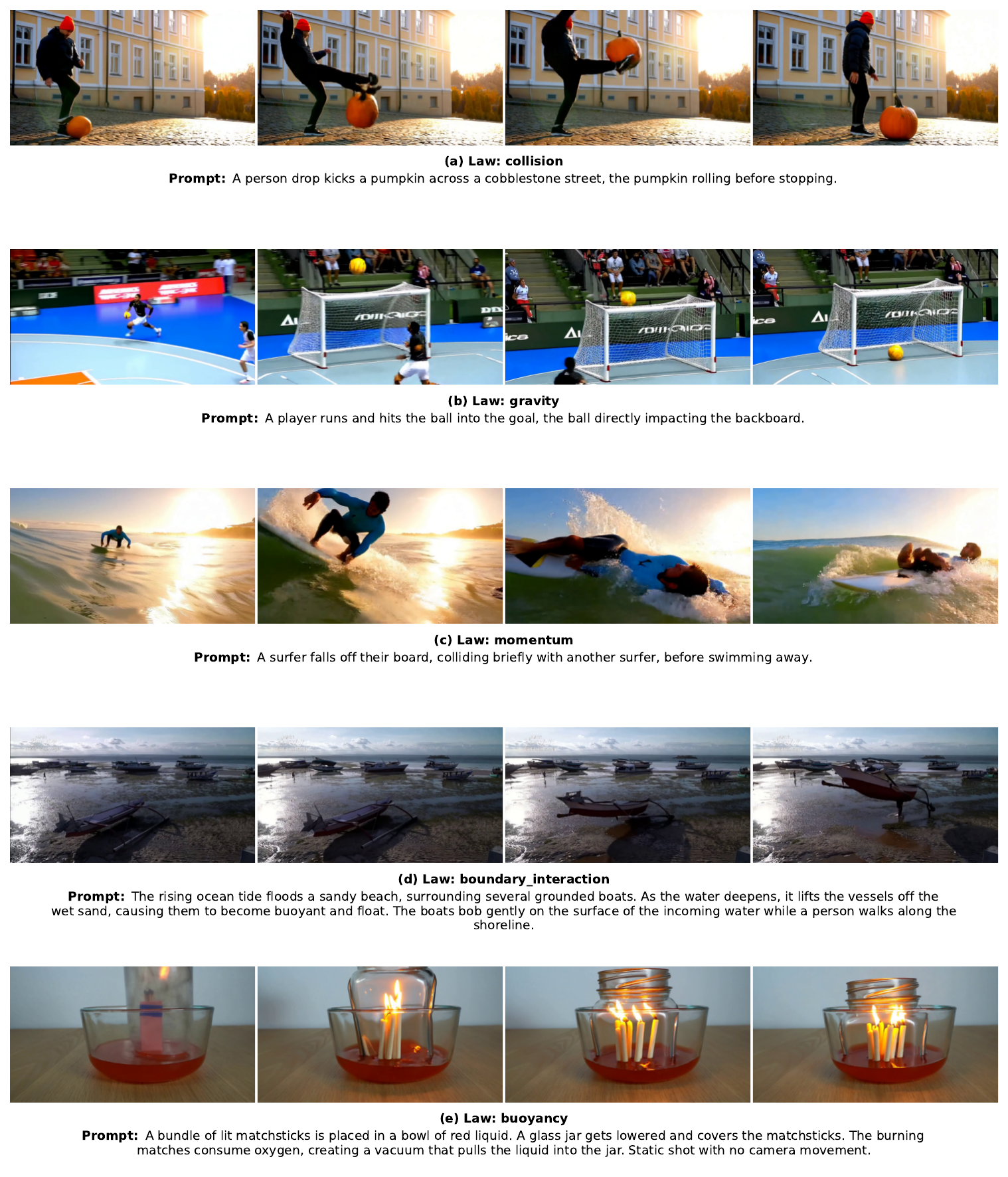}
    \caption{Qualitative examples of physical law violations from \textbf{Cosmos-2B}.}
    \label{fig:poor_video_model_cosmos_2b}
\end{figure}

\begin{figure}[htbp]
    \centering
    \includegraphics[width=\linewidth]{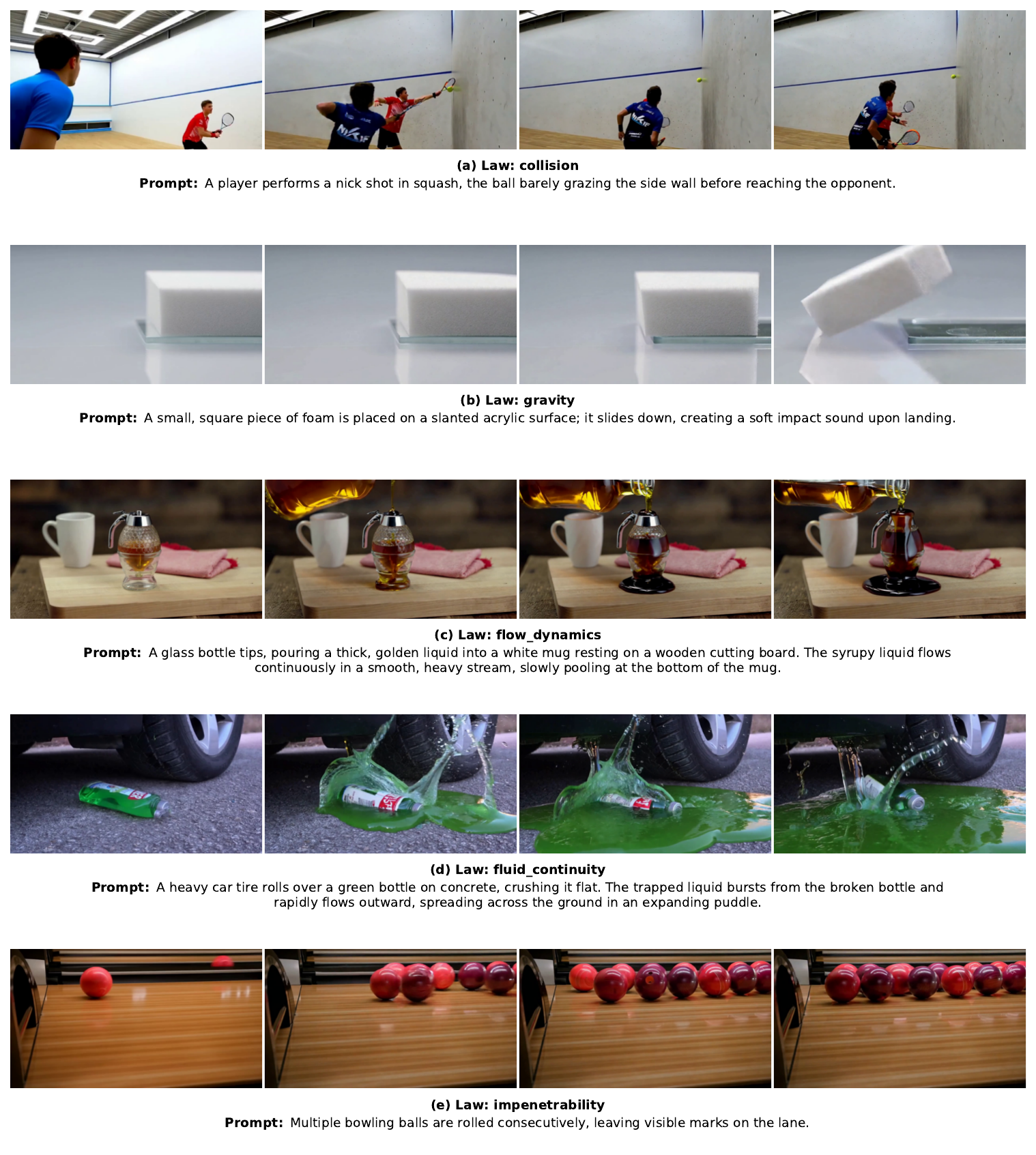}
    \caption{Qualitative examples of physical law violations from \textbf{Cosmos-14B}.}
    \label{fig:poor_video_model_cosmos_14b}
\end{figure}

\begin{figure}[htbp]
    \centering
    \includegraphics[width=\linewidth]{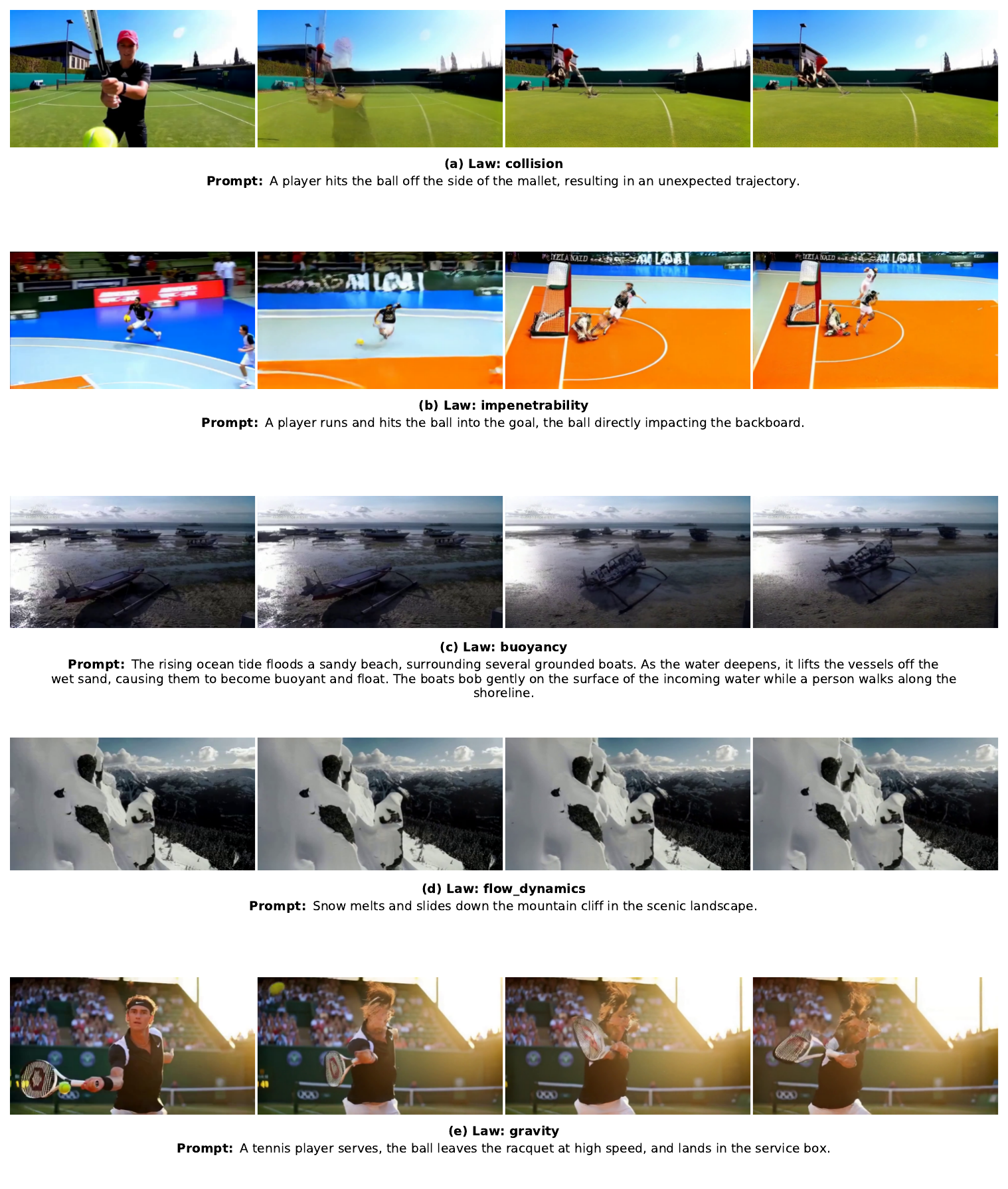}
    \caption{Qualitative examples of physical law violations from \textbf{LTX-2-19B}.}
    \label{fig:poor_video_model_ltx_19b}
\end{figure}

\begin{figure}[htbp]
    \centering
    \includegraphics[width=\linewidth]{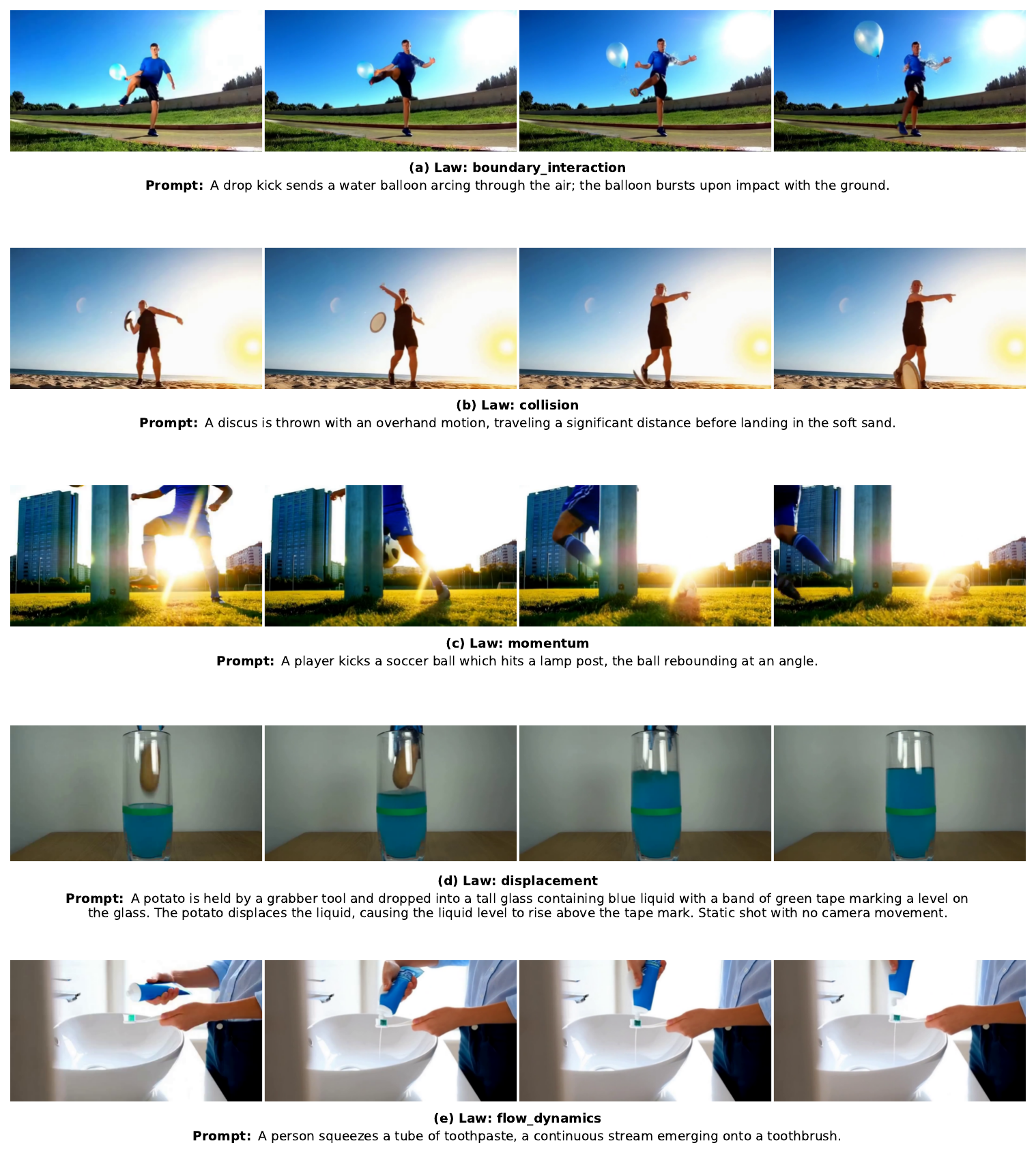}
    \caption{Qualitative examples of physical law violations from \textbf{LTX-2.3-22B}.}
    \label{fig:poor_video_model_ltx_22b}
\end{figure}

\begin{figure}[htbp]
    \centering
    \includegraphics[width=\linewidth]{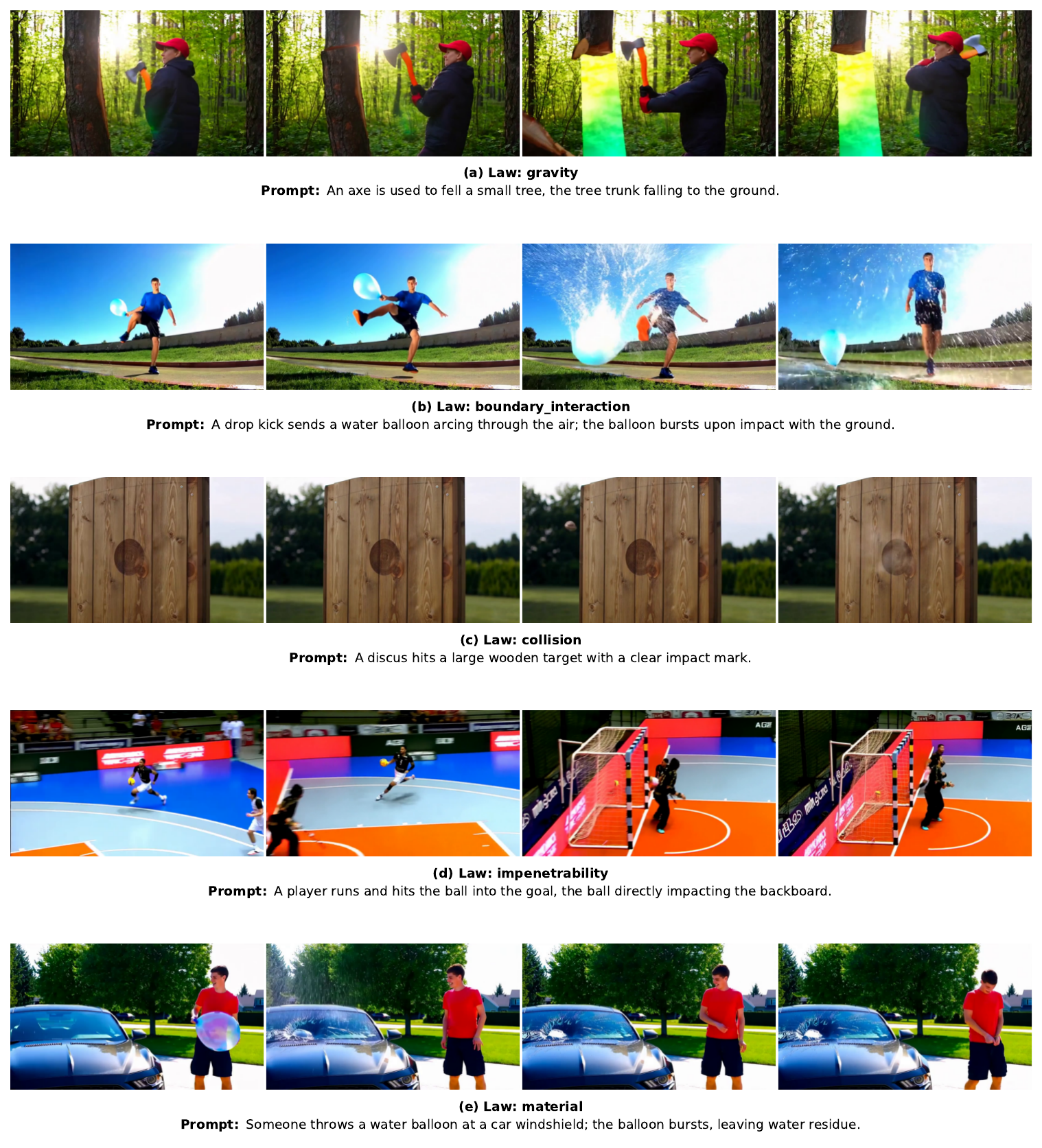}
    \caption{Qualitative examples of physical law violations from \textbf{Wan2.2-27B-A14B}.}
    \label{fig:poor_video_model_wan_i2v}
\end{figure}

\begin{figure}[htbp]
    \centering
    \includegraphics[width=\linewidth]{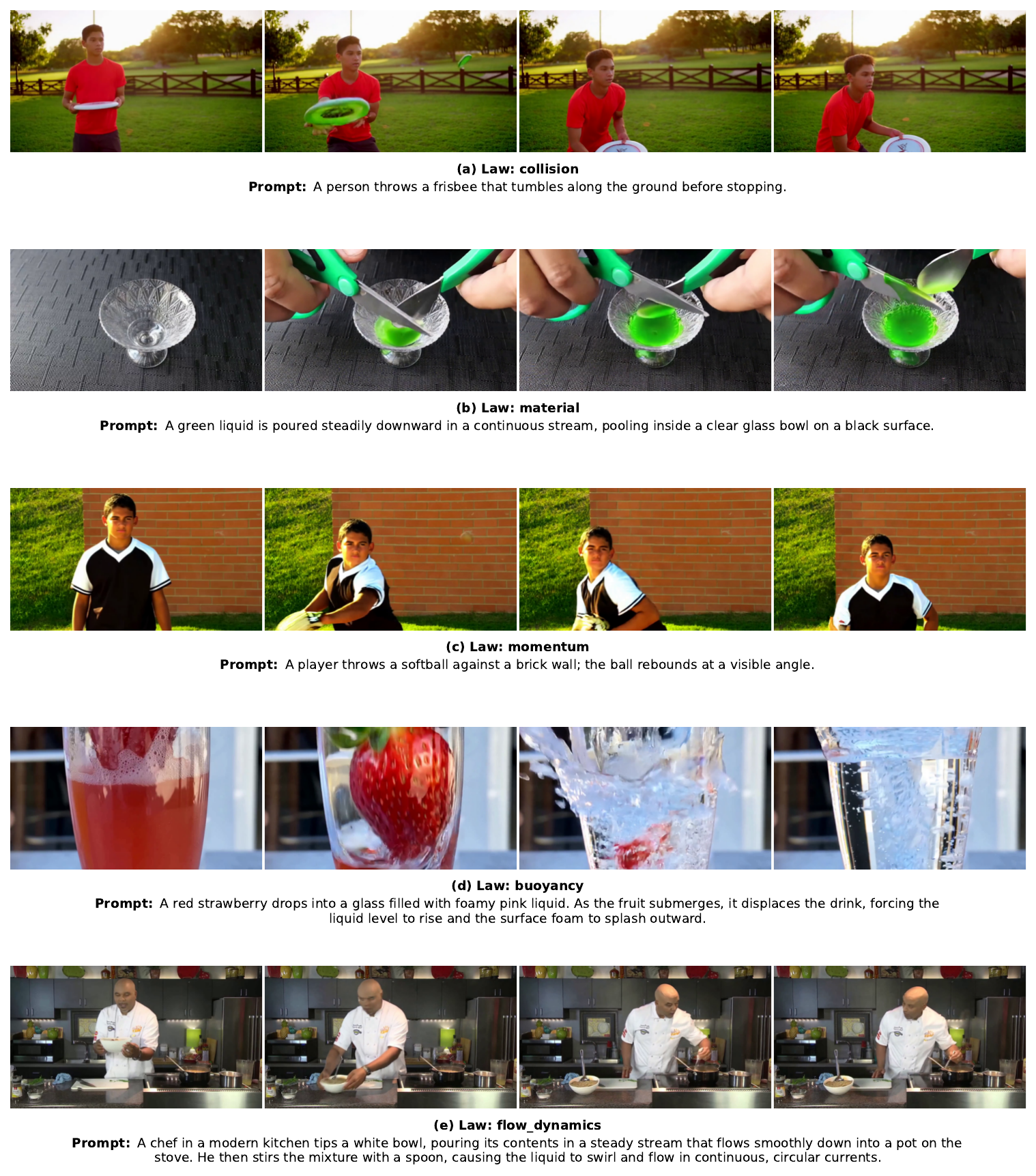}
    \caption{Qualitative examples of physical law violations from \textbf{Wan2.2-TI2V-5B}.}
    \label{fig:poor_video_model_wan_ti2v}
\end{figure}

We present more examples from each generative model where one or more physical laws are violated in \crefrange{fig:poor_video_model_veo}{fig:poor_video_model_wan_ti2v}. For each model, every annotator agreed that the labeled physical law was violated in the corresponding video.

